# From Labor to Collaboration: A Methodological Experiment Using AI Agents to Augment Research Perspectives in Taiwan's Humanities and Social Sciences

Dr. Yi-Chih HUANG (Corresponding Author)
Associate Researcher
National Applied Research Laboratories
Science & Technology Policy Research and Information Center
Contact Address: 1, 14-15F, No. 106, Heping E. Rd., Sec. 2, Taipei 10636, Taiwan (R.O.C.)
E-mail: yichuang@niar.org.tw

---

## Abstract

Generative AI is reshaping knowledge work, yet existing research focuses predominantly on software engineering and the natural sciences, with limited methodological exploration for the humanities and social sciences. Positioned as a "methodological experiment," this study proposes an AI Agent-based collaborative research workflow (Agentic Workflow) for humanities and social science research. Taiwan's Claude.ai usage data (N = 7,729 conversations, November 2025) from the Anthropic Economic Index (AEI) serves as the empirical vehicle for validating the feasibility of this methodology.

This study operates on two levels: the primary level is the design and validation of a methodological framework — a seven-stage modular workflow grounded in three principles: task modularization, human-AI division of labor, and verifiability, with each stage delineating clear roles for human researchers (research judgment and ethical decisions) and AI Agents (information retrieval and text generation); the secondary level is the empirical analysis of AEI Taiwan data — serving as an operational demonstration of the workflow's application to secondary data research, showcasing both the process and output quality (see Appendix A).

This study contributes by proposing a replicable AI collaboration framework for humanities and social science researchers, and identifying three operational modes of human-AI collaboration — direct execution, iterative refinement, and human-led — through reflexive documentation of the operational process. This taxonomy reveals the irreplaceability of human judgment in research question formulation, theoretical interpretation, contextualized reasoning, and ethical reflection. Limitations including single-platform data, cross-sectional design, and AI reliability risks are acknowledged.

Keywords: Generative AI, AI Agent, Human-AI Collaboration, Research Methodology, Agentic Workflow, Humanities and Social Sciences

---

# I. Introduction

## 1. Research Background

Since large language models (LLMs) entered the public eye in late 2022, the impact of generative artificial intelligence (Generative AI) on knowledge work has expanded from technical discussions to core issues in the social sciences. According to the Anthropic Economic Index (AEI), the tasks performed by global Claude.ai users span software development, academic research, content creation, business analysis, and various other knowledge-intensive activities (Anthropic, 2025; Appel et al., 2026), indicating that the use of AI tools is no longer the exclusive domain of specific technical communities but is diffusing across all categories of knowledge workers.

In the field of academic research, generative AI applications have likewise grown rapidly. Davidson and Karell (2025) noted that generative AI can serve as a "measurement" tool, a "prompting" tool, and a "simulation" tool for social science research, providing complementary analytical capabilities for traditional research methods. However, most studies focus on evaluating AI as a standalone tool, with relatively few examining how AI can be systematically integrated into the entire academic research process from the perspective of the "research workflow."

More importantly, existing discussions on AI-assisted research originate primarily from experiences in the natural sciences and engineering disciplines (Gao & Wang, 2024), with a notable lack of exploration regarding applicability to the humanities and social sciences. Research in the humanities and social sciences is characterized by a high degree of interpretivism, theory-building orientation, and context sensitivity — qualities that present methodological challenges for AI adoption that differ from those in the natural sciences. How to leverage AI's information processing and text generation capabilities while maintaining the quality of humanities and social science research is a question that urgently demands exploration.

## 2. Research Questions

Against this backdrop, the recent rise of the "AI Agent" concept offers a new direction for thinking. Unlike traditional AI tool usage (a single question-and-answer mode), AI Agents emphasize task-oriented autonomous execution capabilities, able to carry out multi-step tasks sequentially within a preset workflow according to the researcher's instructions (Guo et al., 2024; Zhang et al., 2024). This concept of "Agentic Workflow" aligns precisely with the characteristic of academic research processes that tasks can be decomposed and steps can be serialized, thereby providing humanities and social science researchers with a new model of human-AI collaboration.

However, there currently lacks a methodological framework for AI Agent collaboration tailored to the research context of the humanities and social sciences. Existing AI workflow designs are predominantly oriented toward software development scenarios (e.g., multi-agent frameworks such as AutoGen and CrewAI) and fail to adequately account for the distinctive needs of humanities and social science research: the criticality of literature review, the depth of theoretical dialogue, the plurality of data interpretation, and the sensitivity of research ethics.

Based on the above research background, this study poses the following two research questions:

Research Question 1: How can a collaborative AI Agent workflow suitable for secondary data research in the humanities and social sciences be designed? What are its design principles?

Research Question 2: In actual practice, how operable is this workflow? How are the boundaries of the division of labor between human researchers and AI Agents defined?

# 3. Research Objectives and Contributions

This study adopts the positioning of a "methodological experiment," the purpose of which is not to test specific hypotheses but rather to examine the feasibility of an AI Agent collaborative research workflow through its actual operation, to reveal its limitations, and to provide a referenceable methodological framework for subsequent research.

It is particularly important to note that this study has a dual-level structure:

- Primary level — Methodological experiment: The core concern of this study lies in the design, operation, and reflection upon the methodological framework itself. The seven-stage AI Agent collaborative workflow proposed in Section III, along with its underlying three design principles of "task modularization, human-AI division of labor, and verifiability," constitute the principal contribution of this study. Section IV validates the operability of this workflow through a description of the actual operational process and documents human researchers' interventional judgments at each critical juncture.
- Secondary level — AEI data analysis: The analysis of the Anthropic Economic Index Taiwan data serves as the empirical vehicle for validating the feasibility of the aforementioned methodology. This analysis is not the ultimate purpose of this study per se but rather the operational material and output of the methodological experiment. The complete data analysis report is presented in appendix form (Appendix A) for readers to examine the actual output quality of the workflow.

Specifically, the anticipated contributions of this study include:

First, methodological contribution: proposing a seven-stage AI Agent collaborative research workflow that covers the complete research process from research planning to reference management, grounded in the three principles of "task modularization," "human-AI division of labor," and "verifiability." This framework can serve as a reference for humanities and social science researchers when adopting AI tools.

Second, reflexive contribution: through comprehensive documentation of the AI collaborative research process (including human intervention points at each stage, the iterative history of prompt design, and the revision process of AI outputs), revealing the specific junctures at which "human judgment is irreplaceable" in human-AI collaboration, thereby providing firsthand case material for academic discussions on the ethics and transparency of AI use.

## 4. Overview of Research Methods

This study adopts a mixed-methods approach. At the methodological level, the AI Agent collaborative workflow is constructed in the spirit of Design Science and iteratively refined through practical operation. At the empirical level, descriptive analysis is employed to process the AEI Taiwan data, using frequency distributions, percentages, measures of central tendency, and measures of dispersion to profile the AI usage behavior patterns of Taiwanese users.

The primary data source used in this study is the publicly released Anthropic Economic Index dataset, focusing on Claude.ai usage records from the Taiwan region (geo_id: TW) during the period of November 13 to 20, 2025. The data encompass multiple analytical facets including task type, collaboration mode, AI autonomy, task success rate, and usage context, totaling 7,729 conversation records.

## 5. Paper Structure

The remaining sections of this paper are organized as follows: Section II conducts a literature review, surveying research on the application of generative AI in academic research, human-AI collaboration theory, and AI Agent workflows; Section III describes the research design and methodology, including the design principles of the workflow, data sources, and analytical methods; Section IV uses the AEI Taiwan data as the empirical vehicle to describe the actual operational process of the seven-stage workflow from a meta-analytical perspective and to identify three operational modes of human-AI collaboration; Section V presents a discussion analyzing the theoretical implications and practical insights derived from operational experience; Section VI summarizes the research contributions and proposes directions for future research. Additionally, Appendix A contains the complete analysis report produced through AI Agent collaboration, for readers to examine the actual output quality of the methodology.

---

# II. Literature Review

This chapter reviews three interrelated areas of the literature: the evolving role of generative AI in academic research processes, theoretical frameworks for human-AI collaboration, and technological developments in AI Agents and workflow modularization. Through a systematic review of the literature, this chapter aims to establish the theoretical positioning of this study and to identify gaps in existing research.

## 1. The Evolution of Generative AI and Academic Research Processes

The impact of generative AI on academic research has undergone a cognitive shift from "assistive tool" to "collaborative partner." Early research primarily positioned AI as an efficiency tool for text processing, such as functional applications including literature summarization, grammar correction, and translation assistance (Mondal et al., 2023). The AI usage pattern during this stage was characterized by "single-point intervention" — researchers used AI at specific junctures to complete specific tasks, with limited integration between AI and the research process.

As the capabilities of large language models improved, scholars began to explore deeper levels of AI participation in research processes. Davidson and Karell (2025) proposed three integration modes — measurement, prompting, and simulation — providing social science researchers with a systematic framework for AI adoption. The measurement mode treats AI as a coding tool for the classification and annotation of large-scale textual data; the prompting mode leverages AI's language generation capabilities to explore possible inferences from theoretical hypotheses; and the simulation mode uses AI as a simulator of social behavior to generate synthetic data for analysis.

In the fields of management and organizational research, the experimental study by Dell'Acqua et al. (2023) found that management consultants using AI significantly outperformed those not using AI on specific tasks, but their performance actually declined on tasks that exceeded AI's capability boundaries — a phenomenon the researchers termed the "jagged frontier" effect. This finding carries important implications for academic research: the adoption of AI does not result in a comprehensive enhancement of capabilities but rather yields differentiated benefits across different task types, and researchers need the judgment to discern the boundaries of AI capabilities.

Mollick and Mollick (2023) further noted that in educational settings, the effective use of AI is highly dependent on the user's prompt design skills and task decomposition strategies. This observation applies equally to academic research: whether researchers can effectively leverage AI depends on their ability to decompose complex research questions into operable sub-tasks, rather than solely on the technical capabilities of the AI tool itself.

The most demonstratively significant practical case comes from Stanford political scientist Hall. Hall (2026a) used an AI coding agent (Claude Code) to fully replicate and extend a published political science empirical paper — Thompson et al.'s (2020) study on the effects of universal vote-by-mail on voter turnout and election outcomes — in approximately one hour. Upon independent review, Straus and Hall (2026) found that the AI's replication results were highly accurate: all 12 regression coefficients were precisely replicated to three decimal places, and the correlation coefficient between the AI-collected election data and the original data exceeded 0.999. However, the review also revealed the boundaries of AI capabilities: among the treatment status codings for 30 California counties, the AI misidentified the treatment timing for 1 county; and when attempting new analyses beyond the scope of the original paper, the AI's performance noticeably declined — not by producing "hallucinations," but by deviating from the intent of the original prompt, generating analyses that were insufficiently rigorous in design.

Hall's experiment provides two key insights for the present study: first, AI Agents have demonstrated remarkable execution capabilities in empirical research tasks that are "well-structured and clearly bounded," capable of substantially compressing the time costs of data collection and analysis; second, the boundaries of AI capabilities emerge precisely at junctures requiring research judgment — when tasks shift from "replicating existing analyses" to "designing new analyses," human researchers' guidance and supervision become indispensable. This observation provides direct empirical support for the "human-AI division of labor" principle advocated in this study.

## 2. From Tool to Collaboration: Theoretical Frameworks for Human-AI Collaboration

Human-AI Collaboration theory provides an analytical framework for understanding the role of AI in research processes. The existing literature presents three main perspectives that both compete and complement one another.

The Tool Perspective treats AI as an extension tool of the researcher, emphasizing human dominance. Under this perspective, AI is an object "to be used," and its value lies in enhancing the work efficiency of human researchers. Brynjolfsson and McAfee's (2014) "Second Machine Age" discourse exemplifies this view, arguing that AI's core value lies in automating repetitive cognitive labor, thereby freeing humans to engage in more creative work.

The Collaboration Perspective moves beyond the tool metaphor and conceives of human-AI interaction as a complementary collaborative relationship. Dellermann et al. (2019) proposed the concept of "Hybrid Intelligence," arguing that humans and AI each possess cognitive advantages — humans excel at abstract reasoning, contextual judgment, and ethical decision-making, while AI excels at large-scale information processing, pattern recognition, and consistent execution. In an ideal collaborative design, the capabilities of both parties form a complementary rather than substitutive relationship.

The Agency Perspective is a newer viewpoint that has emerged in recent years alongside advances in AI Agent technology. Unlike passive tools that await instructions, AI Agents are endowed with a certain degree of autonomous decision-making capability, able to plan and execute task steps autonomously within a preset goal framework (Wang et al., 2024). Shavit et al. (2023) discussed the ethical issues of AI agentic behavior, noting that when AI possesses the capacity for autonomous action, human responsibility for supervising and verifying AI outputs increases rather than diminishes.

This study posits that in the domain of academic research, the above three perspectives are not mutually exclusive but rather correspond to different characteristics of research tasks. For structured, repetitive tasks (such as reference formatting and data cleaning), the efficiency framework of the Tool Perspective is sufficient; for tasks requiring creative input (such as literature analysis and argument construction), the complementary framework of the Collaboration Perspective is more appropriate; and for multi-step, serialized research processes, the autonomous execution framework of the Agency Perspective offers new possibilities. The workflow design proposed in this study seeks to strike a balance among these three perspectives.

## 3. AI Agents and Research Process Modularization

AI Agents emphasize goal-oriented autonomous action, capable of automatically decomposing tasks, selecting tools, and executing iteratively based on preset objectives (Yao et al., 2023), and multi-agent systems can further enable different Agents to each fulfill distinct roles (Guo et al., 2024). Zhang et al. (2024) noted that an effective Agentic Workflow must satisfy three conditions: task decomposability, explicit dependency relationships, and verifiable outputs — conditions that align precisely with the characteristics of academic research processes. Gao et al. (2024) found that AI Agents show potential in hypothesis generation and data analysis but have limited capacity for theory construction and ethical judgment, supporting the necessity of a human-AI division of labor.

In his essay "The 100x Research Institution," Hall (2026b) put forward a more forward-looking vision. Based on the experiment of using AI Agents to replicate a political science paper (Straus & Hall, 2026), he noted that in well-structured empirical tasks, the performance gap between human researchers and AI Agents has become extremely small — AI replication results are highly consistent with those produced manually. He argued that every empirical paper should be accompanied by proof of automated AI Agent replication prior to publication, transforming research from a static product into continuously updated "living research infrastructure." Hall estimated that the cost of a single research task is approximately 10 USD, with annual API expenses under 5,000 USD, making it possible for senior scholars to direct hundreds of Agents to conduct large-scale research.

However, this vision also invites critical reflection. Karpf (2026) expressed concern that disciplines would gravitate toward studying "problems that are easy for AI to handle" rather than genuinely important problems; Gunitsky (2026) argued that what AI automates is normal science, offering limited benefit for breakthrough research. Domestic discussions on this topic remain in their nascent stages, lacking actionable methodological guidance — this gap is precisely the space that the present study seeks to fill.

## 4. Limitations and Gaps in Existing Research

Drawing together the above literature review, this study identifies the following research gaps:

Contextual gap: Existing literature on AI-assisted research originates predominantly from European and American academic settings, with a lack of empirical descriptions of AI usage patterns in East Asia (particularly Taiwan). The distinctive features of Taiwan's academic environment — including bilingual (Chinese-English) research demands, limited research resources, and a unique academic evaluation system — may result in AI usage patterns that differ from those in Europe and North America, necessitating localized empirical research for exploration.

Methodological gap: Although the technical frameworks for AI Agents and Agentic Workflows are increasingly mature, methodological designs tailored to research contexts in the humanities and social sciences remain lacking. Existing workflow frameworks are predominantly oriented toward software development or natural science scenarios and fail to adequately account for the distinctive needs of humanities and social science research, including the depth required for critical literature review, the complexity of theoretical dialogue, and the high sensitivity of research ethics.

Practical gap: Academic discussions of AI collaborative research have largely remained at the conceptual level ("how AI can help research"), lacking comprehensive practice documentation and reflexive analysis. What researchers need is not merely introductions to AI functionalities but rather an operable, replicable, and verifiable workflow, together with an honest disclosure of its strengths and limitations as revealed through actual practice.

It is precisely on the basis of these three gaps that this study attempts to propose an AI Agent collaborative methodological framework for secondary data research in the humanities and social sciences, with an operational demonstration using empirical data from the Taiwan context.

# III. Research Design and Methodology

This chapter describes the methodological positioning of the present study, the design principles of the AI Agent collaborative workflow, data sources and analytical methods, and research ethics considerations.

## 1. Positioning as a Methodological Experiment

This study adopts the research orientation of a "methodological experiment" rather than a traditional hypothesis-testing study. The core objective of a methodological experiment is to examine the feasibility and limitations of a new research method through its actual implementation, thereby providing the academic community with replicable practical experience (Hevner et al., 2004).

Specifically, the "experiment" in this study encompasses three levels: first, the design level—proposing a workflow framework for AI Agent collaborative research; second, the operational level—using AEI Taiwan data as material to actually execute the data analysis component of this workflow; and third, the reflective level—documenting decision points, difficulties, and discoveries encountered during the operational process as reflexive material for methodological inquiry.

It must be emphasized that the empirical analysis results of this study (Section IV) should be regarded as a "method demonstration" rather than "research findings." Their purpose is to demonstrate how the AI Agent collaborative workflow operates in practice, not to offer causal explanations of AI usage behavior in Taiwan.

## 2. Research Tool Selection: Claude Code as the Collaborative Interface

This study selected Anthropic's Claude Code as the primary operational interface for AI Agent collaboration, with the underlying model being Claude Opus 4.6 (model ID: claude-opus-4-6), Anthropic's most advanced reasoning model released during 2025–2026. The draft of this paper was completed in the Claude Code environment in February 2026.

Claude Code is a browser-based AI programming and research collaboration environment in which researchers can issue instructions in natural language, and the AI Agent performs multiple operations including file reading and writing, data analysis, code generation, and web searching, with all operational histories tracked through the Git version control system.

### Environment Setup

The prerequisite steps for using Claude Code are: (1) register a free account on GitHub (https://github.com/) as the foundation for version control and file tracking; (2) subscribe to the Max plan on the official Anthropic website (https://claude.ai/) to obtain access to Claude Code; (3) after logging in, link the GitHub account and enter the Claude Code browser environment; (4) use natural language instructions to establish the research project directory structure. Upon completion, researchers can drive the AI Agent to execute research tasks through conversational interaction.

**Rationale for Selecting Claude Code**

The selection of Claude Code over the Command Line Interface (CLI) or other tools was based on three considerations: First, lowering the technical barrier—by providing a graphical browser interface, researchers can complete data analysis through natural language without needing to install a Python environment or learn command-line syntax; Second, an integrated working environment—file management, code execution, web searching, and version control are integrated into a single interface, with all operations automatically saved in the Git version history; Third, a conversational interaction mode—based on multi-turn natural language dialogue, this mode is highly compatible with the "iterative revision" work habits in academic research.

# 3. AI Agent Collaborative Workflow Design

**(1) Design Principles**

The workflow proposed in this study is based on three core design principles:

Principle 1: Task Modularization. The complete research process is decomposed into clearly defined sub-task modules, each with explicit inputs, processing procedures, and outputs. The advantages of modular design are: (1) reducing the complexity of individual tasks so that the AI Agent can operate effectively within a well-defined scope; (2) providing verifiable intermediate outputs that facilitate quality control by human researchers at each node; and (3) enhancing the reproducibility of the research process, enabling other researchers to replicate the same modular workflow.

Principle 2: Human-AI Division of Labor. Within each task module, the respective responsibilities of the human researcher and the AI Agent are clearly delineated. The basic division-of-labor logic is as follows: humans are responsible for "judgmental" tasks (research question definition, theoretical interpretation, ethical decision-making, and final quality assurance), while the AI handles "executive" tasks (information retrieval, data processing, formatting, and draft text generation). This division of labor echoes the hybrid intelligence framework proposed by Dellermann et al. (2019), fully leveraging the respective cognitive strengths of humans and AI.

Principle 3: Verifiability. All outputs of the AI Agent must undergo review and verification by the human researcher, with complete operational records retained (including prompt templates, raw AI outputs, and human-revised versions). Practices for implementing the verifiability principle include: using the Git version control system to track all revision histories, establishing verification checklists at each stage, and explicitly disclosing AI usage in the paper.

**(2) Seven-Stage Workflow**

Based on the above principles, this study designed a seven-stage AI Agent collaborative research workflow. The content, human-AI division of labor, and expected outputs for each stage are presented in Table 1.

Table 1: Seven-Stage Design of the AI Agent Collaborative Research Workflow

| Stage | Name | Human Role | AI Agent Role | Expected Output |
|-------|------|-----------|---------------|-----------------|
| 0 | Research Planning and Agent Configuration | Define research questions, determine data sources | Assist in structured thinking, establish document architecture | Research proposal, project structure |
| 1 | Literature Collection | Define search scope, verify relevance | Execute searches, organize literature lists | Structured literature database |
| 2 | Literature Analysis | Theoretical interpretation, verify analytical conclusions | Thematic analysis, gap identification | Literature analysis report |
| 3 | Data Understanding and Exploration | Understand data semantics, define | Read data, descriptive statistics | Data structure documentation |
| | | analytical directions | | |
| 4 | Data Analysis and Visualization | Define analytical questions, interpret results | Execute analyses, generate charts | Analytical results and charts |
| 5 | Paper Writing | Review content, theoretical interpretation | Draft each chapter section | Paper draft |
| 6 | Reference Management | Supplement missing information, confirm formatting | Extract citations, format references | Reference list |

This seven-stage design is sequential (each subsequent stage depends on the outputs of the preceding stage) but not strictly linear—researchers may iterate between stages as needed. For example, when new literature needs are discovered during the data analysis stage (Stage 4), one may return to the literature collection stage (Stage 1) for supplementation.

**(3) Agent Role Design**

Within the workflow, this study designed five specialized AI Agent roles for different stages: (1) Literature Collection Agent—systematically searches for and organizes academic literature; (2) Literature Analysis Agent—identifies core arguments, compares perspectives, and identifies research gaps; (3) Data Exploration Agent—reads data structures and generates descriptive statistics; (4) Data Analysis Agent—executes statistical analyses and generates visualizations; (5) Academic Writing Agent—drafts individual chapters and sections of the paper.

Each Agent role includes an explicit "must-not-do" list (negative constraints), which serves as a critical mechanism for preventing AI from overstepping its boundaries. For example, the Academic Writing Agent is required to "not add any literature that has not been provided by the human researcher," in order to mitigate the risk of hallucinated references.

# 4. Data Sources and Analytical Methods

**(1) Data Source: Anthropic Economic Index (AEI)**

The empirical data used in this study come from the fourth edition of the economic index report published by Anthropic (Anthropic Economic Index, 4th edition; Appel et al., 2026). The AEI is a large-scale dataset tracking usage behavior on the Claude.ai platform, providing descriptions of AI usage patterns across global regions through anonymized analysis of user conversation records. The report proposes five "economic primitives"—task complexity, skill level, use purpose (work, education, or personal), AI autonomy, and task success rate—as foundational measurement indicators for tracking the economic impact of AI, based on a privacy-preserving analysis of approximately two million AI conversations, encompassing both consumer-side (Claude.ai) and enterprise-side (API) usage data.

The Taiwan subset on which this study focuses has the following characteristics:

- Data collection period: November 13, 2025 to November 20, 2025 (one week)
- Platform and product: Claude AI (Free and Pro versions)
- Geographic scope: Taiwan (geo_id: TW)
- Total number of conversations: N = 7,729
- Proportion of global total: 0.77%
- Data format: Long format, where each row represents an indicator value for a specific (facet, variable, cluster_name) combination

**(2) Analytical Facets**

The AEI data employs a multi-facet analytical framework. The analytical facets used in this study include:

Categorical facets:

- request (task request type): A three-level classification, where level 0 is the finest granularity (614 categories) and level 2 is the coarsest granularity (22 categories)
- collaboration (collaboration mode): Six types of human-AI collaboration—directive, learning, task iteration, feedback loop, validation, and none

- use_case (use context): work, personal, and coursework
- task_success (task success rate): yes or no
- multitasking: Whether multiple tasks are handled within a single conversation
- human_only_ability (human-only completion capability): Whether the task can be completed independently by humans

Numerical facets:

- ai_autonomy (AI autonomy): A 1–5 scale measuring the degree of AI autonomy in the task
- human_education_years (human education years): The estimated years of human education required to complete the task
- ai_education_years (AI education years): The equivalent years of education demonstrated by the AI
- human_only_time (human-only completion time): The estimated time for a human to complete the task without AI assistance (in hours)
- human_with_ai_time (human-AI collaborative time): The estimated time for a human to complete the task with AI assistance (in minutes)

In addition, this study also uses two summary datasets: grouped by task category (Group by Category, 13 categories) and grouped by occupational classification (Group by Job, 14 categories, based on the Standard Occupational Classification [SOC] system of the U.S. Occupational Information Network [O*NET]).

## (3) Analytical Methods

Given the methodological experiment positioning of this study, the data analysis primarily employs descriptive analysis, specifically including:

1. Frequency distribution and percentage analysis: Depicting the distributional characteristics of each categorical facet
2. Central tendency and dispersion: Describing the distributions of numerical facets using means, medians, and standard deviations
3. Semantic matching analysis: Using keyword matching to identify task types related to academic research, estimating the proportion of academic usage
4. Visual presentation: Presenting analytical results through bar charts, pie charts, grouped bar charts, and other visual formats

All analyses were executed in the Python programming language, driven by natural language instructions within the Claude Code environment, using pandas (a data analysis library) and matplotlib (a visualization plotting library). The researcher described analytical needs in Chinese, and the AI Agent automatically generated corresponding Python scripts and executed them in real time. The resulting charts were output to the project's `charts/` directory, and the entire analytical process was tracked by Git (a distributed version control system) to ensure reproducibility.

# 5. Ethical Considerations and Research Limitations

**(1) AI Usage Disclosure**

This study used AI Agent assistance in the following aspects:

1. Literature collection and preliminary organization (AI assisted with searching and classification; human verified relevance)
2. Data analysis script generation (AI produced Python scripts; human reviewed the logic before execution)
3. Paper chapter draft writing (AI produced initial drafts; human conducted substantive revision and theoretical interpretation)
4. Reference formatting (AI assisted with format adjustments; human verified accuracy)

All AI outputs were reviewed, verified, and revised by the human researcher. The core judgments of the study—including research question definition, theoretical interpretation, data interpretation, research limitation assessment, and final conclusions—were all made by the human researcher.

**(2) Data Ethics**

The AEI data are anonymized aggregated data publicly released by Anthropic and do not involve personally identifiable information. The smallest unit of observation in the data is the aggregated statistical value for a specific region and specific facet, rather than individual users' conversation records. Furthermore, the AEI report states that the data have undergone privacy protection processing, and data with observation counts below the threshold (200 at the national level, 100 at the regional level) have been excluded.

**(3) Preliminary Statement of Methodological Limitations**

This study has the following inherent limitations at the methodological level, which will be further elaborated in the discussion in Section V:

1. Single-platform limitation: The data come from only one platform, Claude.ai, and cannot represent the overall AI usage landscape in Taiwan

2. Cross-sectional limitation: Covering only one week of usage data, it is unable to capture temporal dynamic changes
3. Aggregated data limitation: Individual-level cross-tabulation or regression analysis cannot be conducted
4. Classification system limitation: Task types and occupational classifications are based on AI automated labeling and may contain classification errors
5. Self-selection bias: Claude.ai users do not constitute a random sample of academic workers in Taiwan

# IV. Empirical Illustration: A Meta-Analysis of the Workflow Operational Process

This chapter adopts a meta-analytic perspective to document the operational process of executing the seven-stage workflow described in Chapter III, using the AEI Taiwan dataset as source material. The core concern of this chapter is not "what the data reveal" but rather "how the Agent processes data and how humans intervene." Complete data analysis results are included in Appendix A.

To enhance reproducibility, this chapter presents representative prompts used by the researcher at each of the seven stages, annotated with their corresponding operational mode types. Based on operational experience, this study identifies three types of human-AI collaborative operational modes (Table 2).

Table 2: Classification of AI Agent Collaborative Operational Modes

| Operational Mode | Characteristics | Human Cognitive Investment |
|---|---|---|
| Direct Execution | Agent independently completes tasks based on explicit instructions; human only needs to confirm output | Low |
| Iterative Refinement | Agent's initial output requires multiple rounds of improvement following human review | Medium |

| Human-Led | Analytical direction and judgment logic are determined by the human; Agent is responsible only for execution | High |
| --- | --- | --- |

The following sections present prompt examples and meta-observations for each type, organized by the actual operations at each stage.

# 1. Stage 0: Research Planning and Agent Configuration

Prompt 0-1 [Direct Execution]: "I am conducting a study using AEI's Taiwan data to analyze the behavioral characteristics of Taiwan's Claude.ai users. Please set up the directory structure for the research project, including three folders—manuscript, figures, and data—and generate a preliminary analysis planning document."

The Agent automatically created the directories and produced a generic analysis plan. However, the following three decisions were made independently by the human [Human-Led]: (1) deciding to focus on "descriptive analysis" rather than "causal inference"; (2) selecting analysis facets relevant to the research questions from among multiple available facets; and (3) personally downloading the raw data from the Anthropic official website to confirm its authenticity. This stage demonstrates that the Agent can handle structured administrative tasks, but substantive judgments in research design still require human domain knowledge.

# 2. Stage 1: Literature Collection

Prompt 1-1 [Direct Execution]: "Please search for academic literature from the past three years (2023–2026) on the following topics: (1) generative AI in social science research, (2) human-AI collaboration in academic writing, (3) agentic workflow. Find at least 5 peer-reviewed publications for each topic, listing the author, year, title, journal, and abstract."

After the Agent produced the literature list, the human intervened as follows [Human-Led]:

Prompt 1-2 [Human-Led]: "I was unable to find items 3 and 7 from the literature list on Google Scholar. Please verify whether these two references actually exist. If they do not exist, state so explicitly rather than fabricating substitute references."

This stage revealed one of the most serious risks associated with the Agent—hallucinated references. In its initial literature list, the Agent interspersed nonexistent references that were complete in format and appeared credible. The human researcher must cross-verify the authenticity of every reference one by one; this is an indispensable quality gate.

[Failure Recovery Case Study: Detection and Correction of Hallucinated References]

The following is a complete record of a typical "hallucinated reference" error recovery process, serving as a concrete demonstration of how quality gates operate within an agentic workflow.

(1) Error Discovery: After responding to Prompt 1-1, the Agent produced a list of 15 references, complete in format (including author, year, journal, and DOI), appearing authoritative and credible. When the researcher verified each entry via Google Scholar, two could not be found—the journal names were real, the authors were active scholars in the field, yet the specific papers had never been published. This is a classic case of "high-verisimilitude hallucination": the AI does not fabricate randomly but rather combines statistical patterns to construct fictitious entries that "most resemble real references."

(2) Diagnostic Questioning:

Prompt 1-2a [Human-Led]: "I was unable to find items 3 and 7 from the literature list on Google Scholar. Please verify one by one whether these two references actually exist. If they do not exist, please state directly: 'This reference does not exist; it is a model generation error.' Do not substitute other references."

(3) Agent Response and Correction: Upon receiving explicit instructions, the Agent acknowledged that the two references "cannot be confirmed to exist" and explained that they may have been "improperly generated based on similar reference characteristics." The researcher then requested supplementary replacement references with additional constraints:

Prompt 1-2b [Human-Led]: "Please provide two replacement references under the following conditions: (1) a complete DOI link must be provided, (2) I will immediately verify their existence via the DOI link, (3) if you cannot provide a DOI, please annotate 'This reference requires manual verification.'"

(4) Verification Result: The two replacement references supplemented by the Agent both included DOIs, which the researcher verified in real time by clicking through, confirming their existence.

(5) Methodological Implications: This case reveals three operational principles—*First*, the literature lists produced by the Agent should undergo "complete verification" rather than sampling inspection, because hallucinated references have extremely high verisimilitude and cannot be distinguished from genuine ones based on formatting alone; *Second*, correction instructions should explicitly prohibit the Agent's "automatic compensation" behavior (such as substituting other references without being asked), to prevent errors from propagating through iterations; *Third*, requiring the Agent to provide verifiable anchor points (such as DOI links) is an effective mechanism for reducing hallucination risk. This four-step recovery process of "discovery—diagnosis—constraint—verification" can serve as a general paradigm for handling content errors in agentic workflows.

## 3. Stage 2: Literature Analysis

Prompt 2-1 [Iterative Refinement]: "Please conduct a thematic analysis of the verified literature. Identify three core themes: (1) the role positioning of AI in academic research, (2) theoretical frameworks for human-AI collaboration, (3) design principles of agentic workflows. For each theme, list the supporting references and core arguments."

The Agent's initial draft of the thematic analysis was largely accurate in its classification of references but lacked critical depth—it failed to identify contradictory viewpoints and research gaps across the literature. The researcher needed to follow up:

Prompt 2-2 [Human-Led]: "Within the 'AI role positioning' theme, what are the fundamental differences between the viewpoints of Davidson & Karell (2025) and Bail (2024)? Please identify their divergences regarding the applicability of AI as a research tool, as well as the implications of this divergence for the methodological design of the present study."

This stage confirmed that the Agent excels at "classification" but not at "critique"; the depth of theoretical dialogue requires human guidance.

## 4. Stage 3: Data Understanding and Exploration

Prompt 3-1 [Direct Execution]: "Please read the Excel files in the data/ folder and list each file's column names, number of records, and data types."

The Agent's first attempt failed due to header formatting issues (column names were located in the second row), requiring human correction:

Prompt 3-2 [Iterative Refinement]: "The header is in the second row. Please re-read using the header=1 parameter."

The more critical challenge lay in the Agent's semantic understanding of long-format data:

Prompt 3-3 [Human-Led]: "Note: the same conversation appears repeatedly under different facets. You cannot sum counts across different facets. The total number of conversations should be calculated from a single facet. Each facet is an independent analytical dimension."

The operations at this stage revealed three cognitive levels in the Agent's data comprehension: the syntactic level (reading column names) can be completed after parameter correction; the structural level (understanding hierarchical relationships in long-format data) requires human semantic supplementation; the semantic level (understanding the disciplinary meaning of facets) depends entirely on the human.

## 5. Stage 4: Data Analysis and Visualization

Prompt 4-1 [Direct Execution]: "From the 'request' facet, calculate the count and share for each cluster_name. Present the results as a horizontal bar chart. Use WenQuanYi Zen Hei for Chinese characters, set the chart to 12x8 inches, and save to figures/figure1_request_categories.png."

The Agent translated the natural language into a complete Python script (pandas + matplotlib) and correctly completed the data filtering and statistical calculations on the first attempt. However, the chart presentation required iterative refinement:

Prompt 4-1a [Iterative Refinement]: "The Chinese characters are displaying as squares. Please use matplotlib.font_manager to locate available Chinese font paths on the system and then specify the path. The category labels are overlapping—please reduce the font size to 9pt and increase spacing."

On average, each chart required 3–5 rounds of such corrections. Additionally, semantic judgments in analytical logic constituted critical points of human intervention:

Prompt 4-2 [Human-Led]: "Please combine the two categories 'Assist with academic research, writing, and educational content' and 'Assist with academic research, writing, and interdisciplinary courses,' and calculate the total share of academic research-related tasks."

This merging decision—judging that two similarly named categories conceptually belong to the same domain—reflects the human's semantic understanding of the classification system and is not something the Agent can accomplish autonomously. Similarly, screening which items among dozens of Request L1 task categories qualify as "humanities and social sciences-related" also constitutes a human semantic judgment:

Prompt 4-3 [Human-Led]: "From the Request L1 task categories, please filter items directly related to humanities and social sciences academic research, present their shares as a horizontal bar chart, and calculate the aggregate share. Selection criteria: translation, academic writing, education and teaching, research methods, literature processing, and other items directly related to humanities and social sciences academic work."

**Empirical Findings on Efficiency Augmentation**

With the assistance of the data analysis Agent, this study was able to rapidly extract the most impactful efficiency metrics from the AEI data. Table 3 presents time-benefit data for Taiwan users under AI assistance; this set of data most concretely demonstrates the potential of agentic workflows to liberate researchers from tedious labor.

Table 3: Time Benefits of AI Assistance—Taiwan Claude.ai Users (N=7,729)

| Metric | Value |
|---|---|
| Human solo completion time (median) | *Mdn* = 1.75 hours (105 minutes) |
| AI-assisted completion time (median) | *Mdn* = 12.0 minutes |
| Median time savings rate | Approximately 89% |
| Human solo completion time (mean) | *M* = 3.55 hours, 95% CI [3.43, 3.68] |
| AI-assisted completion time (mean) | *M* = 18.7 minutes, 95% CI [18.2, 19.2] |

By median, AI compressed the typical task completion time from 1.75 hours to 12 minutes—approximately one-ninth of the original duration. The magnitude of this efficiency gain warrants careful consideration: if a humanities and social sciences researcher has three structured tasks per day that can be AI-assisted (literature translation, data organization, format editing), the median estimate suggests approximately 4.65 hours of cognitive resources could be freed daily, redirected toward theoretical reflection, fieldwork observation, or interdisciplinary dialogue—high-value work that AI cannot yet perform.

More critically, the operational efficiency of the data analysis Agent itself deserves attention. During this study's operational process, from "describing the analytical requirement in natural language" to "obtaining complete statistical results and charts," the average time per single analysis iteration was approximately 2–3 minutes. In traditional research workflows, researchers must write their own Python or R scripts, debug, and adjust chart parameters—equivalent work often takes several hours. The efficiency advantage of the agentic workflow derives not from "faster computation" but from "real-time translation from natural language to code," enabling researchers without programming backgrounds to drive data analysis directly using academic language. This finding resonates with the "augmentation" perspective of Brynjolfsson and McAfee (2014)—the core value of AI lies in augmenting the radius of researchers' analytical capabilities rather than replacing their research judgment. A complete analysis of time benefits is provided in Appendix A, Section 6.

The meta-observation for this stage is: The Agent can execute any clearly defined computation, but the decision of "what should be computed" still requires human domain knowledge. Complete results for each analysis are provided in Appendix A.

# 6. Stage 5: Manuscript Writing

Prompt 5-1 [Direct Execution]: "Based on the above analysis results, please write a data analysis report. Structure: Section 1—Overall Profile, Section 2—Usage Scenario Distribution, Section 3—O*NET and Task Classification Methods, Section 4—Task Type Rankings, Section 5—O*NET Occupational Task Analysis, Section 6—Education Level and Time Benefits, Section 7—Implications for Humanities and Social Sciences, Section 8—Conclusions and Recommendations. Cite corresponding figures and tables; use formal Traditional Chinese academic style."

The Agent produced an initial draft of approximately 8,000 words with accurate data citations, but theoretical connections required human specification:

Prompt 5-2 [Human-Led]: "In the paragraph on the analysis of human independent completion capability, please add a theoretical interpretation: this finding resonates with the 'augmentation' perspective of Brynjolfsson and McAfee (2014)—the primary value of AI lies in augmenting efficiency rather than replacing capability. Use this interpretation to contextualize the 82.9% human independent completion rate."

The Agent successfully integrated the theoretical framework into the text, but the appropriateness of the interpretation still required human confirmation. The researcher's revision of the initial draft amounted to approximately 30–40%, concentrated on theoretical connections and argumentative logic. This stage demonstrates: The Agent is an "assembler" of theoretical prose, not a "creator"; the revised final version is included in Appendix A.

# 7. Stage 6: Reference Management

Prompt 6-1 [Direct Execution]: "Please extract all cited references from each chapter of the paper and compile them into a reference list formatted according to the 7th edition of the American Psychological Association (APA) style, sorted alphabetically by author surname."

The Agent correctly extracted and formatted most references, but iterative refinement was needed:

Prompt 6-2 [Iterative Refinement]: "Please check whether the Digital Object Identifier (DOI) link for each reference in the list is correct. Standardize the format to begin with https://doi.org/. References lacking a DOI should be annotated accordingly."

The critical human intervention at this stage [Human-Led] involved supplementing publication information that the Agent could not access—for example, formal publication details for preprints, translation title formats for Chinese-language references, and confirming that all cited references actually exist.

## 8. Meta-Analysis Summary

**(1) Stage Distribution of Operational Modes**

Drawing on the operational experience across all seven stages, the frequency of each operational mode exhibits a systematic distribution (Table 4).

Table 4: Stage Distribution of Operational Modes

| Stage | Direct Execution | Iterative Refinement | Human-Led |
|---|---|---|---|
| 0 Research Planning | ● | | ●● |
| 1 Literature Collection | ● | | ●● |
| 2 Literature Analysis | | ● | ●● |
| 3 Data Exploration | ● | ● | ●● |
| 4 Data Analysis | ●● | ●● | ●● |
| 5 Manuscript Writing | ● | | ●● |

| 6 Reference Management | ●● | ● | ● |

*Note: ● indicates the relative frequency of each mode's occurrence at that stage.*

As the table shows, Human-Led operations pervade all stages, validating the human-AI division of labor principle presented in Chapter III. Direct Execution is concentrated primarily in stages with a high degree of structure (0, 4, 6); Iterative Refinement is concentrated in stages involving presentation quality (3, 4, 6).

**(2) Reproducibility and Limitations**

The prompt sequences provided in this chapter can be used by subsequent researchers to reproduce similar operational processes with different datasets. However, because large language model (LLM) generation is stochastic, the same prompt may produce slightly different responses at different points in time. Therefore, the prompts above should be understood as "references for operational logic" rather than "scripts for exact replication."

# V. Discussion

This chapter discusses the operational experience of the methodological experiment described in the preceding sections (Chapter IV) from three perspectives: implications for research process management in the humanities and social sciences, insights for research practice, and limitations and risks of integrating AI into academic research. The empirical analysis results presented in Appendix A, as outputs of the workflow, serve only as supplementary evidence in this chapter.

# 1. Implications for Research Process Management in the Humanities and Social Sciences

**(1) A Paradigm Shift from "Single-Point Tools" to "Process Architecture"**

The seven-stage AI Agent collaborative workflow proposed in this study represents a mode of thinking distinct from existing discussions on AI use. Current academic discourse on AI predominantly focuses on "what AI can do"—translation, summarization, coding, statistics, and other single-point

functions—while paying relatively little attention to "how AI can be embedded into a complete research process." The methodological experiment in this study demonstrates that the research value of AI lies not merely in efficiency gains for individual tasks, but more importantly in serving as a "workflow infrastructure" that connects multiple stages of research, rendering the overall research process more structured and manageable.

This observation resonates with the discussion by Zhang et al. (2024) on Agentic Workflow. An effective AI workflow does not simply deploy AI to individual tasks; rather, it designs a set of linkage logic between tasks—where the output of the preceding stage serves as the input for the subsequent stage, with a human-reviewed quality gate established at each linkage point. For humanities and social sciences researchers, the value of this process-oriented thinking lies in transforming what was originally a research process highly dependent on individual experience and intuition into a structured process that can be taught, replicated, and improved.

### (2) Specific Mechanisms of the "Irreplaceability of Human Judgment"
A phenomenon that repeatedly emerged in the operational process records of Chapter IV is that AI Agents perform quite reliably at the "execution" level (data reading, statistical computation, chart generation, etc.), yet their capabilities at the "judgment" level have clear boundaries. This study identified four categories of judgment functions in which human researchers are irreplaceable:

1. Defining research questions: Deciding "what is worth studying" is a value judgment rather than a technical operation. In this study, the choice to focus on AI usage patterns in Taiwan's academic field involved understanding the needs of the academic community and identifying research gaps—judgments that cannot be delegated to AI.
2. Theoretical interpretation: Transforming data into theoretical arguments requires cross-disciplinary knowledge integration. For example, the finding that 82.9% of tasks can be independently completed by humans "implies" that users regard AI as an augmentation tool rather than a replacement tool—an interpretation that requires the researcher's deep understanding of human-computer collaboration theory and technology adoption theory.
3. Contextualized judgment: AEI data present globally standardized indicators, but the particularities of the Taiwanese context (academic evaluation systems, research resource allocation, linguistic environment, etc.) require localized contextual understanding. AI can process data but cannot comprehend the social context behind the data.
4.  Ethical reflection: Self-disclosure of research limitations, transparency requirements for AI use, and a humble attitude toward data interpretation—these practices of academic ethics depend on researchers' professional judgment and sense of moral responsibility.

The above four categories of judgment functions constitute the specific content of "human irreplaceability" in humanities and social sciences research. This finding resonates with the argument of Shavit et al. (2023)—the greater the autonomy of AI Agents, the heavier, not lighter, the responsibility of human oversight becomes.

**(3) Implications of Modular Design for Research Training**

A modular workflow holds potential educational value for graduate student training—providing a more structured training pathway than the traditional apprenticeship model, enabling graduate students to practice human-machine division-of-labor judgments progressively within each module. However, excessive reliance on structured processes may constrain creative thinking; therefore, modular workflows should be positioned as "scaffolding"—to be flexibly adjusted or even transcended as researchers mature.

# 2. Insights for Technology Management and Research Practice

**(1) Characteristics of AI Adoption in the Academic Field**

The AEI Taiwan data analysis (see Appendix A for details) reveals several structural characteristics of AI adoption in the academic field. The two broad categories of academic research and writing together account for 17.3%, and when combined with translation (8.5%), the total reaches 25.8%, indicating that AI tool penetration in the academic field has already achieved a certain scale. However, in terms of individual categories, each academic subcategory (8.9% and 8.4%) falls below software development (14.7% as the single largest category), suggesting that AI adoption in the academic field remains at a relatively early stage.

From a technology management perspective, AI adoption in the academic field exhibits characteristics of the "early majority" in Rogers' (2003) innovation diffusion theory: users have surpassed the early adopters stage but have not yet reached the stage of widespread adoption. Factors potentially driving this diffusion include: (1) efficiency demands driven by academic competition pressure; (2) translation needs for cross-language research (Taiwan's 8.5% translation usage share reflects this demand); and (3) technical empowerment needs for data analysis.

**(2) Management Implications of Collaboration Patterns**

The tripartite distribution of human-machine collaboration patterns (directive 27.9%, task-iterative 26.6%, learning-oriented 25.3%) carries management implications for research institutions' AI integration strategies. Different collaboration patterns correspond to different organizational support needs:

- Directive use requires clear prompt templates and best practice guidelines
- Iterative use requires sufficient AI usage time and quota support
- Learning-oriented use requires an organizational culture that encourages exploratory use

When formulating AI use policies, research institutions should not focus solely on the binary question of "whether to allow AI use," but should instead consider how to provide appropriate institutional support for different collaboration patterns.

**(3) Practical Value of Efficiency Augmentation**

The 82.9% human-independent completion rate, together with the high alignment of educational years (99.1%), jointly depict a usage picture of "efficiency augmentation" rather than "capability replacement." The integration of AI does not require researchers to relinquish existing competencies; rather, by having AI handle structured auxiliary tasks (literature screening, descriptive statistics, formatting, translation, etc.), cognitive resources can be concentrated on the research core that demands deep thinking.

**(4) Organizational Learning: From Research Institutions to "Agent Collaboration Centers"**

The operational experience of this study suggests that the organizational form of research institutions may require fundamental adjustments to accommodate the integration of Agentic Workflow. Traditional research institutions (such as policy think tanks) operate primarily through a hierarchical division of labor involving "senior researchers—junior researchers—research assistants": senior researchers plan research directions while junior researchers and assistants handle data collection, literature review, and report writing. The productivity of this model is constrained by personnel allocation—the number of assistants each senior researcher can direct is limited, thereby imposing physical constraints on the research scale.

The vision proposed by Hall (2026b) points toward a transformative direction for this predicament. He argues that, given that the cost of AI Agents has fallen to approximately $10 per research task and less than $5,000 in annual API fees, senior researchers can adopt a parallel operations model of "one person directing multiple Agents" to advance several research subtasks simultaneously. The concrete implementation of this model could involve senior researchers using pre-configured Agent workflows to establish real-time analytics dashboards for dynamically updated data, transforming policy research from a static output of "quarterly reports" to a dynamic analytical infrastructure of "real-time monitoring." For example, regarding Taiwan's AI adoption trends, a researcher could configure an Agent to periodically retrieve the latest AEI data, automatically generate comparative analyses and flag anomalous changes, with the researcher only needing to review the Agent-produced summaries and make judgmental decisions.

For research institutions, this transformation means that the focus of organizational learning will shift from "training researchers to master specific analytical techniques" to "training researchers to design and manage Agent workflows." An institution's core competitiveness will no longer reside solely in "how many research personnel it possesses" but rather in "whether it can construct an efficient human-machine collaboration infrastructure." Specifically, research institutions may consider the following transformation pathways: (1) establishing shared Agent template libraries that standardize commonly used research workflows (such as literature review, policy comparison, and international trend scanning) into reusable templates; (2) setting up Agent collaboration support teams to assist senior researchers in designing and optimizing workflows; and (3) redefining performance indicators, shifting from "individual output volume" to "research quality and depth of insight in human-machine collaboration."

**(5) De-skilling Risk: The Displacement of Researchers' Core Competencies**
However, the efficiency advantages of Agentic Workflow are not without costs. When Agents assume structured work such as literature synthesis, data processing, and report drafting, researchers who cease practicing these foundational skills over time may experience a "de-skilling" effect—much as the proliferation of autonomous driving technology may weaken drivers' manual control abilities.

The crux of this risk lies in the displacement of researchers' core competencies. In traditional research training, "rapidly synthesizing large volumes of literature," "proficiently operating statistical software," and "precisely formatting references" are regarded as fundamental skills for researchers. However, once these tasks are efficiently performed by Agents, researchers' core competitive advantage will inevitably shift from "rapid synthesis" to "identifying the value of literature"—that is, judging which literature merits deep reading, which arguments possess theoretical potential, and which data patterns reflect genuine phenomena rather than statistical artifacts. Similarly, the discernment of "data quality judgment" will become more irreplaceable than the operational efficiency of "rapid data organization."

This competency displacement poses particular challenges for graduate education. If graduate students rely on Agents to complete foundational research tasks from the outset of their training, they may fail to develop sufficient "ground-level cognition"—an intuitive understanding of data structures, deep memory of literature contexts, and internalized mastery of research methodology assumptions. This tacit knowledge is often precisely what is gradually internalized through repetitive manual operations.

Therefore, while promoting Agent collaboration, research institutions should deliberately maintain researchers' core judgment capabilities: during the training phase, manual operation components should be preserved so that graduate students can build a solid cognitive foundation; during the practice phase, Agents should be leveraged to enhance efficiency, but periodic "AI-free exercises" should be conducted to maintain independent research capabilities. This balance—leveraging AI's efficiency without relying on AI's judgment—represents the critical distinction in the maturation of Agentic Workflow from "tool use" to "mature collaboration."

# 3. Limitations and Risks of Integrating AI into Academic Research

**(1) AI Hallucination and Academic Integrity Risks**

The "hallucination" problem of generative AI—producing content that appears plausible but is actually incorrect—poses a serious integrity risk in academic research. In the actual operations of this study, this risk manifested primarily in two areas: literature citation (AI may generate non-existent references) and data interpretation (AI may produce inaccurate descriptions of data).

This study's workflow design addresses this risk through the "verifiability principle"—all AI outputs must undergo human review, and AI is required to annotate all sources for traceability. However, the effectiveness of this mechanism is highly dependent on the human researcher's willingness and ability to conduct reviews. If researchers relax review standards due to time pressure or trust bias, the hallucination risk will be difficult to effectively control.

**(2) Boundaries of Methodological Applicability**

The workflow proposed in this study is primarily suited for secondary data analysis-type research. Its applicability to the following research types is clearly limited:

1. Original theory construction: Theoretical development work requiring highly creative thinking—AI currently cannot effectively substitute for human researchers' conceptualization capabilities.
2. Deeply interpretive qualitative research: Research requiring researcher immersion in field contexts, such as ethnography, in-depth interviews, and discourse analysis—AI cannot provide a comparable depth of interpretation.
3. Research involving sensitive personal data: Inputting research data into AI platforms may constitute a data leakage risk, making this approach unsuitable for research involving the privacy of human subjects.

4. Field research requiring real-time interaction: AI cannot participate in data collection processes that require on-site interaction, such as interviews, observations, or experiments.

### (3) Digital Divide and Equity Considerations

The integration of AI-collaborative research may exacerbate existing digital divides within the academic field. Researchers with better access to AI tools (e.g., paid versions such as Claude Pro), stronger prompt design capabilities, and greater English language proficiency will be better positioned to effectively utilize AI collaborative workflows. This inequality may unfold along the axes of generation, discipline, language, and institutional resources, warranting attention from academic administrators and policymakers.

### (4) Long-term Impacts on the Academic Ecosystem

Should AI-collaborative research become the norm, it may produce structural impacts on the academic ecosystem: the accelerated pace of research output may intensify "publish or perish" pressures; AI-assisted writing may homogenize the style of academic papers; and dependence on AI may erode researchers' ability to independently complete foundational research tasks. These long-term impacts are currently difficult to assess empirically, but the academic community should remain vigilant while embracing AI's efficiency gains.

## 4. Limitations of This Study

This study has the following limitations, which readers should consider when interpreting the findings:

Data limitations: (1) The data are derived solely from the single platform Claude.ai; users of other AI platforms such as ChatGPT and Gemini may exhibit different behavioral patterns. (2) One week of cross-sectional data cannot capture temporal changes and trends. (3) Aggregate-level data limit the possibilities for cross-tabulation analysis. (4) Taiwan accounts for 0.77% of global usage; the representativeness of this sample size warrants cautious assessment.

Methodological limitations: (1) The methodological experiment was conducted by a single research team, and its replicability awaits independent verification by additional researchers. (2) The workflow design is based on a specific research context (secondary data analysis), and its applicability to other research types has not been tested. (3) This study itself was completed using AI collaboration, creating a self-referential methodological circularity.

Inferential limitations: (1) Descriptive analysis cannot establish causal relationships; all findings in this study are correlational descriptions rather than causal claims. (2) The classification labels in the AEI data were automatically assigned by AI, and the accuracy of the classification cannot be independently verified. (3) This study cannot distinguish between users' "actual behavior" and "optimal behavior"—the data reflect how users use AI, not how to most effectively use AI.

---

# VI. Conclusion and Recommendations for Future Research

## 1. Review of Research Questions and Core Findings

This study, positioned as a "methodological experiment," explored how AI Agents can augment the horizons of humanities and social sciences research in Taiwan, using Taiwan data from the Anthropic Economic Index as the empirical vehicle for validating methodological feasibility. Reviewing the two research questions of this study, the core findings are as follows:

Regarding Research Question 1 (workflow design and design principles), this study proposed a seven-stage collaborative workflow grounded in three principles: "task modularization," "human-machine division of labor," and "verifiability." This workflow encompasses the entire research process from research planning to reference management, with a dedicated AI Agent role assigned to each stage and a human-reviewed quality gate established at each point. This design attempts to establish a structured collaboration mechanism between AI's execution efficiency and human depth of judgment.

Regarding Research Question 2 (operability and division-of-labor boundaries), through the actual analytical operations on AEI data (Chapter IV), this study found that the workflow is operable in the context of secondary data descriptive analysis. The meta-analysis in Chapter IV revealed that AI Agents performed reliably in "direct execution" and "iterative correction" tasks such as data reading, statistical computation, chart generation, and text drafting; whereas "human-led" components—including research question definition, theoretical interpretation, contextualized judgment, and ethical reflection—constitute domains of judgment in which human researchers are irreplaceable. The boundary of human-machine division of labor is not fixed but dynamically adjusts according to the degree of task structuredness and interpretive demands.

## 2. Changes in Research Work Patterns

The findings of this study have the following implications for researchers' work patterns:

First, efficiency transformation in literature and data work. AI Agents can significantly enhance the efficiency of structured tasks such as literature searching, abstract extraction, formatting, and descriptive statistics, enabling researchers to devote more cognitive resources to critical reading and theoretical dialogue. For humanities and social sciences researchers who lack programming backgrounds, AI Agents can further serve as "technical intermediaries," lowering the entry barrier to quantitative analysis. However, the premise of efficiency transformation is that researchers possess the ability to evaluate the quality of AI outputs.

Second, from "independent completion" to "collaborative management." The AI Agent collaboration model transforms the researcher's role from "independent executor of all tasks" to "manager of human-machine collaboration"—requiring the acquisition of new skills in task delegation, prompt instruction design, and output quality review. This role transformation demands new managerial competencies but also provides possibilities for scaling research output and conducting interdisciplinary research.

Third, new challenges in quality control and ethics. AI collaboration introduces new quality control requirements (factual accuracy, argumentative logic, citation appropriateness) and ethical judgment demands (what constitutes legitimate tool-assisted use, how to transparently disclose AI usage). The answers to these questions remain unsettled and require ongoing dialogue within the academic community to build consensus.

# 3. Summary of Core Contributions

In summary, the core contributions of this study lie in three dimensions:

Methodological dimension: This study proposed a set of AI Agent collaborative research workflows that have been validated through actual operational testing, designed on the basis of three principles—"task modularization, human-machine division of labor, and verifiability"—providing a referenceable framework for humanities and social sciences researchers seeking to integrate AI tools. The value of this framework lies not in a claim of "best practice" but in providing a starting point that is open to critique, revision, and improvement.

Reflexivity dimension: Through the meta-analytical perspective adopted in Chapter IV, which comprehensively documented the operational process of AI-collaborative research, this study identified three categories of operational patterns—"direct execution," "iterative correction," and "human-led"—and revealed four specific mechanisms of "the irreplaceability of human judgment" (research question definition, theoretical interpretation, contextualized judgment, and ethical reflection), providing a first-hand case study for academic discussions on the boundaries and ethics of AI use.

Empirical dimension: Appendix A, through a complete analysis of AEI Taiwan data (N=7,729), provides a descriptive empirical foundation for AI usage behavior in Taiwan's academic field—academic-related tasks (including translation) account for 25.8%, the time savings rate is approximately 89%, and 82.9% of tasks can be completed independently by humans, delineating a usage picture of "efficiency acceleration rather than capability replacement." This provides baseline data for subsequent research on AI adoption in Taiwan's academic field and enables readers to independently evaluate the actual efficacy of the methodology.

# 4. Future Research Directions

Based on the findings and limitations of this study, the following five future research directions are proposed:

Direction 1: Cross-platform comparative research. Expanding the scope of analysis to multiple AI platforms such as ChatGPT, Gemini, and Copilot, comparing behavioral pattern differences across users of different platforms, and constructing a more comprehensive picture of AI usage in Taiwan.

Direction 2: Longitudinal tracking research. Using time-series data to track dynamic changes in AI usage behavior, observing whether AI adoption in the academic field follows the typical pattern of a diffusion curve, and whether collaboration patterns evolve with the accumulation of usage experience.

Direction 3: Disciplinary difference comparisons. Investigating disciplinary differences within the humanities and social sciences (sociology, economics, education, management, etc.), comparing AI usage patterns, collaboration preferences, and adoption barriers among researchers in different disciplines.

Direction 4: Experimental evaluation of the workflow. Designing controlled experiments to compare the research output quality, efficiency differences, and researcher experiences between AI Agent collaborative workflows and traditional research processes, evaluating the benefits of the workflow in a more rigorous manner.

Direction 5: Construction of an AI collaboration ethics framework. Developing more refined evaluation frameworks for the ethical dimensions of academic AI use, including standardized formats for AI use disclosure, criteria for determining substantive human contributions, and quality assurance mechanisms for AI-generated content.

# 5. Concluding Remarks

The metaphorical shift from "labor" to "collaboration" traces the trajectory of AI tools' evolving role in academic research. In the past, researchers completed all stages of research through individual cognitive labor; now, the integration of AI Agents enables certain stages of the research process to be accomplished through human-machine collaboration. This transformation does not imply a weakening of the researcher's role—quite the contrary. When structured execution tasks are shared with AI, the researcher's core value becomes further concentrated in the irreplaceable cognitive functions of judgment, interpretation, and reflection.

This study, adopting the humble posture of a methodological experiment, has attempted to provide an operable framework and a set of empirical data for dialogue in service of this transformation. This study fully recognizes that a single paper cannot answer all questions about AI's impact on academic research—but if it can provide humanities and social sciences researchers with a starting point for exploration and stimulate more critical discussion on AI collaboration methodology, then the purpose of this study will have been achieved.

# References

[1] Acemoglu, D., & Restrepo, P. (2019). Automation and new tasks: How technology displaces and reinstates labor. *Journal of Economic Perspectives*, 33(2), 3-30. https://doi.org/10.1257/jep.33.2.3

[2] Anthropic (2025). The Anthropic Economic Index (4th ed.). *Anthropic Research*. https://www.anthropic.com/research/economic-index-primitives

[3] Appel, R., Massenkoff, M., McCrory, P., McCain, M., Heller, R., Neylon, T., & Tamkin, A. (2026, January 15). Anthropic Economic Index report: Economic primitives. *Anthropic Research*. https://www.anthropic.com/research/anthropic-economic-index-january-2026-report

[4] Brynjolfsson, E., & McAfee, A. (2014). *The second machine age: Work, progress, and prosperity in a time of brilliant technologies*. W. W. Norton & Company. https://wwnorton.com/books/the-second-machine-age/

[5] Davidson, T., & Karell, D. (2025). Integrating generative artificial intelligence into social science research: Measurement, prompting, and simulation. *Sociological Methods & Research*. https://doi.org/10.1177/00491241251339184

[6] Dell'Acqua, F., McFowland, E., Mollick, E. R., Lifshitz-Assaf, H., Kellogg, K., Rajendran, S., Krayer, L., Candelon, F., & Lakhani, K. R. (2023). Navigating the jagged technological frontier: Field experimental evidence of the effects of AI on knowledge worker productivity and quality. *Harvard Business School Technology & Operations Mgt. Unit Working Paper*, No. 24-013. https://doi.org/10.2139/ssrn.4573321

[7] Dellermann, D., Ebel, P., Sollner, M., & Leimeister, J. M. (2019). Hybrid intelligence. *Business & Information Systems Engineering*, 61(5), 637-643. https://doi.org/10.1007/s12599-019-00595-2

[8] Eloundou, T., Manning, S., Mishkin, P., & Rock, D. (2024). GPTs are GPTs: An early look at the labor market impact potential of large language models. *Science*, 384(6702), 1306-1308. https://doi.org/10.1126/science.adj0998

[9] Gao, J., & Wang, D. (2024). Quantifying the use and potential benefits of artificial intelligence in scientific research. *Nature Human Behaviour*, 8(12), 2281-2292. https://doi.org/10.1038/s41562-024-02020-5

[10] Gao, S., Fang, A., Huang, Y., Giunchiglia, V., Noori, A., Schwarz, J. R., Ektefaie, Y., Kondic, J., & Zitnik, M. (2024). Empowering biomedical discovery with AI agents. *Cell*, 187(22), 6125-6151. https://doi.org/10.1016/j.cell.2024.09.022

[11] Gunitsky, S. (2026, January). The age of academic slop is upon us. *Hegemon* (Substack). https://hegemon.substack.com/p/the-age-of-academic-slop-is-upon

[12] Guo, T., Chen, X., Wang, Y., Chang, R., Pei, S., Chawla, N. V., Wiest, O., & Zhang, X. (2024). Large language model based multi-agents: A survey of progress and challenges. In *Proceedings of the 33rd International Joint Conference on Artificial Intelligence (IJCAI 2024)*, 8048-8057. https://doi.org/10.24963/ijcai.2024/890

[13] Hall, A. B. (2026a, January 5). Claude Code and its ilk are coming for the study of politics like a freight train [Post]. *X (formerly Twitter)*. https://x.com/ahall_research/status/2007221974947508303

[14] Hall, A. B. (2026b, January 13). The 100x research institution. *Free Systems* (Substack). https://freesystems.substack.com/p/the-100x-research-institution

[15] Hevner, A. R., March, S. T., Park, J., & Ram, S. (2004). Design science in information systems research. *MIS Quarterly*, 28(1), 75-105. https://doi.org/10.2307/25148625

[16] Karpf, D. (2026, January). What comes next, if Claude Code is as good as people say. *Substack*.

https://davekarpf.substack.com/p/what-comes-next-if-claude-code-is

[17] Mollick, E. R., & Mollick, L. (2023). Using AI to implement effective teaching strategies in classrooms: Five strategies, including prompts. *The Wharton School Research Paper*. https://doi.org/10.2139/ssrn.4391243

[18] Mondal, H., Mondal, S., & Enoch, I. T. (2023). ChatGPT in academic writing: Maximizing its benefits and minimizing the risks. *Indian Journal of Ophthalmology*, 71(12), 3600-3606. https://doi.org/10.4103/IJO.IJO_718_23

[19] Rogers, E. M. (2003). *Diffusion of innovations* (5th ed.). Free Press. https://www.simonandschuster.com/books/Diffusion-of-Innovations-5th-Edition/Everett-M-Rogers/9780743222099

[20] Shavit, Y., Agarwal, S., Brundage, M., Adler, S., O'Keefe, C., Campbell, R., ... & Robinson, D. G. (2023). Practices for governing agentic AI systems. *OpenAI Research.*

https://openai.com/research/practices-for-governing-agentic-ai-systems

[21] Straus, G., & Hall, A. B. (2026). How accurately did Claude Code replicate and extend a published political science paper? *Working Paper, Stanford University*. https://www.andrewbenjaminhall.com/Straus_Hall_Claude_Audit.pdf

[22] Thompson, D. M., Wu, J. A., Yoder, J., & Hall, A. B. (2020). Universal vote-by-mail has no impact on partisan turnout or vote share. *Proceedings of the National Academy of Sciences*, 117(25), 14052-14056. https://doi.org/10.1073/pnas.2007249117

[23] Wang, L., Ma, C., Feng, X., Zhang, Z., Yang, H., Zhang, J., ... & Wen, J. R. (2024). A survey on large language model based autonomous agents. *Frontiers of Computer Science*, 18(6), 186345. https://doi.org/10.1007/s11704-024-40231-1

[24] Yao, S., Zhao, J., Yu, D., Du, N., Shafran, I., Narasimhan, K., & Cao, Y. (2023). ReAct: Synergizing reasoning and acting in language models.

*Proceedings of the International Conference on Learning Representations (ICLR 2023)*. https://doi.org/10.48550/arXiv.2210.03629

[25] Zhang, J., Xiang, J., Yu, Z., Teng, F., Chen, X., Chen, J., Zhuge, M., Cheng, X., Hong, S., Wang, J., Zheng, B., Liu, B., Luo, Y., & Wu, C. (2024). AFlow: Automating agentic workflow generation. In *Proceedings of the 13th International Conference on Learning Representations (ICLR 2025)*. https://arxiv.org/abs/2410.10762

# Appendix A. Analysis of Taiwan's Claude.ai Usage Behavior — Based on the Anthropic Economic Index, 4th Edition

Data period: November 13, 2025 to November 20, 2025 Data scope: Taiwan (geo_id = TW), Claude AI (Free and Pro) platform Sample size: N = 7,729 conversations

---

Appendix note: This appendix is a demonstration output of human–AI collaborative analysis. The researcher provided the analytical direction, report structure, and scholarly judgment; Claude (Anthropic, claude-opus-4-6) performed data processing, chart generation, and initial draft writing. The researcher was responsible for final review, interpretive correction, and data verification. The data source is the Taiwan subset of the Anthropic Economic Index, 4th Edition publicly available raw data.

---

## Literature Background: Measuring AI's Economic Impact by "Tasks" Rather Than "Occupations"

Traditional labor economics research on the impact of automation has predominantly used "occupations" as the unit of analysis — for example, asking whether "AI will replace accountants" or "whether journalists' jobs are threatened." In recent years, however, the field has gradually shifted toward a more granular "task-level" analytical framework. Acemoglu and Restrepo (2019) argued that automation does not replace an occupation wholesale but selectively affects specific tasks within that occupation — a job comprises multiple skills and tasks, and AI may take over some of those tasks while simultaneously generating new task demands. This "task-oriented" analytical perspective captures the heterogeneous impact of AI on the labor market more precisely than the "occupation-oriented" approach.

In this context, Eloundou et al. (2024) used the pun "GPTs are GPTs" to argue that large language models, as a "General-Purpose Technology" (GPT), exert influence not confined to specific industries or occupations but permeating the concrete tasks of various types of knowledge work in an uneven yet widespread manner. They estimated that approximately 80% of the U.S. workforce has at least 10% of their work tasks affected by LLMs, while approximately 19% of the workforce has more than 50% of their tasks affected — a finding that reinforces the necessity of "task-by-task analysis."

The Anthropic Economic Index (AEI) is currently one of the very few indicators capable of conducting large-scale empirical observation at the task level. Unlike traditional occupational surveys or expert forecasts, AEI directly analyzes approximately two million real AI conversations, using the U.S. Occupational Information Network (O*NET) occupational task database and a multi-level Request taxonomy to precisely map each conversation to specific work tasks and skill requirements (Anthropic, 2025; Appel et al., 2026). This "bottom-up" empirical strategy enables researchers to observe which tasks AI is actually used for, and how much human time and education those tasks originally require — rather than relying solely on expert judgment to predict "which occupations AI might affect."

The following report applies this task-level analytical framework to examine the behavioral characteristics and usage patterns of Taiwan's Claude.ai users.

---

# Section 1. Overall Profile

This section summarizes five key performance indicators (KPIs) of Taiwan's Claude.ai usage behavior, outlining the fundamental profile of Taiwan's users.

Table A-1: Key Indicators of Taiwan's Claude.ai Usage Behavior (N=7,729)

| Indicator | Value | Description |
|---|---|---|
| Sample conversations | 7,729 | Total number of Claude.ai conversations generated by Taiwan users within one week |
| Global share | 0.77% | Proportion of Taiwan's conversations relative to the global total |

| Task success rate | 68.7% | Proportion of conversations assessed as successfully completed |
| --- | --- | --- |
| Human-can-complete-alone rate | 82.9% | Proportion of tasks assessed as completable by the user without AI assistance |
| AI autonomy median | 4.0 / 5.0 | Degree of autonomy granted to AI in conversations (1=fully instructed, 5=highly autonomous) |

With a 0.77% global share, Taiwan occupies a notable position in the Claude.ai usage landscape. Considering that Taiwan's population accounts for approximately 0.3% of the global total, this share reflects an above-average adoption density of generative AI tools among Taiwan's knowledge workers.

Among the five indicators, the most analytically significant is the human-can-complete-alone rate (82.9%). This figure indicates that the vast majority of users do not turn to AI because they "cannot complete" the task, but rather choose to have AI handle work they are originally capable of completing themselves. In other words, the usage pattern of Claude in Taiwan is predominantly one of efficiency acceleration, rather than capability substitution. Users regard AI as a tool for increasing output speed, not as a crutch for filling skill gaps.

The task success rate of 68.7% indicates that approximately one-third of conversations did not achieve their intended objectives. This figure serves as a reminder that AI collaboration is not infallible, and human review and correction remain essential. The AI autonomy median of 4.0 (out of 5.0) indicates that users tend to grant AI a high degree of autonomous space rather than micromanaging instructions step by step — a pattern consistent with the "efficiency acceleration" positioning.

# Section 2. Usage Scenario Distribution
## 2.1 Usage Scenarios

Figure A-1 presents the distribution of three usage scenarios among Taiwan's Claude users.

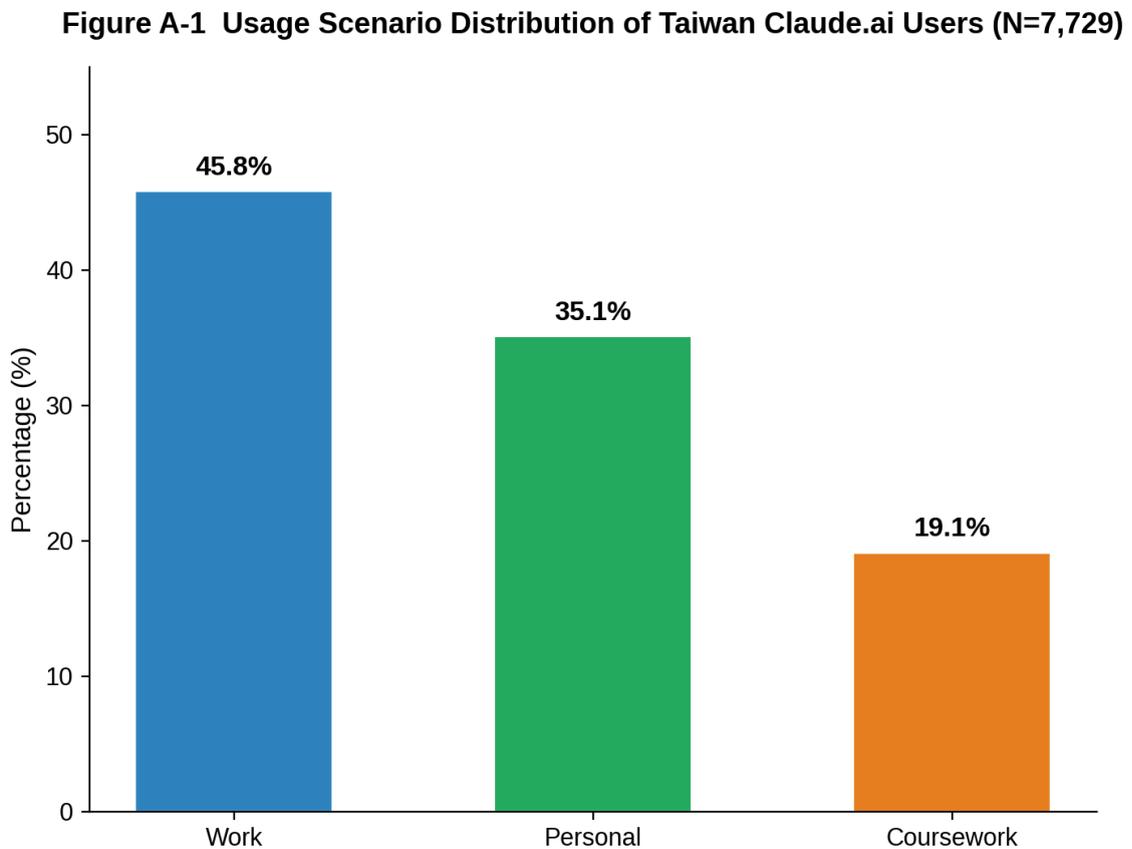

**Figure A-1  Usage Scenario Distribution of Taiwan Claude.ai Users (N=7,729)**

The work scenario accounts for the highest share (45.8%), followed by personal use (35.1%), with the academic/coursework scenario at 19.1%. The three categories exhibit a clear gradient structure: work-related use accounts for approximately half, yet personal and academic uses combined also exceed half (54.2%), indicating that Claude usage is not confined to the workplace.

Notably, academic/coursework scenarios account for nearly one-fifth. This share suggests that students and researchers constitute a non-negligible user group. For the humanities and social sciences (HSS) domain, this figure implies that a considerable proportion of users may be graduate students or academic workers using Claude for coursework reports, thesis writing, literature translation, and other academic tasks.

## 2.2 Human–AI Collaboration Modes

Figure A-2 presents the distribution of six human–AI collaboration modes.

**Figure A-2  Human–AI Collaboration Mode Distribution (N=7,729)**

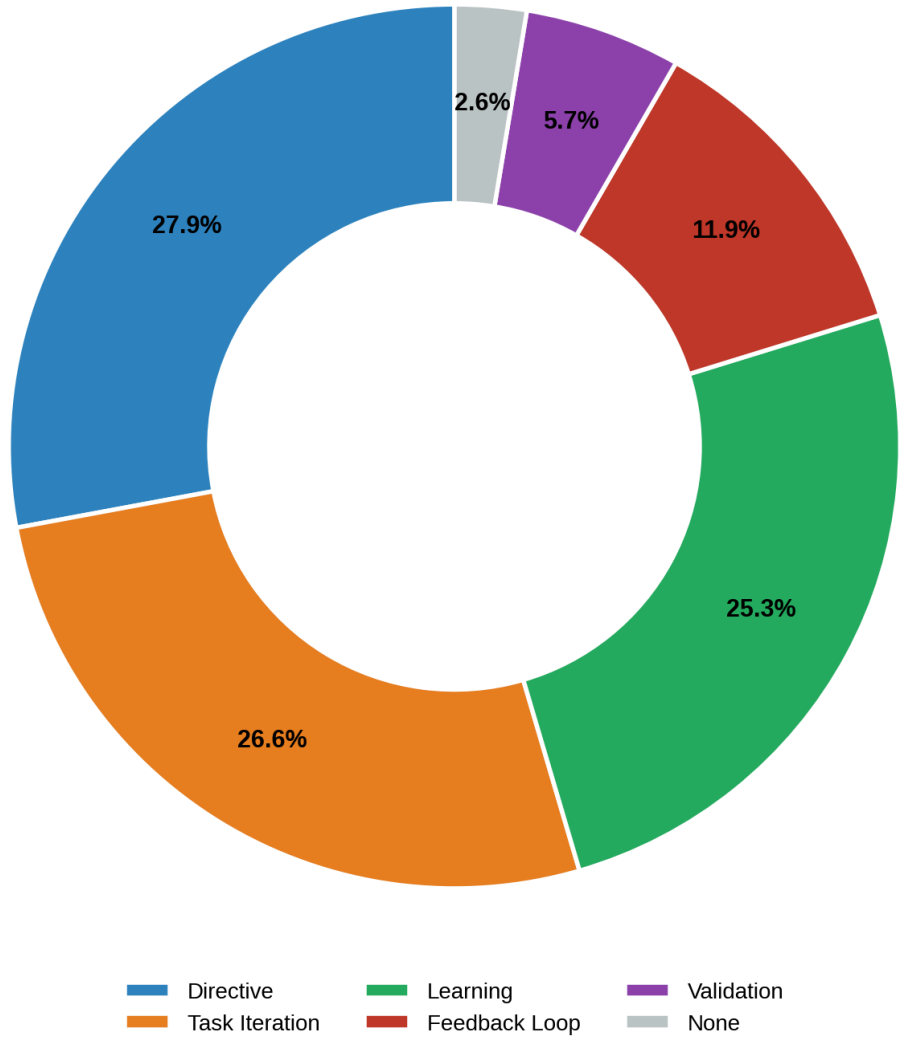

The most prominent feature of the collaboration mode distribution is the balanced configuration in which three modes share comparable proportions: directive (27.9%), task iteration (26.6%), and learning (25.3%) collectively account for 79.8%, with no single mode commanding an absolute majority. This distribution indicates that Taiwan's users do not merely operate AI through one-directional commands but engage in multi-turn conversational interaction to a considerable degree.

Feedback loop (11.9%) combined with task iteration (26.6%) means that 38.5% of conversations involve iterative revision. This proportion aligns with the nature of academic writing — paper drafts typically undergo multiple rounds of revision, and AI can play the role of a responsive collaborator in this iterative process.

Learning-mode usage at 25.3% indicates that one-quarter of users treat Claude as a learning channel. This mode is particularly important for HSS researchers who lack programming or statistical backgrounds: AI can serve as a "technical intermediary," lowering the entry barrier to cross-disciplinary methods.

---

# Section 3. Understanding This Data — O*NET and the Task Classification Method

This section provides background knowledge to help readers unfamiliar with the AEI data structure understand the classification frameworks used in subsequent analyses.

### 3.1 What Is O*NET

O*NET (Occupational Information Network) is an occupational information database established by the U.S. Department of Labor, covering approximately 1,000 occupations and standardized descriptions of their thousands of specific work tasks. Each occupational task is defined by a precise English-language statement; for example, "modify existing software to correct errors, adapt it to new hardware, or improve its performance" is a typical task for software developers.

In the AEI data, Anthropic uses an AI classifier to determine which O*NET occupational task each Claude conversation most closely corresponds to. For instance, "modify existing software to correct errors..." accounts for 5.49% of Taiwan's conversations, meaning that 5.49% of conversations are most similar in nature to this task routinely performed by software developers.

The core question that O*NET classification answers is: In the traditional occupational world, whose work does what Taiwanese people use Claude for correspond to? This classification enables us to link AI usage behavior to existing labor market structures and assess AI's potential impact on different occupational domains.

An important limitation should be noted: O*NET was established based on the U.S. labor market, and its applicability to Taiwan has gaps. Furthermore, 39.18% of Taiwan's data is labeled as not_classified (unable to be matched to any existing occupational task), and an additional 4.58% is labeled as none. Combined, nearly 44% of conversations cannot be categorized, reflecting that a considerable proportion of AI usage behavior falls outside the descriptive scope of traditional occupational tasks.

## 3.2 What Is the Request Task Classification Hierarchy

In addition to O*NET occupational task mapping, Anthropic uses a hierarchical classification system (Request taxonomy) to assign each conversation to three levels of task categories:

- Level 2 (coarsest): Broad categories with the widest coverage. For example, "translate, edit, format, summarize documents, and assist with language learning."
- Level 1 (intermediate): Groups similar Level 0 categories together. For example, "translate documents across languages."
- Level 0 (finest): Very specific task descriptions. For example, "translate documents between Asian languages and English."

Using a library classification analogy: Level 2 is like the "Literature" shelf, Level 1 is the "Fiction" section, and Level 0 is "Japanese mystery novels." The three levels together constitute a classification system from coarse to fine, enabling analysts to observe the distribution of AI usage behavior at different levels of granularity.

## 3.3 Metrics for Measuring Task Complexity

The AEI data includes six numerical metrics for quantifying the task complexity and degree of AI involvement in each conversation. Table A-2 summarizes the definitions of each metric and the actual Taiwan data.

Table A-2: Overview of Taiwan's Claude.ai Task Complexity Metrics (N=7,729)

| Metric (English name) | Description | Taiwan data | Plain-language interpretation |
|---|---|---|---|
| human_education_years | Human education years required to complete the task | $M =$ 12.49, 95% CI [12.41, 12.58] | Approximately equivalent to a university graduate level |
| ai_education_years | AI performance equivalent in education years | $M =$ 12.38, 95% CI [12.31, 12.46] | AI performance is approximately equivalent to a university graduate |
| human_only_time | Time required for a human to complete alone | $Mdn =$ 1.75 hours | Typically a one- to two-hour task |
| human_with_ai_time | Completion time with AI assistance | $Mdn =$ 12 minutes | Dramatically compressed to approximately ten minutes |
| ai_autonomy | AI autonomy level (1–5 scale) | $Mdn =$ 4.0 | Users let AI produce output with high autonomy |
| human_only_ability | Whether a human can complete the task alone | 82.9% can | Most tasks are "can do but let AI do" |

Synthesizing the above metrics, a typical portrait of Taiwan's Claude usage can be drawn: A university-level task that a human is fully capable of performing but would take one to two hours — the user chooses to let AI complete it with high autonomy in approximately ten minutes. This is the portrait of an efficiency accelerator, not capability substitution.

---

# Section 4. Task Type Rankings

Figure A-3 presents the 12 task types with the highest shares at the Request Level 2 tier (coarsest classification), excluding not_classified.

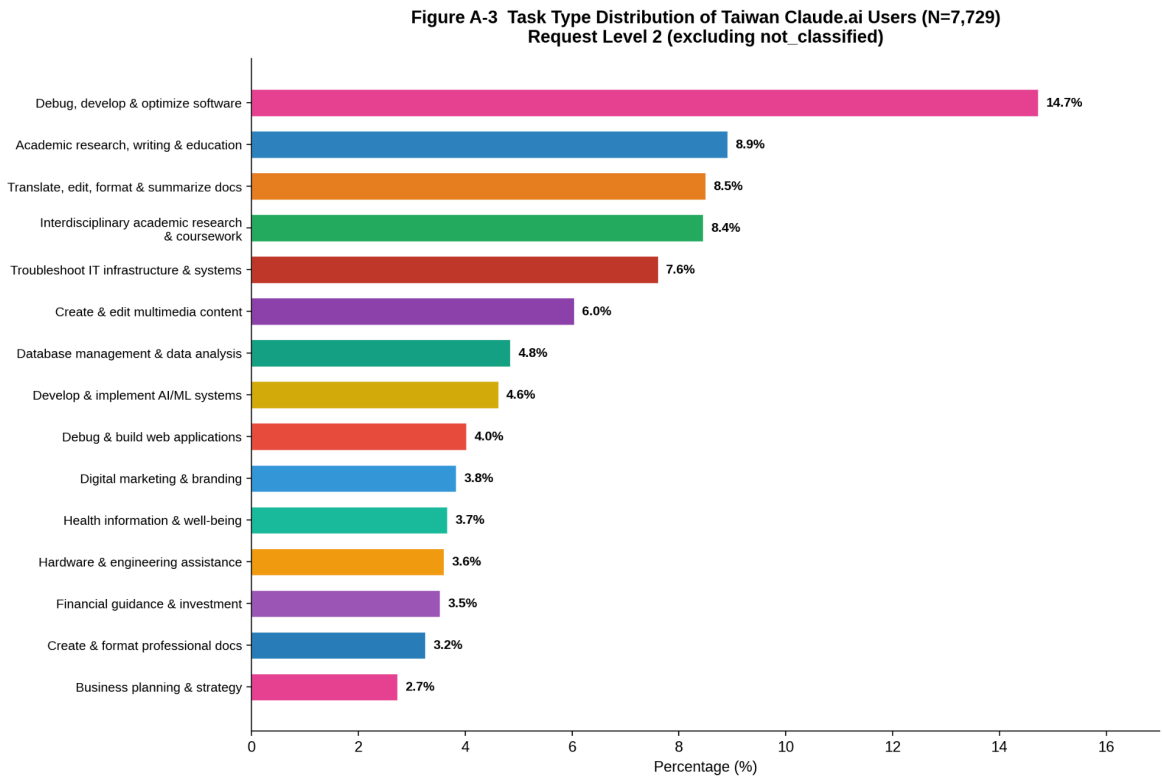

**Figure A-3  Task Type Distribution of Taiwan Claude.ai Users (N=7,729)**
**Request Level 2 (excluding not_classified)**

Software debugging, development, and optimization ranks first at 14.7%, reflecting that programming development is Claude's most prevalent use case. However, the categories ranked second through fourth are all directly related to academic or language work:

- Academic research, writing, and educational content: 8.9%
- Translation, editing, formatting, and document summarization: 8.5%
- Interdisciplinary academic research and coursework: 8.4%

The two academic research and writing categories combined account for 17.3%, translation and document processing accounts for 8.5%, and the three together total approximately 25.8% — more than one-quarter of Claude conversations are related to academic or language tasks. This share is only slightly below that of overall technical tasks (software + IT + AI/ML combined at approximately 27%), indicating that the academic domain is already the second-largest arena for AI usage in Taiwan.

For HSS researchers, the high share of translation and document processing (8.5%) is particularly noteworthy. In Taiwan's Chinese–English bilingual academic environment, translation and manuscript editing are core components of researchers' daily work, and the high ranking of this category suggests that researchers are already extensively employing AI for language-related tasks.

# Section 5. O*NET Occupational Task Analysis

## 5.1 Top 15 O*NET Tasks

Figure A-4 presents the 15 O*NET occupational tasks with the highest shares after excluding not_classified and none.

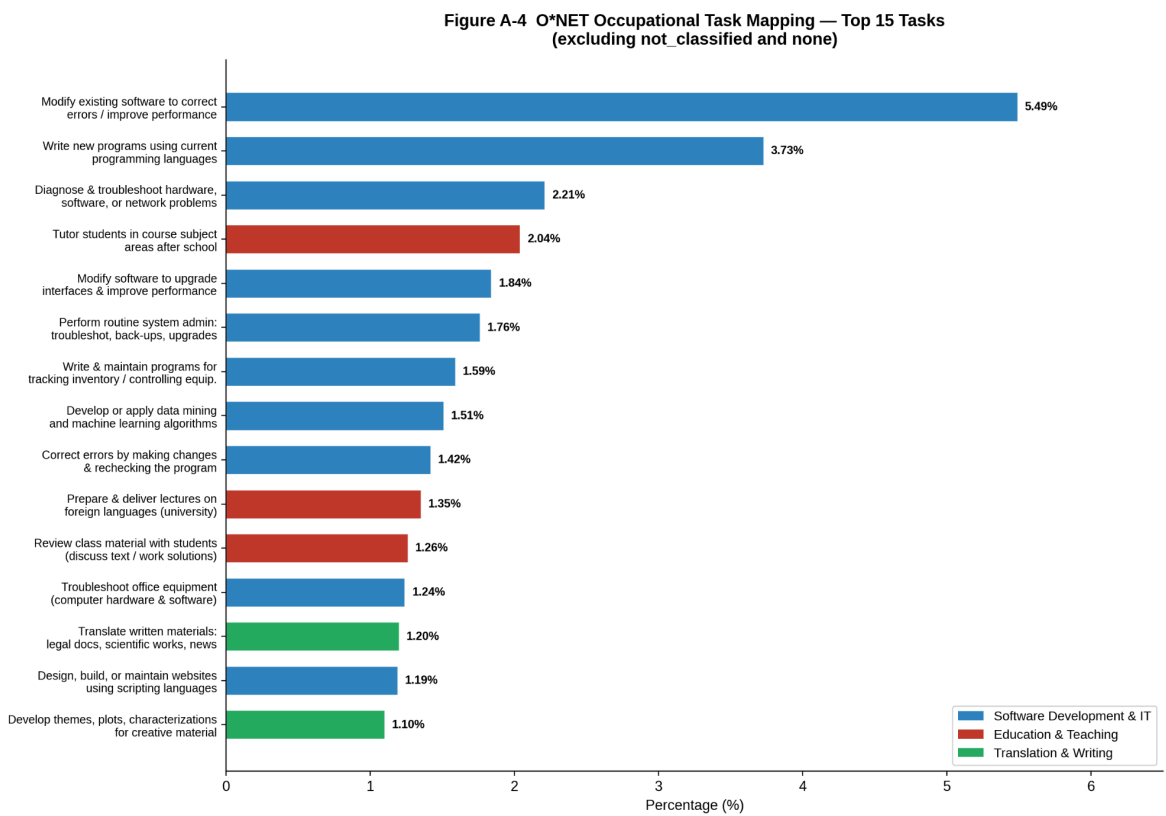

The top 15 tasks are predominantly software development-related (blue), but education and teaching (red) and translation and writing (green) also hold notable positions. Among education-related tasks, "tutor students in course subject areas after school" (2.04%), "prepare and deliver lectures on foreign languages for university students" (1.35%), and "review course materials with students" (1.26%) together account for 4.65%, demonstrating the significant presence of teaching activities in AI usage.

**5.2 Five Major Domain Clusters**

Table A-3 groups O*NET tasks into five major domains, presenting the representative tasks, combined share, and affected academic disciplines in Taiwan for each domain.

Table A-3: O*NET Task Clustering into Five Major Domains

| Domain | Representative tasks | Combined share | Affected academic disciplines in Taiwan |
|---|---|---|---|
| Software development & IT | Modify software to fix errors, write programs, system administration | ~31.7% | Computer science, information management |
| Education & teaching | After-school tutoring, teaching foreign languages, reviewing coursework | ~13.5% | Education, foreign language teaching, pedagogy across disciplines |
| Translation, writing & editing | Translate documents, develop creative materials | ~7.5% | Foreign languages and literatures, translation studies, Chinese literature, communication studies |
| Counseling & guidance | Provide social services, psychology-related counseling conversations | ~1.3% | Social work, counseling psychology |
| Academic research | Literature analysis, data processing, research methods | ~0.5% | Cross-disciplinary academic research |

Education and teaching (13.5%) plus translation, writing, and editing (7.5%) combined total approximately 21%. Notably, these are precisely the core work skills of HSS researchers — preparing and delivering lectures, translating literature, writing papers, and editing manuscripts. Although the O*NET classification is based on the U.S. occupational system, the results clearly point to the conclusion that AI is intervening in the most routine work content of HSS researchers.

The share of academic research per se (~0.5%) appears extremely low, but this reflects a characteristic of the O*NET classification — O*NET does not have a dedicated "academic researcher" task category. Academic research work is decomposed into specific tasks such as "teaching," "translation," "writing," and "data analysis," which are then distributed across multiple categories. Therefore, the actual AI usage volume for academic research far exceeds the 0.5% directly classified under O*NET.

---

# Section 6. Education Level and Time Efficiency

### 6.1 Education Years Comparison

Figure A-5 compares human education years with AI-equivalent education years.

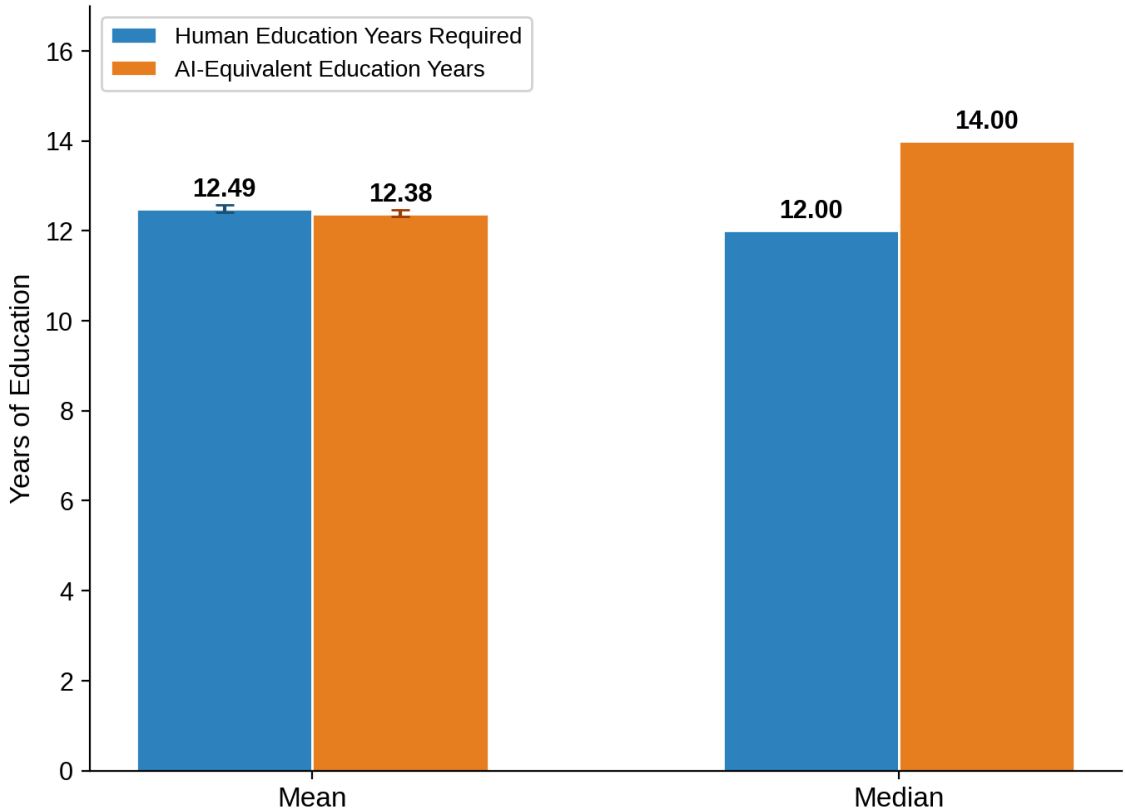

**Figure A-5  Education Years Comparison: Human vs. AI (N=7,729)
(Error bars show 95% CI for means)**

Five key data points:

| Indicator | Value |
|---|---|
| Human education years required for the task | *M* = 12.49 years, 95% CI [12.41, 12.58] |
| AI-equivalent education years | *M* = 12.38 years, 95% CI [12.31, 12.46] |
| Human education years median | *Mdn* = 12.00 years |
| AI-equivalent education years median | *Mdn* = 14.00 years |
| Mean difference | 0.11 years (CIs overlap; difference not significant) |

The mean human education years of 12.49 and the mean AI-equivalent education years of 12.38 are highly similar, with their 95% confidence intervals overlapping, indicating that the difference is not statistically significant. This result implies that the task complexity users assign to AI is roughly equivalent to the level their own education equips them to handle. AI performance is assessed as comparable to that of a university graduate.

In terms of medians, the AI-equivalent education years (14.0 years) are slightly higher than the human education years (12.0 years), suggesting that for "typical" tasks, AI performance is even slightly superior to the education level the task itself requires.

## 6.2 Time Efficiency Comparison

Figure A-6 compares human-only completion time with AI-assisted time.

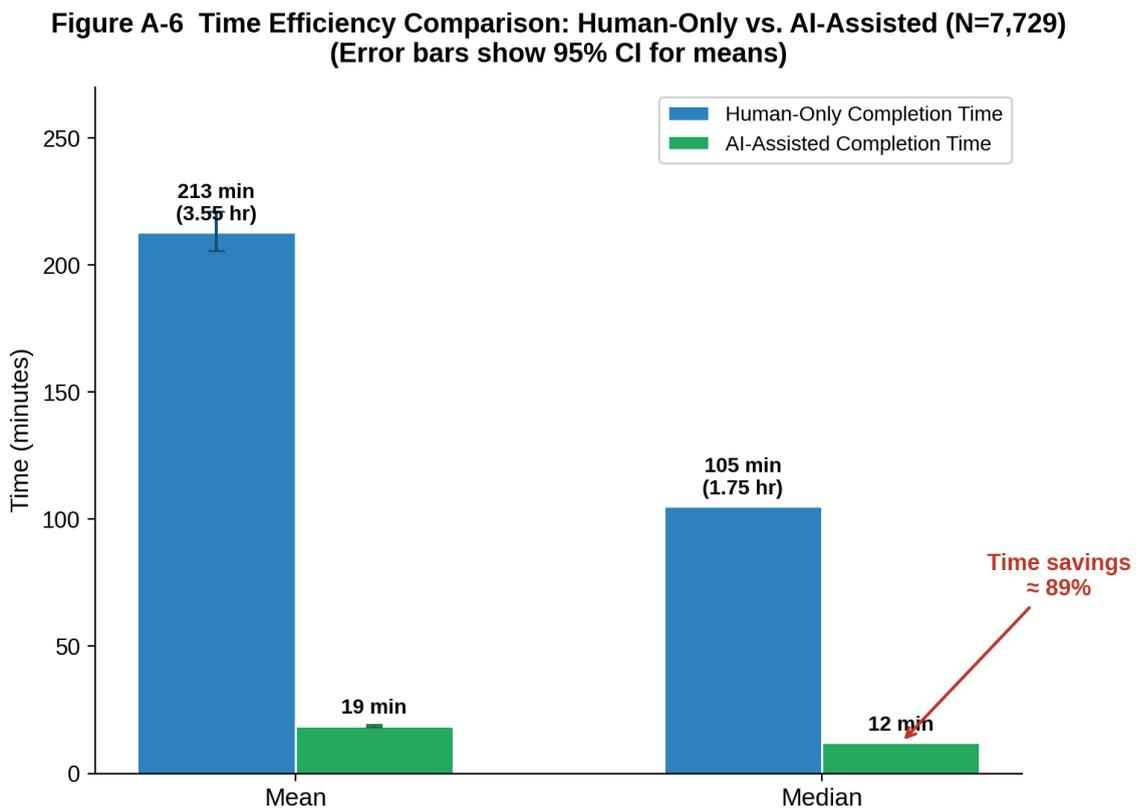

**Figure A-6  Time Efficiency Comparison: Human-Only vs. AI-Assisted (N=7,729) (Error bars show 95% CI for means)**

| Indicator | Value |
|---|---|
| Human-only time (mean) | *M* = 3.55 hours (212.8 minutes), 95% CI [3.43, 3.68] hr |
| AI-assisted time (mean) | *M* = 18.7 minutes, 95% CI [18.2, 19.2] min |

| Human-only time (median) | *Mdn* = 1.75 hours (105 minutes) |
|---|---|
| AI-assisted time (median) | *Mdn* = 12.0 minutes |
| Median time savings rate | Approximately 89% (105 minutes → 12 minutes) |

Based on median values, human-only completion requires 1.75 hours, while AI-assisted completion requires only 12 minutes, yielding a time savings rate of approximately 89%. Even using the more conservative median estimate, AI still compresses the typical task completion time to approximately one-ninth of the original.

For HSS researchers, this efficiency gain is particularly impactful in scenarios such as translation and proofreading (a typical 2–3 hour English paper translation can be compressed to 15–20 minutes) and lesson plan preparation (organizing teaching materials can be shortened from several hours to half an hour). However, it should be noted that this comparison is between "estimated human-only completion time" and "actual AI-assisted interaction time" — the former is an estimated value while the latter is a measured value, and their measurement bases are not entirely equivalent.

# Section 7. Inferences Regarding the Impact on Academic Researchers in the Humanities and Social Sciences

Figure A-7 presents the 14 Request L1 task categories directly relevant to HSS academic work.

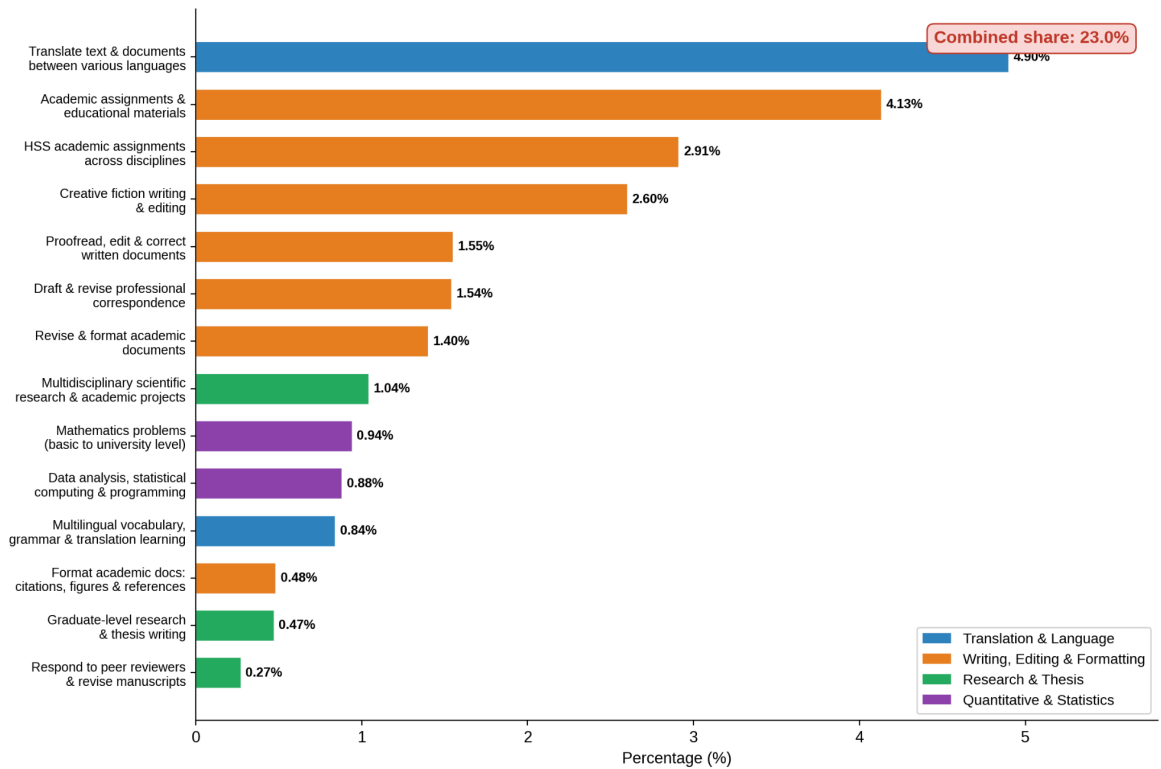

**Figure A-7 HSS-Related Academic Task Distribution (Request L1)**
**14 categories directly relevant to humanities & social science research**

Combined share: 23.0%

Legend:
- Translation & Language
- Writing, Editing & Formatting
- Research & Thesis
- Quantitative & Statistics

Percentage (%)

The 14 HSS-related tasks combined account for 23.0% of all conversations, approaching one-quarter. The following discussion infers AI's impact on academic researchers in Taiwan's HSS domain from five perspectives.

## Inference 1: Translation and Multilingual Processing Is the Most Prevalent AI-Assisted Scenario

Translation of documents and multilingual translation accounts for 4.90% at the L1 level, the highest share among HSS-related tasks. At the L0 level, "translate documents between Asian languages and English" accounts for 2.91%, and multilingual vocabulary and grammar learning accounts for an additional 0.84%. Cross-linguistic work is the most intuitive AI entry point for Taiwan's HSS researchers.

This phenomenon reflects a structural characteristic of Taiwan's academic environment: researchers need to extensively read English-language literature and write reports in Chinese (or vice versa), making translation a task that occupies a large proportion of daily work yet carries relatively low added value. AI is dramatically reducing this language barrier; in the long run, it may reshape the weight of "English proficiency" in academic competency assessment — when translation is no longer a bottleneck, the scarcity value of language ability will decline, and the academic evaluation system may need to be recalibrated.

**Inference 2: "Full-Pipeline AI-Assisted" Academic Writing Is Taking Shape**

From the data, each stage of the academic writing process can be identified as having a corresponding AI-assisted task:

| Writing stage | Corresponding L1 task | Share |
|---|---|---|
| Literature translation | Translate documents and multilingual translation | 4.90% |
| Draft writing | Academic assignments and teaching material creation; HSS academic assignments | 4.13% + 2.91% |
| Revision and formatting | Academic document revision and formatting | 1.40% |
| Citation management | Academic citation formatting | 0.48% |
| Reviewer response | Respond to peer review and submission revisions | 0.27% |

AI is no longer merely a "writing assistant" but a full-pipeline collaborative tool spanning from literature translation, draft writing, revision and formatting, and citation management to responding to reviewer comments. This trend implies that every stage of academic writing has entered the range of AI intervention, and "AI-assisted writing" is evolving from a single-point tool into a systematic workflow.

**Inference 3: The Share of Academic/Coursework Scenarios Suggests High Graduate Student Dependence**

Academic/coursework scenarios account for 19.1% of overall usage, HSS academic assignments account for 2.91%, academic document revision accounts for 1.40%, and the learning collaboration mode accounts for 25.3%. Taken together, a considerable proportion of users treat Claude as a learning tool or knowledge advisor rather than a mere surrogate for labor.

Graduate students are likely the primary drivers of this pattern. Graduate students simultaneously assume the dual roles of "learner" and "researcher," and their AI usage patterns combine learning mode (understanding new concepts, exploring methodologies) with task iteration mode (drafting and revising papers). The 25.3% share of learning-mode collaboration supports this inference: users treat AI to a considerable degree as "an advisor available for questions at any time."

## Inference 4: Time Savings Have a Fundamental Impact on Research Productivity

The median time savings rate is approximately 89% (1.75 hours → 12 minutes). If this efficiency gain is projected onto specific academic work scenarios: translating and proofreading an English paper can be shortened from 3 hours to 20 minutes; if 20 papers require translation processing per semester, the cumulative time savings would amount to approximately 50 hours.

This productivity enhancement has fundamental implications: researchers can invest the time saved into more creative work — deep thinking, theory building, fieldwork — rather than expending it on structured clerical processing. However, time savings also bring the academic ethics challenge of blurring the boundary between "AI-assisted" and "AI-ghostwritten." When AI can produce university-level text in minutes (ai_education_years median = 14.0 years), defining the boundaries of reasonable use will become an issue that academic institutions cannot avoid.

## Inference 5: Humans Can Still Complete Most Tasks Independently; AI Plays an Accelerating Role

82.9% of tasks are assessed as completable by humans alone, yet the AI autonomy median is as high as 4.0/5.0. This combination depicts a new normal of "can do but let AI do": users possess the capability to complete tasks but choose to delegate execution to AI to save time and cognitive load.

For the HSS academic community, this data challenges the extreme narrative that "AI will replace humanities scholars" while simultaneously refuting the conservative stance that "humanities scholars do not need AI." A more accurate description is: AI is redistributing the cognitive resource allocation of HSS

researchers — delegating structured, repetitive work (translation, formatting, literature organization) to AI for accelerated processing, thereby freeing up time and attention for core work that requires human judgment (theory building, contextualized interpretation, ethical reflection).

**Impact Summary**

Table A-4: Five Major Dimensions of AI's Impact on Taiwan's HSS Academic Researchers

| Dimension | Data basis | Specific impact on researchers | Risks and challenges |
|---|---|---|---|
| Lowered language barriers | Translation ranks second-highest at L1 (4.90%); Asian language translation (2.91%) | The threshold for reading and writing English papers is substantially lowered; non-native English-speaking researchers gain increased opportunities for international publication | Accuracy of specialized terminology in translations still requires human review; over-reliance may weaken researchers' own language proficiency |

| Enhanced writing efficiency | The entire academic writing pipeline has corresponding AI tasks (draft → revision → citation → reviewer response) | Paper production cycles are shortened; multi-round revision efficiency is improved | The boundary between "AI-assisted" and "AI-ghostwritten" is blurred; standards for determining academic originality need to be redefined |
|---|---|---|---|
| Teaching model transformation | Education and teaching accounts for 13.5% of O*NET; academic/coursework scenarios account for 19.1% | Instructors can use AI to create teaching materials and design assessments; students use AI as an after-class learning aid | The validity of traditional assessment methods (in-class presentations, term papers) is challenged |
| Expansion of research methods | Data analysis (0.88%), mathematics and statistics (0.94%), learning-mode collaboration (25.3%) | HSS researchers lacking quantitative backgrounds can leverage AI to overcome barriers in statistical analysis and programming | AI-assisted statistical analysis may lead to "black box" method application, where researchers do not necessarily understand the assumptions and limitations of the methods they employ |

| New academic ethics issues | 82.9% of tasks are completable by humans alone yet delegated to AI; AI autonomy median of 4.0/5.0 | "Can do but let AI do" becomes the new normal, necessitating clear AI usage disclosure norms | Academic integrity standards have not yet kept pace with technological development; perceptions of reasonable AI use may diverge between instructors and students |
|---|---|---|---|

# Section 8. Conclusions and Recommendations

Taiwan's Claude.ai usage data depicts a usage portrait centered on "efficiency acceleration": 82.9% of tasks could have been completed by humans alone, yet with AI assistance, the median completion time was compressed from 1.75 hours to 12 minutes. Academic-related tasks (including translation) account for 25.8%, indicating that AI penetration in Taiwan's academic domain has reached a considerable scale. For HSS researchers, translation and proofreading, academic writing, and lesson preparation are the primary AI usage scenarios, and the trend of "full-pipeline AI-assisted writing" is transitioning from possibility to reality.

Based on the above analysis, the following five recommendations are proposed:

1. Establish clear AI usage guidelines: Academic institutions should formulate specific and actionable AI usage norms that distinguish between "reasonable assistance" (e.g., translation proofreading, formatting) and "academic dishonesty" (e.g., undisclosed AI ghostwriting), rather than responding with blanket prohibitions.
2. Incorporate AI literacy into research methodology courses: Rather than outright prohibition, research methodology courses should systematically teach the reasonable use, limitations, and risks of AI tools, cultivating graduate students' capacity for "critical use of AI."
3. Reassess the role of language proficiency in academic evaluation: As AI translation quality continues to improve, the weight of "English writing ability" as an indicator of academic competence may need adjustment. Academic evaluation should focus more on research design, analytical depth, and theoretical originality rather than the fluency of linguistic expression.

4. Attend to AI's potential erosion of deep thinking and original insight: AI's efficiency advantages may incline researchers toward rapid output rather than deep reflection. The academic community should consciously preserve the value of "slow thinking," ensuring that AI accelerates routine work rather than supplanting the knowledge creation that requires time for maturation.

5. Leverage efficiency gains by investing saved time in research innovation: The 89% time savings rate signifies an enormous release of cognitive resources. Researchers should strategically deploy this efficiency dividend — investing the saved time in fieldwork, cross-disciplinary dialogue, theoretical reflection, and other high-value work that AI cannot yet replace, rather than merely using AI to accelerate existing work patterns.

---

Data Source Declaration

The data used in this report's analysis comes from the Anthropic Economic Index (AEI), 4th Edition publicly available raw data.

Anthropic. (2025). *The Anthropic Economic Index (4th ed.)*. Retrieved from https://www.anthropic.com/research/economic-index-primitives

Dataset: `aei_raw_claude_ai_2025-11-13_to_2025-11-20.csv` (Taiwan subset) Data download: https://huggingface.co/datasets/Anthropic/EconomicIndex Data period: November 13, 2025 to November 20, 2025 Platform: Claude AI (Free and Pro) Geographic scope: Taiwan (geo_id = TW) Sample size: N = 7,729 conversations

All data in this report were directly extracted from the above raw data without any statistical inference or model estimation. Charts were generated using Python (matplotlib 3.10); source code is available at

`scripts/generate_all_charts.py`.

---

Appendix References

Acemoglu, D., & Restrepo, P. (2019). Automation and new tasks: How technology displaces and reinstates labor. *Journal of Economic Perspectives*, 33(2), 3-30. https://doi.org/10.1257/jep.33.2.3

Eloundou, T., Manning, S., Mishkin, P., & Rock, D. (2024). GPTs are GPTs: An early look at the labor market impact potential of large language models. *Science*, 384(6702), 1306-1308. https://doi.org/10.1126/science.adj0998

# 從勞動到協作：利用 **AI Agent** 擴增台灣人文社會科學研究視野的方法論實驗

## From Labor to Collaboration: A Methodological Experiment Using AI Agents to Augment Research Perspectives in Taiwan's Humanities and Social Sciences

黃意植 博士（主要聯絡人）
Dr. Yi-Chih HUANG (Corresponding Author)
副研究員／ Associate Researcher
財團法人國家實驗研究院 科技政策研究與資訊中心
National Applied Research Laboratories
Science & Technology Policy Research and Information Center
聯絡地址：106 台北市大安區和平東路二段 106 號一樓
Contact Address: 1, 14-15F, No. 106, Heping E. Rd., Sec. 2, Taipei 10636, Taiwan (R.O.C.)
聯絡電話：+886-2-2737-7427
TEL: +886-2-2737-7427
E-mail: yichuang@niar.org.tw

---

## 摘要

生成式人工智慧正在重塑知識工作型態，然而現有研究多聚焦於軟體工程與自然科學，對人文社會科學的方法論啟示仍缺乏系統性探討。本研究以「方法論實驗」為核心定位，提出一套基於 AI Agent 的人文社會科學研究代理式工作流（Agentic Workflow），並以 Anthropic 經濟指數（Anthropic Economic Index, AEI）台灣地區 Claude.ai 使用數據（N=7,729，2025 年 11 月）作為驗證該方法論可行性的實證載體。

本研究包含兩個層次：主體層次為方法論框架的設計與驗證——提出基於「任務模組化、人機分工、可驗證性」三項原則的七階段模組化工作流程，每階段明確界定人機分工，人類負責研究判斷與倫理決策，AI 負責資訊檢索與文本生成；載體層次為 AEI 台灣數據的實證分析——作為方法論的操作示範，展示工作流程在二手資料研究中的實際運作過程與產出品質（詳見附錄 A）。

本研究貢獻在於：提出可供人文社會科學研究者參照的 AI 協作方法論框架；透過操作過程的反身性記錄，歸納出「直接執行型」「迭代修正型」「人類主導型」三類人機協作操作模式，揭示人類研究者在研究問題界定、理論詮釋、脈絡化判斷與倫理反思等環節之不可替代性。本研究亦坦承單一平台、橫斷面資料及 AI 可靠性等侷限，並提出未來研究方向。

關鍵詞：生成式人工智慧、AI Agent、人機協作、研究方法論、代理式工作流（Agentic Workflow）、人文社會科學

---

# Abstract

Generative AI is reshaping knowledge work, yet existing research focuses predominantly on software engineering and the natural sciences, with limited methodological exploration for the humanities and social sciences. Positioned as a "methodological experiment," this study proposes an AI Agent-based collaborative research workflow (Agentic Workflow) for humanities and social science research. Taiwan's Claude.ai usage data (N = 7,729 conversations, November 2025) from the Anthropic Economic Index (AEI) serves as the empirical vehicle for validating the feasibility of this methodology.

This study operates on two levels: the primary level is the design and validation of a methodological framework — a seven-stage modular workflow grounded in three principles: task modularization, human-AI division of labor, and verifiability, with each stage delineating clear roles for human researchers (research judgment and ethical decisions) and AI Agents (information retrieval and text generation); the secondary level is the empirical analysis of AEI Taiwan data — serving as an operational demonstration of the workflow's application to secondary data research, showcasing both the process and output quality (see Appendix A).

This study contributes by proposing a replicable AI collaboration framework for humanities and social science researchers, and identifying three operational modes of human-AI collaboration — direct execution, iterative refinement, and human-led — through reflexive documentation of the operational process. This taxonomy reveals the irreplaceability of human judgment in research question formulation, theoretical interpretation, contextualized reasoning, and ethical reflection. Limitations including single-platform data, cross-sectional design, and AI reliability risks are acknowledged.

# 壹、緒論

## 一、研究背景

自 2022 年底大型語言模型（Large Language Models, LLMs）進入公眾視野以來，生成式人工智慧（Generative AI）對知識工作的衝擊已從技術討論擴展至社會科學的核心議題。根據 Anthropic 發布的經濟指數報告（Anthropic Economic Index, AEI），全球 Claude.ai 使用者的任務涵蓋軟體開發、學術研究、內容創作、商業分析等多種知識密集型工作（Anthropic, 2025; Appel et al., 2026），顯示 AI 工具的使用已非特定技術社群的專利，而是正在向各類知識工作者擴散。

在學術研究領域，生成式 AI 的應用同樣快速增長。Davidson 與 Karell（2025）指出，生成式 AI 可作為社會科學研究的「測量工具」（measurement）、「提示工具」（prompting）與「模擬工具」（simulation），為傳統研究方法提供互補性的分析能力。然而，多數研究聚焦於 AI 作為單一工具的功能評估，較少從「研究流程」（research workflow）的角度探討 AI 如何系統性地融入學術研究的全過程。

更重要的是，現有 AI 輔助研究的討論主要源自自然科學與工程學科的經驗（Gao & Wang, 2024），對人文社會科學領域的適用性探討明顯不足。人文社會科學研究具有高度詮釋性、理論建構導向以及情境敏感性等特質，這些特質使得 AI 的導入面臨不同於自然科學的方法論挑戰。如何在保持人文社會科學研究品質的前提下，善用 AI 的資訊處理與文本生成能力，是一個亟待探索的課題。

## 二、研究問題

在此背景下，近年來「AI Agent」概念的興起提供了新的思考方向。不同於傳統的 AI 工具使用（單次提問—回答模式），AI Agent 強調任務導向的自主執行能力，能夠根據研究者的指令，在預設的工作流程（workflow）中連續完成多步驟任務（Guo et al., 2024; Zhang et al., 2024）。這種「代理式工作流」（Agentic Workflow）的概念，恰好呼應了學術研究流程中「任務可拆解、步驟可序列化」的特性，為人文社會科學研究者提供了一種新的人機協作模式。

然而，目前缺乏一套針對人文社會科學研究情境的 AI Agent 協作方法論框架。既有的 AI 工作流程設計多以軟體開發為場景（如 AutoGen、CrewAI 等多代理框架），未能充分考量人文社會科學研究的特殊需求：文獻回顧的批判性、理論對話的深度、數據詮釋的多元性，以及研究倫理的敏感性。

基於上述研究背景，本研究提出以下兩個研究問題：

研究問題一：如何設計一套適用於人文社會科學二手資料研究的 AI Agent 協作工作流程？其設計原則為何？

研究問題二：在實際操作中，此工作流程的可操作性如何？人類研究者與 AI Agent 的分工邊界如何界定？

## 三、研究目的與貢獻

本研究採取「方法論實驗」（methodological experiment）的定位，目的並非驗證特定假設，而是透過實際操作一套 AI Agent 協作研究流程，檢視其可行性、揭示其限制，並為後續研究提供可參照的方法論框架。

須特別說明的是，本研究具有雙重層次結構：

- 主體層次——方法論實驗：本研究的核心關懷在於方法論框架本身的設計、操作與反思。第參章提出的七階段 AI Agent 協作工作流程，及其背後的「任務模組化、人機分工、可驗證性」三項設計原則，構成本研究的主要貢獻。第肆章透過實際操作過程的描述，驗證此工作流程的可操作性，並記錄人類研究者在各關鍵節點的介入判斷。
- 載體層次——AEI 數據分析：Anthropic 經濟指數台灣數據的分析，係作為驗證上述方法論可行性的實證載體。此分析本身並非本研究的最終目的，而是方法論實驗的操作素材與產出物。完整的數據分析報告以附錄形式呈現（附錄 A），供讀者檢視工作流程的實際產出品質。

具體而言，本研究的預期貢獻包括：

第一，方法論貢獻：提出一套七階段的 AI Agent 協作研究工作流程，涵蓋從研究規劃到參考文獻整理的完整研究歷程，並以「任務模組化」「人機分工」「可驗證性」三項原則作為設計基礎。此框架可供人文社會科學研究者在導入 AI 工具時參考。

第二，反身性貢獻：透過完整記錄 AI 協作研究的操作過程（包括各階段的人類介入節點、Prompt 設計的迭代歷程、AI 產出的修訂過程），揭示人機協作中「人類判斷不可替代性」的具體環節，為學術界討論 AI 使用倫理與透明度提供第一手案例。

## 四、研究方法概述

本研究採用混合方法取向。在方法論層面，以設計科學（Design Science）的精神建構 AI Agent 協作工作流程，並透過實際操作進行迭代修正。在實證層面，採用描述性分析法（Descriptive Analysis）處理 AEI 台灣數據，以次數分配、百分比、集中趨勢與離散趨勢等統計量描繪台灣使用者的 AI 使用行為輪廓。

本研究使用的主要數據來源為 Anthropic 公開發布的經濟指數資料集，聚焦於台灣地區（geo_id: TW）在 2025 年 11 月 13 日至 20 日期間的 Claude.ai 使用記錄。數據涵蓋任務類型、協作模式、AI 自主性、任務成功率、使用情境等多面向（facets）的分析維度，共計 7,729 筆對話紀錄。

## 五、論文結構

本文以下各章的安排如下：第貳章進行文獻檢閱，回顧生成式 AI 在學術研究中的應用、人機協作理論以及 AI Agent 工作流程的相關研究；第參章說明研究設計與方法論，包括工作流程的設計原則、數據來源與分析方法；第肆章以 AEI 台灣數據為實證載體，以後設分析的視角描述工作流程七階段的實際操作過程，歸納三類人機協作操作模式；第伍章進行討論，分析操作經驗的理論意涵與實務啟示；第陸章總結研究貢獻，並提出未來研究方向。此外，附錄 A 收錄由 AI Agent 協作產出的完整分析報告，供讀者檢視方法論的實際產出品質。

## 貳、文獻檢閱

本章回顧三個相互關聯的文獻領域：生成式 AI 在學術研究流程中的角色演進、人機協作的理論框架，以及 AI Agent 與工作流程模組化的技術發展。透過文獻的系統性整理，本章旨在建立本研究的理論定位，並識別現有研究的缺口。

## 一、生成式 AI 與學術研究流程之演進

生成式 AI 對學術研究的影響，經歷了從「輔助工具」到「協作夥伴」的認知轉變。早期研究主要將 AI 定位為文本處理的效率工具，例如文獻摘要、語法校正與翻譯輔助等功能性應用（Mondal et al., 2023）。這一階段的 AI 使用模式屬於「單點介入」——研究者在特定環節使用 AI 完成特定任務，AI 與研究流程的整合程度有限。

隨著大型語言模型能力的提升，學者開始探索 AI 在研究流程中更深層次的參與。Davidson 與 Karell（2025）提出了三種整合模式：測量（measurement）、提示（prompting）與模擬（simulation），為社會科學研究者提供了系統性的 AI 導入框架。測量模式將 AI 視為編碼工具，用於大規模文本資料的分類與標註；提示模式利用

AI 的語言生成能力，探索理論假設的可能推論；模擬模式則將 AI 作為社會行為的模擬器，產生可供分析的合成數據。

在管理學與組織研究領域，Dell'Acqua 等人（2023）的實驗研究發現，使用 AI 的管理顧問在特定任務上的表現顯著優於未使用 AI 者，但在超出 AI 能力邊界的任務上反而表現下降——研究者稱之為「鋸齒狀前沿」（jagged frontier）效應。此發現對學術研究具有重要啟示：AI 的導入並非全面性的能力提升，而是在不同任務類型上呈現差異化的效益，研究者需要具備辨識 AI 能力邊界的判斷力。

Mollick 與 Mollick（2023）進一步指出，在教育場景中，AI 的有效使用高度依賴使用者的 prompt 設計能力與任務拆解策略。這一觀察對學術研究同樣適用：研究者能否有效運用 AI，取決於其將複雜研究問題拆解為可操作子任務的能力，而非僅是 AI 工具本身的技術能力。

最具示範意義的實踐案例來自 Stanford 政治學者 Hall。Hall（2026a）使用 AI 編程代理（Claude Code）在約一小時內完整複製並擴展了一篇已發表的政治學實證論文——Thompson 等人（2020）關於全面郵寄投票（universal vote-by-mail）對投票率與選舉結果影響的研究。經由獨立審查，Straus 與 Hall（2026）發現 AI 的複製結果高度準確：12 個迴歸係數全數精確複製至小數點後三位，選舉數據的蒐集與原始數據的相關係數超過 0.999。然而，審查亦揭示了 AI 的能力邊界：在 30 個加州郡的處理狀態編碼中，AI 誤判了 1 個郡的處理時間；在嘗試超越原始論文範疇的新分析時，AI 的表現明顯下降——並非產生「幻覺」，而是偏離了原始 prompt 的意圖，產出了設計不夠嚴謹的分析。

Hall 的實驗為本研究提供了兩項關鍵啟示：第一，AI Agent 在「結構明確、邊界清楚」的實證研究任務中已展現出驚人的執行能力，能夠大幅壓縮數據蒐集與分析的時間成本；第二，AI 的能力邊界恰恰出現在需要研究判斷的環節——當任務從「複製既有分析」轉向「設計新分析」時，人類研究者的引導與監督不可或缺。此一觀察為本研究主張的「人機分工」原則提供了直接的實證支持。

## 二、從工具到協作：人機協作的理論框架

人機協作（Human-AI Collaboration）理論為理解 AI 在研究流程中的角色提供了分析框架。既有文獻呈現出三種主要觀點的競爭與互補。

工具論觀點（Tool Perspective）將 AI 視為研究者的延伸工具，強調人類的主導地位。在此觀點下，AI 是「被使用」的客體，其價值在於提升人類研究者的工作效率。

Brynjolfsson 與 McAfee（2014）的「第二次機器時代」論述即屬此類，認為 AI 的核心價值在於自動化重複性認知勞動，釋放人類從事更具創造性的工作。

協作論觀點（Collaboration Perspective）則超越工具隱喻，將人機互動視為一種互補性的協作關係。Dellermann 等人（2019）提出「混合智慧」（Hybrid Intelligence）概念，主張人類與 AI 各有認知優勢——人類擅長抽象推理、情境判斷與倫理決策，AI 擅長大規模資訊處理、模式識別與一致性執行。在理想的協作設計中，雙方的能力形成互補而非替代。

代理論觀點（Agency Perspective）是近年隨著 AI Agent 技術發展而興起的新視角。不同於被動等待指令的工具，AI Agent 被賦予一定程度的自主決策能力，能在預設的目標框架內自主規劃與執行任務步驟（Wang et al., 2024）。Shavit 等人（2023）討論了 AI 代理行為的倫理議題，指出當 AI 具備自主行動能力時，人類對 AI 產出的監督與驗證責任不減反增。

本研究認為，在學術研究場域中，上述三種觀點並非互斥，而是對應於不同的研究任務特性。對於結構化、重複性的任務（如文獻格式整理、數據清洗），工具論的效率框架即已足夠；對於需要創造性投入的任務（如文獻分析、論點建構），協作論的互補框架更為適切；而對於多步驟、序列化的研究流程，代理論的自主執行框架則提供了新的可能性。本研究提出的工作流程設計，即試圖在這三種觀點之間取得平衡。

## 三、AI Agent 與研究流程模組化

AI Agent 強調目標導向的自主行動，能根據預設目標自動分解任務、選擇工具並迭代執行（Yao et al., 2023），多代理系統更可讓不同 Agent 各司其職（Guo et al., 2024）。Zhang 等人（2024）指出有效的代理式工作流（Agentic Workflow）需滿足任務可分解性、明確依賴關係及可驗證產出三項條件，恰與學術研究流程吻合。Gao 等人（2024）發現 AI Agent 在假設生成與數據分析展現潛力，但在理論建構與倫理判斷能力有限，支持了人機分工的必要性。

Hall（2026b）在〈The 100x Research Institution〉中提出更具前瞻性的構想。基於 AI Agent 複製政治學論文的實驗（Straus & Hall, 2026），他指出在結構明確的實證任務中，人類研究員與 AI Agent 的表現差異已極為微小——AI 複製結果與人工結果高度一致。他主張每篇實證論文發表前應附帶 AI Agent 自動複製結果的證明，使研究從靜態產品轉型為持續更新的「活研究基礎設施」（living research

infrastructure）。Hall 估計單次研究任務成本僅約 10 美元，全年 API 費用不到 5,000 美元，使資深學者指揮數百個 Agent 從事大規模研究成為可能。

然而此願景亦引發反思。Karpf（2026）擔憂學科將傾向研究「AI 容易處理的問題」而非真正重要的問題；Gunitsky（2026）認為 AI 自動化的是常態科學，對突破性研究助益有限。國內相關討論仍處起步階段，缺乏可操作的方法論指引，此一缺口正是本研究試圖填補的空間。

## 四、既有研究限制與缺口

綜合上述文獻回顧，本研究識別出以下研究缺口：

情境缺口：現有 AI 輔助研究的文獻多源自歐美學術場域，對東亞地區（特別是台灣）的 AI 使用模式缺乏實證描述。台灣學術環境的特殊性——包括中英雙語研究需求、有限的研究資源、以及獨特的學術評鑑制度——可能導致 AI 使用模式與歐美不同，需要在地化的實證研究加以探索。

方法論缺口：儘管 AI Agent 與代理式工作流（Agentic Workflow）的技術框架日趨成熟，但針對人文社會科學研究情境的方法設計仍然欠缺。現有的工作流程框架多以軟體開發或自然科學為場景，未能充分考量人文社會科學研究的特殊需求，包括文獻批判的深度要求、理論對話的複雜性，以及研究倫理的高度敏感性。

實踐缺口：學術界對 AI 協作研究的討論多停留在概念層次（「AI 可以如何幫助研究」），缺乏完整的實踐記錄與反身性分析。研究者需要的不僅是 AI 功能的介紹，而是一套可操作、可複製、可驗證的工作流程，以及對此流程在實際操作中的優勢與侷限的誠實揭露。

本研究正是基於上述三重缺口，嘗試提出一套針對人文社會科學二手資料研究的 AI Agent 協作方法論框架，並以台灣場域的實證數據進行操作展示。

## 參、研究設計與方法論

本章說明本研究的方法論定位、AI Agent 協作工作流程的設計原則、數據來源與分析方法，以及研究倫理考量。

## 一、方法論實驗的定位

本研究採取「方法論實驗」（methodological experiment）的研究取向，而非傳統的假設驗證式研究。方法論實驗的核心目標在於：透過實際操作一套新的研究方法，

檢視其可行性與侷限，並為學術社群提供可參照的實踐經驗（Hevner et al., 2004）。

具體而言，本研究的「實驗」內涵包括三個層次：第一，設計層次——提出一套 AI Agent 協作研究的工作流程框架；第二，操作層次——以 AEI 台灣數據為素材，實際執行此工作流程中的數據分析環節；第三，反思層次——記錄操作過程中的決策點、困難與發現，作為方法論的反身性素材。

需要強調的是，本研究的實證分析結果（第肆章）應被視為「方法展示」而非「研究發現」。其目的在於展示 AI Agent 協作工作流程的實際運作方式，而非對台灣 AI 使用行為提出因果性的解釋。

# 二、研究工具選擇：**Claude Code** 作為協作介面

本研究選擇 Anthropic 公司的 Claude Code 作為 AI Agent 協作的主要操作介面，底層模型為 Claude Opus 4.6（model ID: claude-opus-4-6），為 Anthropic 於 2025–2026 年間發布的最高階推理模型。本論文草稿於 2026 年 2 月在 Claude Code 環境中完成。

Claude Code 是一套基於瀏覽器的 AI 編程與研究協作環境，研究者可在其中以自然語言下達指令，由 AI Agent 執行文件讀寫、數據分析、程式碼生成、網路搜尋等多種操作，並透過 Git 版本控制系統追蹤所有操作歷程。

## 環境建置

使用 Claude Code 的前置步驟為：（1）至 GitHub（https://github.com/）註冊免費帳號，作為版本控制與檔案追蹤基礎；（2）至 Anthropic 官方網站（https://claude.ai/）訂閱 Max 方案以取得 Claude Code 使用權限；（3）登入後連結 GitHub 帳號，進入 Claude Code 瀏覽器環境；（4）以自然語言指令建立研究專案目錄結構。完成後，研究者即可以對話方式驅動 AI Agent 執行研究任務。

## 選擇 **Claude Code** 的理由

選擇 Claude Code 而非命令列介面（Command Line Interface, CLI）或其他工具，係基於三項考量：第一，降低技術門檻——提供圖形化瀏覽器介面，研究者無需安裝 Python 環境或學習命令列語法，即可透過自然語言完成數據分析；第二，整合式工作環境——將文件管理、程式碼執行、網路搜尋與版本控制整合於單一介面，所有操作自動保存於 Git 版本歷程中；第三，對話式互動模式——以多輪自然語言對話為基礎，與學術研究中「反覆修訂」的工作習慣高度吻合。

# 三、AI Agent 協作工作流程設計

## （一）設計原則

本研究提出的工作流程基於三項核心設計原則：

原則一：任務模組化（Task Modularization）。將完整的研究流程拆解為明確的子任務模組，每個模組具有清晰的輸入、處理程序與輸出。模組化設計的優勢在於：（1）降低單一任務的複雜度，使 AI Agent 能在明確的範圍內有效執行；（2）提供可檢驗的中間產出，便於人類研究者在每個節點進行品質控制；（3）增進研究過程的可重現性，其他研究者可依照相同的模組流程進行複製。

原則二：人機分工（Human-AI Division of Labor）。在每個任務模組中，明確界定人類研究者與 AI Agent 各自的職責範圍。基本分工邏輯為：人類負責「判斷性」任務（研究問題界定、理論詮釋、倫理決策、最終品質把關），AI 負責「執行性」任務（資訊檢索、數據處理、格式整理、文本初稿生成）。此分工設計呼應了 Dellermann 等人（2019）提出的混合智慧框架，充分運用人類與 AI 各自的認知優勢。

原則三：可驗證性（Verifiability）。所有 AI Agent 的產出均需經過人類研究者的審查與驗證，並保留完整的操作記錄（包括 prompt 模板、AI 原始輸出、人類修訂版本）。可驗證性原則的實踐方式包括：使用 Git 版本控制系統追蹤所有修改歷程、在每個階段設置驗證檢查清單（checklist），以及在論文中明確揭露 AI 使用情況。

## （二）七階段工作流程

基於上述原則，本研究設計了一套七階段的 AI Agent 協作研究工作流程，各階段的內容、人機分工與預期產出如表 1 所示。

表 1：AI Agent 協作研究工作流程七階段設計

| 階段 | 名稱 | 人類角色 | **AI Agent 角色** | 預期產出 |
|---|---|---|---|---|
| 0 | 研究規劃與 Agent 設定 | 定義研究問題、確定數據來源 | 協助結構化思考、建立文件架構 | 研究計畫書、專案結構 |
| 1 | 文獻蒐集 | 定義搜尋範圍、驗證相關性 | 執行搜尋、整理文獻清單 | 結構化文獻資料庫 |

| 2 | 文獻分析 | 理論詮釋、驗證分析結論 | 主題分析、缺口識別 | 文獻分析報告 |
| 3 | 數據理解與探索 | 理解數據語意、定義分析方向 | 讀取數據、描述性統計 | 數據結構文檔 |
| 4 | 數據分析與視覺化 | 定義分析問題、解讀結果 | 執行分析、生成圖表 | 分析結果與圖表 |
| 5 | 論文撰寫 | 審查內容、理論詮釋 | 撰寫各章節草稿 | 論文初稿 |
| 6 | 參考文獻整理 | 補充缺失資訊、確認格式 | 提取引用、格式化文獻 | 參考文獻清單 |

此七階段設計具有序列性（後一階段依賴前一階段的產出）但非嚴格線性——研究者可根據實際需要在階段之間進行迭代。例如，在數據分析階段（階段 4）發現新的文獻需求時，可回到文獻蒐集階段（階段 1）進行補充。

**（三）Agent 角色設計**

在工作流程中，本研究為不同階段設計了五種專業化的 AI Agent 角色：（1）文獻蒐集 Agent——系統性搜尋與整理學術文獻；（2）文獻分析 Agent——識別核心論點、比較觀點與識別研究缺口；（3）數據探索 Agent——讀取數據結構與生成描述性統計；（4）數據分析 Agent——執行統計分析與生成視覺化圖表；（5）學術寫作 Agent——撰寫論文各章節草稿。

每個 Agent 角色均包含明確的「不可為」清單（negative constraints），此乃防止 AI 越界的關鍵機制。例如，學術寫作 Agent 被要求「不得新增任何未經人類研究者提供的文獻」，以避免虛構引用（hallucinated references）的風險。

# 四、數據來源與分析方法

**（一）數據來源：Anthropic 經濟指數（AEI）**

本研究使用的實證數據來自 Anthropic 公司發布的第四期經濟指數報告（Anthropic Economic Index, 4th edition; Appel et al., 2026）。AEI 是一項追蹤 Claude.ai 平台使用行為的大規模數據集，透過對使用者對話紀錄的匿名化分析，提供全球各地區的 AI 使用模式描述。該報告提出五項「經濟基元」（economic primitives）——任務複雜度、技能水準、使用目的（工作、教育或個人）、AI 自主性與任務成功率——作

為追蹤 AI 經濟影響的基礎測量指標，並基於約兩百萬筆 AI 對話的隱私保護分析，涵蓋消費端（Claude.ai）與企業端（API）的使用數據。

本研究聚焦的台灣子集具有以下特徵：

- 數據收集期間：2025 年 11 月 13 日至 2025 年 11 月 20 日（一週）
- 平台與產品：Claude AI（Free and Pro 版本）
- 地理範圍：台灣（geo_id: TW）
- 總對話數：N = 7,729 筆
- 佔全球比例：0.77%
- 數據格式：長格式（long format），每行代表一個（facet, variable, cluster_name）組合的指標值

（二）分析面向（**Facets**）

AEI 數據採用多面向分析架構，本研究使用的分析面向包括：

類別型面向：

- request（任務請求類型）：三層級分類，level 0 為最細粒度（614 個類別），level 2 為最粗粒度（22 個類別）
- collaboration（協作模式）：6 種人機協作型態——指令式（directive）、學習式（learning）、任務迭代（task iteration）、回饋迴圈（feedback loop）、驗證式（validation）、無互動（none）
- use_case（使用情境）：工作（work）、個人（personal）、課業（coursework）
- task_success（任務成功率）：是（yes）、否（no）
- multitasking（多工性）：是否在單一對話中處理多個任務
- human_only_ability（人類獨立完成能力）：該任務是否為人類可獨立完成

數值型面向：

- ai_autonomy（AI 自主性）：1-5 分量表，衡量 AI 在任務中的自主程度
- human_education_years（人類教育年數）：完成該任務所需的估計人類教育年數
- ai_education_years（AI 教育年數）：AI 展現的等效教育年數
- human_only_time（人類獨立完成時間）：估計人類不借助 AI 完成該任務的時間（小時）
- human_with_ai_time（人機協作時間）：估計人類借助 AI 完成該任務的時間（分鐘）

此外，本研究亦使用兩份彙總資料：依任務類別分組（Group by Category, 13 類）與依職業分類分組（Group by Job, 14 類，基於美國職業資訊網絡 O*NET〔Occupational Information Network〕之標準職業分類 SOC〔Standard Occupational Classification〕系統）。

（三）分析方法

鑑於本研究的方法論實驗定位，數據分析採用描述性分析為主，具體包括：

1. 次數分配與百分比分析：描繪各類別型面向的分布特徵
2. 集中趨勢與離散趨勢：以平均數、中位數、標準差描述數值型面向的分布
3. 語意匹配分析：以關鍵字匹配法識別與學術研究相關的任務類型，估算學術使用佔比
4. 視覺化呈現：以長條圖、圓餅圖、分組長條圖等方式呈現分析結果

所有分析均透過 Claude Code 環境中以自然語言指令驅動 Python 程式語言執行，使用 pandas（資料分析套件）與 matplotlib（視覺化繪圖套件）。研究者以中文描述分析需求，AI Agent 自動生成對應的 Python 腳本並即時執行，產出的圖表輸出至專案的 `charts/` 目錄，分析流程全程由 Git（分散式版本控制系統）追蹤，以確保可重現性。

# 五、倫理考量與研究限制

（一）**AI** 使用揭露

本研究在以下環節使用 AI Agent 協助：

1. 文獻蒐集與初步整理（AI 協助搜尋與分類，人類驗證相關性）
2. 數據分析腳本生成（AI 產出 Python 腳本，人類審查邏輯後執行）
3. 論文章節草稿撰寫（AI 產出初稿，人類進行實質修訂與理論詮釋）
4. 參考文獻格式化（AI 協助格式調整，人類驗證準確性）

所有 AI 產出均經人類研究者審查、驗證與修訂。研究的核心判斷——包括研究問題定義、理論詮釋、數據解讀、研究限制評估與最終結論——皆由人類研究者完成。

（二）數據倫理

AEI 數據為 Anthropic 公司公開發布的匿名化彙總數據，不涉及個人可識別資訊。數據中的最小觀測單位為特定地區、特定面向的彙總統計值，而非個別使用者的對話記錄。此外，AEI 報告載明數據已經過隱私保護處理，觀測數低於門檻值（國家層級 200 筆、區域層級 100 筆）的數據已被排除。

（三）方法論限制前置說明

本研究在方法論層面存在以下先天限制，將於第伍章討論中進一步展開：

1. 單一平台限制：數據僅來自 Claude.ai 一個平台，無法代表台灣整體的 AI 使用情況
2. 橫斷面限制：僅涵蓋一週的使用數據，無法捕捉時序動態變化

3. 彙總數據限制：無法進行個體層次的交叉分析或迴歸分析
4. 分類系統限制：任務類型與職業分類系基於 AI 自動標註，可能存在分類誤差
5. 自我選擇偏誤：Claude.ai 使用者並非台灣學術工作者的隨機樣本

# 肆、實證示例：工作流程操作過程的後設分析

本章以後設分析的視角，記錄以 AEI 台灣數據為素材，實際執行第參章七階段工作流程的操作過程。本章的核心關懷並非「數據揭示了什麼」，而是「Agent 如何處理數據、人類如何介入」。完整的數據分析結果收錄於附錄 A。

為強化可複製性，本章依七階段逐一呈現研究者所使用的代表性 prompt，並標註其所屬的操作模式類型。基於操作經驗，本研究歸納出三類人機協作的操作模式（表 2）。

表 2：AI Agent 協作操作模式分類

| 操作模式 | 特徵 | 人類認知投入 |
|---|---|---|
| 直接執行型 | Agent 依據明確指令獨立完成，人類僅需確認產出 | 低 |
| 迭代修正型 | Agent 初次產出需經人類審查後多輪改進 | 中 |
| 人類主導型 | 分析方向與判斷邏輯由人類決定，Agent 僅負責執行 | 高 |

以下依據各階段的實際操作，呈現各類型 prompt 的範例與後設觀察。

# 一、階段 0：研究規劃與 Agent 設定

Prompt 0-1〔直接執行型〕：「我正在進行一項研究，使用 AEI 的台灣數據，分析台灣 Claude.ai 使用者的行為特徵。請建立研究專案的目錄結構，包含 manuscript、figures、data 三個資料夾，並生成一份初步的分析規劃文件。」

Agent 自動建立目錄並產出通用性的分析規劃。然而，以下三項決策由人類獨立完成〔人類主導型〕：(1)決定聚焦於「描述性分析」而非「因果推論」；(2)從多個分析面向（facets）中選定與研究問題相關者；(3)親自至 Anthropic 官方網站下載原始數據以確認真實性。此階段顯示 Agent 可處理結構化的行政任務，但研究設計的實質判斷仍需人類的領域知識。

## 二、階段 1：文獻蒐集

Prompt 1-1〔直接執行型〕：「請搜尋近三年（2023-2026）關於以下主題的學術文獻：(1) generative AI in social science research, (2) human-AI collaboration in academic writing, (3) agentic workflow。每個主題至少找 5 篇同儕審查文獻，列出作者、年份、標題、期刊與摘要。」

Agent 產出文獻清單後，人類進行以下介入〔人類主導型〕：

Prompt 1-2〔人類主導型〕：「文獻清單中的第 3、7 篇我無法在 Google Scholar 中找到，請確認這兩篇是否存在。如果不存在，請明確告知而非虛構替代文獻。」

此階段揭示了 Agent 最嚴重的風險之一——文獻幻覺（hallucinated references）。Agent 在首次產出的文獻清單中混入了不存在的文獻，且格式完整、看似可信。人類研究者必須逐一交叉驗證每筆文獻的真實性，此為不可省略的品質閘門。

【失敗修復案例：文獻幻覺的偵測與修正】

以下完整記錄一次典型的「文獻幻覺」錯誤修復歷程，作為代理式工作流中品質閘門運作的具體示範。

(1)錯誤發現：Agent 回應 Prompt 1-1 後產出 15 篇文獻清單，格式完整（含作者、年份、期刊、DOI），看似權威可信。研究者以 Google Scholar 逐一驗證時，發現其中兩篇無法查得——期刊名稱真實存在、作者為該領域活躍學者，但該篇論文從未發表。此即典型的「高擬真幻覺」：AI 並非隨機捏造，而是基於統計模式組合出「最像真實文獻」的虛構條目。

(2)診斷性提問：

Prompt 1-2a〔人類主導型〕：「文獻清單中的第 3 篇與第 7 篇我無法在 Google Scholar 中找到。請逐一確認這兩篇是否確實存在。如果不存在，請直接說明『此文獻不存在，為模型生成錯誤』，不要用其他文獻替代。」

(3)Agent 回應與修正：Agent 在接收明確指令後承認兩篇文獻「無法確認其存在」，並說明可能為「基於相似文獻特徵的不當生成」。研究者隨後要求補充替代文獻，並附加約束條件：

Prompt 1-2b〔人類主導型〕：「請補充兩篇替代文獻，條件：(1) 必須提供完整 DOI 連結，(2) 我將立即以 DOI 連結驗證其存在性，(3) 若無法提供 DOI，請標註『此文獻需人工確認』。」

(4)**驗證結果**：Agent 補充的兩篇替代文獻均附有 DOI，經研究者即時點擊驗證後確認存在。

(5)**方法論啟示**：此案例揭示三項操作原則——第一，對 Agent 產出的文獻清單應執行「全數驗證」而非抽樣檢查，因幻覺文獻的擬真度極高，難以從格式外觀辨別真偽；第二，在修正指令中應明確禁止 Agent 的「自動補償」行為（如未經詢問即以其他文獻替代），以避免錯誤在迭代中擴散；第三，要求 Agent 提供可驗證的錨點（如 DOI 連結）是降低幻覺風險的有效機制。此一「發現—診斷—約束—驗證」的四步修復流程，可作為代理式工作流中處理內容錯誤的通用範式。

## 三、階段 2：文獻分析

Prompt 2-1〔迭代修正型〕：「請針對已驗證的文獻，進行主題分析。識別三個核心主題：(1) AI 在學術研究中的角色定位、(2) 人機協作的理論框架、(3) Agentic Workflow 的設計原則。每個主題列出支持文獻與核心論點。」

**Agent 產出的主題分析初稿在文獻歸類上大致正確，但缺乏批判性——未能識別文獻之間的矛盾觀點與研究缺口。研究者需要追問：**

Prompt 2-2〔人類主導型〕：「在『AI 角色定位』主題中，Davidson & Karell（2025）與 Bail（2024）的觀點有何根本差異？請指出他們在 AI 作為研究工具的適用性上的分歧，以及此分歧對本研究方法論設計的啟示。」

**此階段確認：Agent 擅長「歸類」但不擅長「批判」，理論對話的深度需要人類引導。**

## 四、階段 3：數據理解與探索

Prompt 3-1〔直接執行型〕：「請讀取 data/ 資料夾中的 Excel 檔案，列出每份檔案的欄位名稱、資料筆數與資料型態。」

**Agent 首次嘗試因表頭格式問題而讀取失敗（欄位名稱位於第二行），需要人類修正：**

Prompt 3-2〔迭代修正型〕：「表頭在第二行，請用 header=1 參數重新讀取。」

**更關鍵的挑戰在於 Agent 對長格式數據的語意理解：**

Prompt 3-3〔人類主導型〕：「注意：同一筆對話在不同 facet 下會重複出現，不能將不同 facet 的 count 加總。總對話數應從單一 facet 計算。每個 facet 是獨立的分析維度。」

此階段操作揭示了 Agent 數據理解的三個認知層次：語法層（讀取欄位名稱）可在修正參數後完成；結構層（理解長表格式層級關係）需人類語意補充；語意層（理解 facet 的學科意涵）則完全依賴人類。

## 五、階段 4：數據分析與視覺化

Prompt 4-1〔直接執行型〕：「請從 'request' facet 中，計算各 cluster_name 的 count 與 share，以水平長條圖呈現。中文字型用 WenQuanYi Zen Hei，圖表 12x8 英寸，存入 figures/figure1_request_categories.png。」

Agent 將自然語言轉譯為完整的 Python 腳本（pandas + matplotlib），首次即正確完成數據篩選與統計計算。然而圖表呈現需要迭代修正：

Prompt 4-1a〔迭代修正型〕：「中文顯示為方塊，請用 matplotlib.font_manager 找到系統中可用的中文字型路徑再指定。類別標籤重疊，請縮小到 9pt 並增加間距。」

平均每張圖表需 3-5 輪此類修正。此外，分析邏輯的語意判斷構成關鍵的人類介入：

Prompt 4-2〔人類主導型〕：「請將『協助學術研究、寫作和教育內容』與『協助學術研究、寫作及跨學科課程』兩個類別合計，計算學術研究相關任務的總佔比。」

此合併決策——判斷兩個名稱相近的類別在概念上屬同一範疇——屬於人類對分類系統的語意理解，非 Agent 可自主完成。同樣地，從數十項 Request L1 任務類別中篩選哪些屬於「人文社會科學相關」，亦為人類的語意判斷：

Prompt 4-3〔人類主導型〕：「請從 Request L1 任務類別中，篩選出與人文社會科學學術研究直接相關的項目，以水平長條圖呈現其佔比，並計算合計佔比。篩選標準：翻譯、學術寫作、教育教學、研究方法、文獻處理等與人社學術工作直接相關者。」

效率擴增的實證發現

在數據分析 Agent 的協助下，本研究得以迅速萃取 AEI 資料中最具衝擊力的效率指標。表 3 呈現台灣使用者在 AI 協助下的時間效益數據，此組數據最能具體展現現代理式工作流將研究者從繁瑣勞動中解放的潛力。

表 3：AI 協助的時間效益——台灣 Claude.ai 使用者（N=7,729）

| 指標 | 數值 |
|---|---|
| 人類獨力完成時間（中位數） | *Mdn* = 1.75 小時（105 分鐘） |
| AI 協助完成時間（中位數） | *Mdn* = 12.0 分鐘 |

| 中位數時間節省率 | 約 89% |
| --- | --- |
| 人類獨力完成時間（平均） | *M* = 3.55 小時, 95% CI [3.43, 3.68] |
| AI 協助完成時間（平均） | *M* = 18.7 分鐘, 95% CI [18.2, 19.2] |

以中位數計，AI 將典型任務的完成時間從 1.75 小時壓縮至 12 分鐘——約為原本的九分之一。此一效率增益的規模值得深思：若一位人文社會科學研究者每日有三項可由 AI 輔助的結構化任務（文獻翻譯、數據整理、格式編排），按中位數估算，每日可釋出約 4.65 小時的認知資源，重新投入理論思考、田野觀察或跨學科對話等 AI 尚無法勝任的高價值工作。

更關鍵的是數據分析 Agent 自身的運作效率。在本研究的操作過程中，從「以自然語言描述分析需求」到「獲得完整統計結果與圖表」，單次分析迭代的平均耗時約 2-3 分鐘。傳統研究流程中，研究者需自行撰寫 Python 或 R 腳本、除錯、調整圖表參數——同等工作量往往耗費數小時。代理式工作流的效率優勢並非來自「更快的計算」，而是來自「自然語言到程式碼的即時轉譯」，使不具程式設計背景的研究者亦能直接以學術語言驅動數據分析。此一發現呼應了 Brynjolfsson 與 McAfee（2014）的「擴增」觀點——AI 的核心價值在於擴增研究者的分析能力半徑，而非替代其研究判斷。完整的時間效益分析詳見附錄 A 第 6 節。

此階段的後設觀察為：Agent 可執行任何明確定義的計算，但「什麼應該被計算」的決策仍需人類的領域知識。各分析的完整結果見附錄 A。

# 六、階段 5：論文撰寫

Prompt 5-1〔直接執行型〕：「請依據上述分析結果，撰寫數據分析報告。結構：第 1 節 整體輪廓、第 2 節 使用場景分布、第 3 節 O*NET 與任務分類方法、第 4 節 任務類型排行、第 5 節 O*NET 職業任務分析、第 6 節 教育程度與時間效益、第 7 節 人文社會科學影響推論、第 8 節 結論與建議。引用對應圖表，使用正體中文學術風格。」

Agent 產出約 8,000 字的報告初稿，數據引用準確，但理論連結需要人類指定：

Prompt 5-2〔人類主導型〕：「請在人類獨立完成能力分析段落中，加入理論詮釋：此發現呼應 Brynjolfsson 與 McAfee（2014）的『擴增』觀點——AI 的主要價值在於擴增效率而非替代能力。以此詮釋 82.9% 的人類可獨立完成率。」

Agent 成功將理論框架整合進文字，但詮釋的適切性仍需人類確認。研究者對初稿的修訂幅度約 30-40%，集中在理論連結與論述邏輯。此階段顯示：Agent 是理論文字的「組裝者」而非「創造者」，修訂後的最終版本收錄於附錄 A。

## 七、階段 6：參考文獻整理

Prompt 6-1〔直接執行型〕：「請從論文各章節中提取所有引用的文獻，整理為美國心理學會（American Psychological Association, APA）第七版格式的參考文獻清單，按作者姓氏字母排序。」

Agent 正確提取並格式化大部分文獻，但需要迭代修正：

Prompt 6-2〔迭代修正型〕：「請檢查文獻清單中每筆文獻的數位物件識別碼（Digital Object Identifier, DOI）連結是否正確。格式統一為 https://doi.org/ 開頭。缺少 DOI 的文獻請標註。」

人類在此階段的關鍵介入為補充 Agent 無法取得的出版資訊〔人類主導型〕——例如預印本（preprint）的正式出版資訊、中文文獻的翻譯標題格式，以及確認所有被引文獻確實存在。

## 八、後設分析總結

（一）操作模式的階段分布

綜合七階段的操作經驗，各操作模式的出現頻率呈現規律性分布（表 4）。

表 4：操作模式的階段分布

| 階段 | 直接執行型 | 迭代修正型 | 人類主導型 |
|------|-----------|-----------|-----------|
| 0 研究規劃 | ● | | ●● |
| 1 文獻蒐集 | ● | | ●● |
| 2 文獻分析 | | ● | ●● |
| 3 數據探索 | ● | ● | ●● |
| 4 數據分析 | ●● | ●● | ●● |
| 5 論文撰寫 | ● | | ●● |
| 6 參考文獻 | ●● | ● | ● |

註：● 表示該模式在該階段出現的相對頻率。

由表可見，人類主導型操作貫穿所有階段，驗證了第參章的人機分工原則。直接執行型主要集中於結構化程度高的階段（0、4、6）；迭代修正型則集中於涉及呈現品質的環節（3、4、6）。

（二）可複製性與限制

本章提供的 prompt 序列可供後續研究者在不同數據集上重現類似的操作過程。然而，由於大型語言模型（Large Language Model, LLM）生成具有隨機性，相同 prompt 在不同時間點可能產生略有差異的回應。因此，上述 prompt 應被理解為「操作邏輯的參照」而非「精確複製的腳本」。

# 伍、討論

本章就前述方法論實驗的操作經驗（第肆章），從三個面向進行討論：對人文社會科學研究流程管理的意涵、對研究實務的啟示，以及 AI 導入學術研究的限制與風險。附錄 A 的實證分析結果作為工作流程的產出物，在本章中僅作為討論的輔助佐證。

## 一、對人文社會科學研究流程管理的意涵

（一）從「單點工具」到「流程架構」的典範轉移

本研究提出的七階段 AI Agent 協作工作流程，代表了一種不同於現有 AI 使用討論的思考方式。目前學術界對 AI 的討論多集中在「AI 能做什麼」——翻譯、摘要、編碼、統計等單點功能——而較少關注「AI 如何嵌入完整的研究流程」。本研究的方法論實驗顯示，AI 的研究價值不僅在於單一任務的效率提升，更在於作為「流程架構」（workflow infrastructure）串連研究的多個階段，使整體研究過程更為結構化與可管理。

此一觀察呼應了 Zhang 等人（2024）關於代理式工作流（Agentic Workflow）的論述。有效的 AI 工作流程並非將 AI 投入個別任務，而是設計一套任務之間的銜接邏輯——前一階段的產出即為後一階段的輸入，每個銜接點都設有人類審查的品質閘門（quality gate）。對人文社會科學研究者而言，這種流程思維的價值在於：將原本高度依賴個人經驗與直覺的研究過程，轉化為可教授、可複製、可改進的結構化流程。

（二）「人類判斷不可替代性」的具體機制

第肆章的操作過程記錄中反覆浮現的一個現象是：AI Agent 在「執行」層面的表現相當穩定（數據讀取、統計計算、圖表生成等），但在「判斷」層面的能力存在明確邊界。本研究識別出四類人類研究者不可替代的判斷功能：

1. 研究問題的界定：決定「什麼值得研究」是價值判斷而非技術操作。在本研究中，選擇以台灣學術場域的 AI 使用模式為研究焦點，涉及對學術社群需求的理解與對研究缺口的識別，這些判斷無法由 AI 代勞。
2. 理論詮釋：將數據轉化為理論論述需要跨領域的知識整合。例如，82.9% 的人類可獨立完成率「意味著」使用者將 AI 視為擴增工具而非替代工具──這一詮釋需要研究者對人機協作理論、技術採納理論的深度理解。
3. 脈絡化判斷：AEI 數據呈現的是全球標準化指標，但台灣場域的特殊性（學術評鑑制度、研究資源配置、語言環境等）需要在地化的脈絡理解。AI 可以處理數據，但無法理解數據背後的社會脈絡。
4. 倫理反思：研究限制的自我揭露、AI 使用的透明度要求、數據詮釋的謙遜態度──這些學術倫理的實踐依賴研究者的專業判斷與道德責任感。

上述四類判斷功能構成了人文社會科學研究中「人類不可替代性」的具體內涵。此一發現呼應了 Shavit 等人（2023）的論述──AI Agent 的自主性越高，人類監督的責任越重，而非越輕。

（三）模組化設計對研究訓練的啟示

模組化工作流程對研究生訓練具有潛在的教育價值──提供一種比傳統師徒制更為結構化的訓練路徑，使研究生可循序漸進地在每個模組中練習人機分工的判斷。然而，過度依賴結構化流程可能限制創造性思維，因此模組化工作流程應被定位為「鷹架」（scaffolding）──在研究者成熟後應能彈性調整甚至超越。

# 二、對科技管理與研究實務的啟示

（一）學術場域的 AI 採納特徵

AEI 台灣數據分析（詳見附錄 A）揭示了學術場域 AI 採納的若干結構性特徵。學術研究寫作兩大類合計佔比達 17.3%，加上翻譯（8.5%）更達 25.8%，顯示 AI 工具在學術場域的滲透已具有一定規模。然而，就單一類別而言，各學術子類（8.9% 與 8.4%）均低於軟體開發（14.7% 為單一最大類別），顯示學術場域的 AI 採納仍處於相對早期階段。

從科技管理的角度，學術場域的 AI 採納呈現出 Rogers（2003）創新擴散理論中「早期多數」（early majority）的特徵：使用者已超越了早期採納者（early adopters），但尚未達至普及階段。推動此一擴散的可能因素包括：(1)學術競爭壓力下的效率需求；(2)跨語言研究的翻譯需求（台灣的 8.5% 翻譯使用佔比反映了此一需求）；(3)數據分析的技術賦能需求。

（二）協作模式的管理意涵

人機協作模式的三足鼎立分布（指令式 27.9%、任務迭代 26.6%、學習式 25.3%）對研究機構的 AI 導入策略具有管理意涵。不同的協作模式對應不同的組織支持需求：

- 指令式使用需要清晰的 prompt 範本與最佳實踐指引
- 迭代式使用需要足夠的 AI 使用時間與額度支持
- 學習式使用需要鼓勵探索性使用的組織文化

研究機構在制定 AI 使用政策時，不應僅關注「是否允許使用 AI」的二元問題，而應考慮如何為不同的協作模式提供適當的制度支持。

## （三）效率擴增的實務價值

82.9% 的人類可獨立完成率，以及教育年數的高度對齊（99.1%），共同描繪了一幅「效率擴增」而非「能力替代」的使用圖像。AI 的導入不需要研究者放棄既有能力，而是透過 AI 處理結構化的輔助任務（文獻篩選、描述性統計、格式整理、翻譯等），將認知資源集中於需要深度思考的研究核心。

## （四）組織學習：從研究機構到「Agent 協作中心」

本研究的操作經驗暗示，研究機構的組織型態可能需要根本性的調整以適應代理式工作流的導入。傳統研究機構（如政策智庫）的運作模式以「資深研究員—初級研究員—研究助理」的層級式分工為主：資深研究員規劃研究方向，初級研究員與助理負責數據蒐集、文獻整理與報告撰寫。此一模式的產能受限於人力配置——每位資深研究員可指揮的助理數量有限，研究規模因此受到物理性的約束。

Hall（2026b）提出的構想為此一困境指明了轉型方向。他主張，在 AI Agent 的成本已降至單次研究任務約 10 美元、全年 API 費用不到 5,000 美元的條件下，資深研究員可以「一人指揮多 Agent」的平行作業模式，同時推進數個研究子任務。此一模式的具體實踐可能是：資深研究員透過事先設定的 Agent 工作流程，對動態更新的數據建立即時分析儀表板（Dashboard），使政策研究從「季度報告」的靜態產出轉型為「即時監測」的動態分析基礎設施。例如，針對台灣 AI 採納趨勢，研究員可設定 Agent 定期擷取 AEI 最新數據、自動產生比較分析並標記異常變化，研究員僅需審閱 Agent 產出的摘要並做出判斷性決策。

對研究機構而言，此一轉型意味著組織學習（organizational learning）的重心將從「培訓研究員掌握特定分析技術」轉向「培訓研究員設計與管理 Agent 工作流程」。機構的核心競爭力不再僅是「擁有多少研究人力」，而是「能否建構高效的人機協作基礎設施」。具體而言，研究機構可考慮以下轉型路徑：（1）建立共享的 Agent 模板

庫，將常用研究工作流程（如文獻回顧、政策比較、國際趨勢掃描）標準化為可重用的模板；(2)設置 Agent 協作支援團隊，協助資深研究員設計與最佳化工作流程；(3)重新定義績效指標，從「個人產出量」轉向「人機協作的研究品質與洞見深度」。

## （五）去技能化風險：研究者核心能力的位移

然而，代理式工作流的效率優勢並非毫無代價。當 Agent 承擔了文獻彙整、數據處理與報告初稿等結構化工作後，研究者長期不練習這些基礎技能，可能導致「去技能化」(de-skilling)效應——如同自動駕駛技術的普及可能弱化駕駛者的手動操控能力。

此一風險的核心在於研究者核心能力的位移。在傳統研究訓練中，「快速彙整大量文獻」「熟練操作統計軟體」「精確格式化參考文獻」被視為研究者的基本功。然而，當這些任務由 Agent 高效執行後，研究者的核心競爭力將不可避免地從「快速彙整」轉向「識別文獻價值」——亦即判斷哪些文獻值得深讀、哪些論點具有理論潛力、哪些數據模式反映真實現象而非統計偶然。同理，「資料品質判斷」的洞察力將比「快速整理資料」的操作效率更具不可替代性。

此一能力位移對研究生教育尤其具有挑戰性。若研究生從訓練初期即依賴 Agent 完成基礎研究任務，可能無法建立足夠的「底層認知」(ground-level cognition)——對數據結構的直覺理解、對文獻脈絡的深層記憶、對研究方法假設的內化掌握。這些隱性知識(tacit knowledge)往往正是透過反覆的手動操作而逐漸內化的。

因此，研究機構在推動 Agent 協作的同時，應有意識地維護研究者的核心判斷能力：在訓練階段保留手動操作的環節，使研究生建立堅實的基礎認知；在實戰階段善用 Agent 提升效率，但定期進行「無 AI 演練」以維持獨立研究能力。此一平衡——善用 AI 的效率而不依賴 AI 的判斷——正是代理工作流從「工具使用」走向「成熟協作」的關鍵分野。

# 三、AI 導入學術研究的限制與風險

## （一）AI 幻覺與學術誠信風險

生成式 AI 的「幻覺」(hallucination)問題——即產生看似合理但實際不正確的內容——在學術研究中構成嚴重的誠信風險。在本研究的實際操作中，此風險主要出現在兩個環節：文獻引用(AI 可能生成不存在的文獻)與數據解讀(AI 可能對數據做出不準確的描述)。

本研究的工作流程設計以「可驗證性原則」應對此風險——所有 AI 產出均需經人類審查，且要求 AI 標註所有來源以便追溯。然而，此一機制的有效性高度依賴人類研究者的審查意願與能力。若研究者因時間壓力或信任偏誤而放鬆審查標準，幻覺風險將難以被有效控制。

（二）方法論適用性的邊界

本研究提出的工作流程主要適用於二手資料分析型的研究，對於以下研究類型的適用性存在明確限制：

1. 原創性理論建構：需要高度創造性思維的理論開發工作，AI 目前無法有效替代人類研究者的概念化能力。
2. 深度詮釋性的質性研究：民族誌、深度訪談、論述分析等需要研究者沉浸於田野脈絡的研究，AI 無法提供替代性的詮釋深度。
3. 涉及敏感個資的研究：將研究數據輸入 AI 平台可能構成資料外洩風險，不適用於涉及人類受試者隱私的研究。
4. 需要即時互動的田野研究：AI 無法參與訪談、觀察或實驗等需要現場互動的數據蒐集過程。

（三）數位落差與公平性考量

AI 協作研究的導入可能加劇學術場域中既有的數位落差。擁有更好的 AI 工具取用管道（如付費版本的 Claude Pro）、更強的 prompt 設計能力、以及更熟悉英語的研究者，將更能有效運用 AI 協作工作流程。此一不平等可能沿著世代、學科、語言與機構資源等軸線展開，值得學術管理者與政策制定者關注。

（四）學術生態系統的長期影響

若 AI 協作研究成為常態，可能對學術生態系統產生結構性影響：研究產出的速度加快可能加劇「發表或滅亡」（publish or perish）的壓力；AI 輔助撰寫可能使論文風格趨於同質化；對 AI 的依賴可能削弱研究者獨立完成基礎研究任務的能力。這些長期影響目前尚難以實證評估，但值得學術社群在擁抱 AI 效率的同時保持警覺。

# 四、本研究的限制

本研究存在以下限制，讀者在詮釋研究結果時應予以考量：

數據限制：（1）數據僅來自 Claude.ai 單一平台，ChatGPT、Gemini 等其他 AI 平台的使用者可能呈現不同的行為模式；（2）一週的橫斷面數據無法反映時序變化與趨勢；（3）彙總層級的數據限制了交叉分析的可能性；（4）台灣佔全球 0.77% 的使用量，樣本規模的代表性需謹慎評估。

方法論限制:(1)方法論實驗僅由單一研究團隊執行,其可複製性尚待更多研究者的獨立驗證;(2)工作流程的設計基於特定的研究情境(二手資料分析),對其他研究類型的適用性未經檢驗;(3)本研究本身即使用 AI 協作完成,存在自我指涉的方法論循環。

推論限制:(1)描述性分析無法建立因果關係,本研究的所有發現均為相關性描述而非因果主張;(2)AEI 數據中的分類標籤係由 AI 自動標註,分類的準確性無法獨立驗證;(3)本研究無法區分使用者的「實際行為」與「最佳行為」——數據反映的是使用者如何使用 AI,而非如何最有效地使用 AI。

# 陸、結論與未來研究建議

## 一、研究問題回顧與核心發現

本研究以「方法論實驗」為定位,探討如何利用 AI Agent 擴增台灣人文社會科學研究的視野,並以 Anthropic 經濟指數的台灣數據作為驗證方法論可行性的實證載體。回顧本研究的兩個研究問題,核心發現如下:

針對研究問題一(工作流程設計與設計原則),本研究提出了基於「任務模組化」「人機分工」「可驗證性」三項原則的七階段協作工作流程。此流程涵蓋從研究規劃到參考文獻整理的完整研究歷程,每一階段均配置專責的 AI Agent 角色,並設有人類審查的品質閘門。此一設計嘗試在 AI 的執行效率與人類的判斷深度之間建立結構化的協作機制。

針對研究問題二(可操作性與分工邊界),透過 AEI 數據的實際分析操作(第肆章),本研究發現此工作流程在二手資料描述性分析的場景中具有可操作性。第肆章的後設分析顯示:AI Agent 在數據讀取、統計計算、圖表生成、文本初稿等「直接執行型」與「迭代修正型」任務中表現穩定;而研究問題界定、理論詮釋、脈絡化判斷、倫理反思等「人類主導型」環節則構成人類研究者不可替代的判斷領域。人機分工的邊界並非固定不變,而是隨任務的結構化程度與詮釋需求而動態調整。

## 二、研究工作型態的改變

本研究的發現對研究者的工作型態具有以下啟示:

第一,文獻與數據工作的效率轉型。AI Agent 可將文獻搜尋、摘要提取、格式整理、描述性統計等結構化任務的效率顯著提升,使研究者將更多認知資源投入批判性閱讀與理論對話。對缺乏程式設計背景的人文社會科學研究者,AI Agent 更可作為

「技術中介」，降低量化分析的入門門檻。然而，效率轉型的前提是研究者具備評估 AI 產出品質的能力。

第二，從「獨立完成」到「協作管理」。AI Agent 協作模式將研究者的角色從「所有任務的獨立執行者」轉變為「人機協作的管理者」——需要學習委派任務、設計 prompt 指令與審查產出品質。此一角色轉變需要新的管理技能，但也提供了規模化研究產出與跨學科研究的可能性。

第三，品質控制與倫理的新挑戰。AI 協作帶來新的品質控制需求（事實準確性、論述邏輯性、引用適切性）與倫理判斷需求（何種使用屬於合理的工具輔助、如何透明揭露 AI 使用情況）。這些問題的答案尚未定型，需要學術社群持續對話建立共識。

## 三、核心貢獻總結

綜合而言，本研究的核心貢獻在於三個層面：

方法論層面：提出了一套經過實際操作檢驗的 AI Agent 協作研究工作流程，以「任務模組化、人機分工、可驗證性」三項原則為設計基礎，為人文社會科學研究者導入 AI 工具提供了可參照的框架。此框架的價值不在於「最佳實踐」的宣稱，而在於提供一個可供批評、修正與改進的起點。

反身性層面：透過第肆章以後設分析視角完整記錄 AI 協作研究的操作過程，歸納出「直接執行型」「迭代修正型」「人類主導型」三類操作模式，揭示了「人類判斷不可替代性」的四類具體機制（研究問題界定、理論詮釋、脈絡化判斷、倫理反思），為學術界討論 AI 使用的邊界與倫理提供了第一手案例。

實證層面：附錄 A 以 AEI 台灣數據（N=7,729）的完整分析，提供了台灣學術場域 AI 使用行為的描述性實證基礎——學術相關任務（含翻譯）佔比達 25.8%、時間節省率約 89%、82.9% 的任務人類可獨力完成，勾勒出「效率加速而非能力替代」的使用圖像，為後續研究台灣學術 AI 採納提供可對照的基線數據，亦供讀者獨立評估方法論的實際效能。

## 四、未來研究方向

基於本研究的發現與限制，提出以下五個未來研究方向：

方向一：跨平台比較研究。將分析範圍擴展至 ChatGPT、Gemini、Copilot 等多個 AI 平台，比較不同平台使用者的行為模式差異，建立更為全面的台灣 AI 使用圖像。

方向二：縱貫性追蹤研究。以時間序列數據追蹤 AI 使用行為的動態變化，觀察學術場域的 AI 採納是否呈現擴散曲線的典型模式，以及協作模式是否隨使用經驗的累積而演化。

方向三：學科差異比較。針對人文社會科學內部的學科差異（社會學、經濟學、教育學、管理學等），比較不同學科研究者的 AI 使用模式、協作偏好與導入障礙。

方向四：工作流程的實驗評估。設計對照實驗，比較使用 AI Agent 協作工作流程與傳統研究流程的研究產出品質、效率差異與研究者體驗，以更嚴謹的方式評估工作流程的效益。

方向五：AI 協作倫理框架的建構。在學術 AI 使用的倫理面向，發展更為精細的評估框架，包括 AI 使用揭露的標準化格式、人類實質貢獻的判定標準，以及 AI 生成內容的品質保證機制。

## 五、結語

從「勞動」到「協作」的隱喻轉變，描繪了 AI 工具在學術研究中角色演進的軌跡。過去，研究者以個人的認知勞動完成研究的全部環節；現在，AI Agent 的導入使得研究流程中的部分環節可以透過人機協作來完成。此一轉變並不意味研究者的角色被弱化——恰好相反，當結構化的執行任務被 AI 分擔之後，研究者的核心價值更加集中於判斷、詮釋與反思等不可替代的認知功能。

本研究以方法論實驗的謙遜姿態，嘗試為此一轉變提供一套可操作的框架與一組可供對話的實證數據。本研究充分認識到，一篇論文無法回答 AI 對學術研究的所有影響——但若能為人文社會科學研究者提供一個開始探索的起點，引發更多關於 AI 協作方法論的批判性討論，則本研究的目的已達。

## 參考文獻

### 英文文獻（依作者姓氏字母排序）

[1] Acemoglu, D., & Restrepo, P. (2019). Automation and new tasks: How technology displaces and reinstates labor. *Journal of Economic Perspectives*, 33(2), 3-30. https://doi.org/10.1257/jep.33.2.3

[2] Anthropic (2025). The Anthropic Economic Index (4th ed.). *Anthropic Research*. https://www.anthropic.com/research/economic-index-primitives

[3] Appel, R., Massenkoff, M., McCrory, P., McCain, M., Heller, R., Neylon, T., & Tamkin, A. (2026, January 15). Anthropic Economic Index report: Economic primitives. *Anthropic Research*.

https://www.anthropic.com/research/anthropic-economic-index-january-2026-report

[4] Brynjolfsson, E., & McAfee, A. (2014). *The second machine age: Work, progress, and prosperity in a time of brilliant technologies*. W. W. Norton & Company. https://wwnorton.com/books/the-second-machine-age/

[5] Davidson, T., & Karell, D. (2025). Integrating generative artificial intelligence into social science research: Measurement, prompting, and simulation. *Sociological Methods & Research*. https://doi.org/10.1177/00491241251339184

[6] Dell'Acqua, F., McFowland, E., Mollick, E. R., Lifshitz-Assaf, H., Kellogg, K., Rajendran, S., Krayer, L., Candelon, F., & Lakhani, K. R. (2023). Navigating the jagged technological frontier: Field experimental evidence of the effects of AI on knowledge worker productivity and quality. *Harvard Business School Technology & Operations Mgt. Unit Working Paper*, No. 24-013.

https://doi.org/10.2139/ssrn.4573321

[7] Dellermann, D., Ebel, P., Söllner, M., & Leimeister, J. M. (2019). Hybrid intelligence. *Business & Information Systems Engineering*, 61(5), 637-643. https://doi.org/10.1007/s12599-019-00595-2

[8] Eloundou, T., Manning, S., Mishkin, P., & Rock, D. (2024). GPTs are GPTs: An early look at the labor market impact potential of large language models. *Science*, 384(6702), 1306-1308. https://doi.org/10.1126/science.adj0998

[9] Gao, J., & Wang, D. (2024). Quantifying the use and potential benefits of artificial intelligence in scientific research. *Nature Human Behaviour*, 8(12), 2281-2292. https://doi.org/10.1038/s41562-024-02020-5

[10] Gao, S., Fang, A., Huang, Y., Giunchiglia, V., Noori, A., Schwarz, J. R., Ektefaie, Y., Kondic, J., & Zitnik, M. (2024). Empowering biomedical discovery with AI agents. *Cell*, 187(22), 6125-6151.
https://doi.org/10.1016/j.cell.2024.09.022

[11] Gunitsky, S. (2026, January). The age of academic slop is upon us. *Hegemon* (Substack).
https://hegemon.substack.com/p/the-age-of-academic-slop-is-upon

[12] Guo, T., Chen, X., Wang, Y., Chang, R., Pei, S., Chawla, N. V., Wiest, O., & Zhang, X. (2024). Large language model based multi-agents: A survey of progress and challenges. In *Proceedings of the 33rd International Joint Conference on Artificial Intelligence (IJCAI 2024)*, 8048-8057.
https://doi.org/10.24963/ijcai.2024/890

[13] Hall, A. B. (2026a, January 5). Claude Code and its ilk are coming for the study of politics like a freight train [Post]. *X (formerly Twitter)*.
https://x.com/ahall_research/status/2007221974947508303

[14] Hall, A. B. (2026b, January 13). The 100x research institution. *Free Systems* (Substack).
https://freesystems.substack.com/p/the-100x-research-institution

[15] Hevner, A. R., March, S. T., Park, J., & Ram, S. (2004). Design science in information systems research. *MIS Quarterly*, 28(1), 75-105.
https://doi.org/10.2307/25148625

[16] Karpf, D. (2026, January). What comes next, if Claude Code is as good as people say. *Substack*.
https://davekarpf.substack.com/p/what-comes-next-if-claude-code-is

[17] Mollick, E. R., & Mollick, L. (2023). Using AI to implement effective teaching strategies in classrooms: Five strategies, including prompts. *The Wharton School Research Paper*. https://doi.org/10.2139/ssrn.4391243

[18] Mondal, H., Mondal, S., & Enoch, I. T. (2023). ChatGPT in academic writing: Maximizing its benefits and minimizing the risks. *Indian Journal of Ophthalmology*, 71(12), 3600-3606. https://doi.org/10.4103/IJO.IJO_718_23

[19] Rogers, E. M. (2003). *Diffusion of innovations* (5th ed.). Free Press. https://www.simonandschuster.com/books/Diffusion-of-Innovations-5th-Edition/Everett-M-Rogers/9780743222099

[20] Shavit, Y., Agarwal, S., Brundage, M., Adler, S., O'Keefe, C., Campbell, R., ... & Robinson, D. G. (2023). Practices for governing agentic AI systems. *OpenAI Research*. https://openai.com/research/practices-for-governing-agentic-ai-systems

[21] Straus, G., & Hall, A. B. (2026). How accurately did Claude Code replicate and extend a published political science paper? *Working Paper, Stanford University*. https://www.andrewbenjaminhall.com/Straus_Hall_Claude_Audit.pdf

[22] Thompson, D. M., Wu, J. A., Yoder, J., & Hall, A. B. (2020). Universal vote-by-mail has no impact on partisan turnout or vote share. *Proceedings of the National Academy of Sciences*, 117(25), 14052-14056. https://doi.org/10.1073/pnas.2007249117

[23] Wang, L., Ma, C., Feng, X., Zhang, Z., Yang, H., Zhang, J., ... & Wen, J. R. (2024). A survey on large language model based autonomous agents. *Frontiers of Computer Science*, 18(6), 186345. https://doi.org/10.1007/s11704-024-40231-1

[24] Yao, S., Zhao, J., Yu, D., Du, N., Shafran, I., Narasimhan, K., & Cao, Y. (2023). ReAct: Synergizing reasoning and acting in language models. *Proceedings of the International Conference on Learning Representations (ICLR 2023)*. https://doi.org/10.48550/arXiv.2210.03629

[25] Zhang, J., Xiang, J., Yu, Z., Teng, F., Chen, X., Chen, J., Zhuge, M., Cheng, X., Hong, S., Wang, J., Zheng, B., Liu, B., Luo, Y., & Wu, C. (2024). AFlow: Automating agentic workflow generation. In *Proceedings of the 13th*

# 附錄 A. 台灣 Claude.ai 使用行為分析——以 Anthropic 經濟指數第四版資料為基礎

資料區間：2025 年 11 月 13 日至 2025 年 11 月 20 日 資料範圍：台灣（geo_id = TW），Claude AI (Free and Pro) 平台 樣本數：N = 7,729 筆對話

---

附錄說明：本附錄為人機協作分析之示範產出。研究者提供分析方向、報告架構與學術判斷，Claude（Anthropic, claude-opus-4-6）執行資料處理、圖表產生與初稿撰寫。研究者負責最終審校、詮釋校正與數據驗證。資料來源為 Anthropic Economic Index 第四版公開原始資料之台灣子集。

---

## 文獻背景：以「任務」而非「職業」衡量 AI 的經濟影響

傳統勞動經濟學對自動化衝擊的研究，多以「職業」（occupations）為分析單位——例如探問「AI 是否會取代會計師」或「記者的工作是否受到威脅」。然而，近年來學界逐漸轉向更精細的「任務層級」（task-level）分析框架。Acemoglu 與 Restrepo（2019）指出，自動化並非整體取代某一職業，而是選擇性地影響該職業中的特定任務——一個職業（job）包含多項技能（skills）與任務（tasks），AI 可能接管其中部分任務，同時催生新的任務需求。此一「任務取向」的分析視角，較「職業取向」更能精確描繪 AI 對勞動力市場的異質性影響。

在此脈絡下，Eloundou 等人（2024）以「GPTs are GPTs」的雙關語，論證大型語言模型作為「通用技術」（General-Purpose Technology, GPT），其影響並非局限於特定產業或特定職業，而是以不均勻但廣泛的方式滲透各類知識工作的具體任務。他們估計，約 80% 的美國勞動力至少有 10% 的工作任務會受到 LLM 的影響，而約 19% 的勞動力則有超過 50% 的任務會受到衝擊——此一發現強化了「逐任務分析」的必要性。

Anthropic 經濟指數（AEI）正是目前極少數能在任務層級進行大規模實證觀測的指標。不同於傳統的職業調查或專家預測，AEI 直接分析約兩百萬筆真實 AI 對話，透過美國職業資訊網絡（O*NET）的職業任務資料庫與多層級任務分類法（Request taxonomy），將每筆對話精確對應至具體的工作任務與技能需求（Anthropic, 2025;

Appel et al., 2026）。此一「由下而上」的實證策略，使研究者得以觀察 AI 實際被用於哪些任務、這些任務原本需要多少人力時間與教育程度——而非僅憑專家判斷預測「AI 可能影響哪些職業」。

以下報告即以此任務層級的分析框架，檢視台灣 Claude.ai 使用者的行為特徵與使用模式。

---

## 第 1 節　整體輪廓

本節摘要台灣 Claude.ai 使用行為的五項關鍵績效指標（KPI），勾勒台灣使用者的基本面貌。

表 A-1: 台灣 Claude.ai 使用行為關鍵指標（N=7,729）

| 指標 | 數值 | 說明 |
|------|------|------|
| 樣本對話數 | 7,729 | 一週內台灣使用者產生的 Claude.ai 對話總數 |
| 全球佔比 | 0.77% | 台灣對話數佔全球總量的比例 |
| 任務成功率 | 68.7% | 被評估為成功完成的對話佔比 |
| 人類可獨力完成比例 | 82.9% | 使用者被評估為可不依賴 AI 獨力完成該任務的佔比 |
| AI 自主性中位數 | 4.0 / 5.0 | AI 在對話中被賦予的自主程度（1=完全受指令，5=高度自主） |

台灣以 0.77% 的全球佔比，在 Claude.ai 使用版圖中佔有一席之地。考量台灣人口約佔全球 0.3%，此佔比反映台灣知識工作者對生成式 AI 工具具有高於全球平均的採納密度。

五項指標中最具分析意涵的是人類可獨力完成比例（82.9%）。此數據顯示，絕大多數使用者並非因為「無法完成」而求助 AI，而是選擇讓 AI 處理自己原本有能力完成的工作。換言之，Claude 在台灣的使用模式以效率加速（acceleration）為主，而非能力替代（substitution）。使用者將 AI 視為提升產出速度的工具，而非填補技能缺口的拐杖。

任務成功率 68.7% 則顯示，約三分之一的對話並未達成預期目標。此一數據提醒: AI 協作並非萬無一失，人類審查與修正仍為必要環節。AI 自主性中位數達 4.0（滿

分 5.0），顯示使用者傾向賦予 AI 較高的自主空間，而非逐步微調指令——此模式與「效率加速」的定位一致。

---

## 第 2 節　使用場景分布

### 2.1 使用場景

圖 A-1 呈現台灣 Claude 使用者的三種使用場景分布。

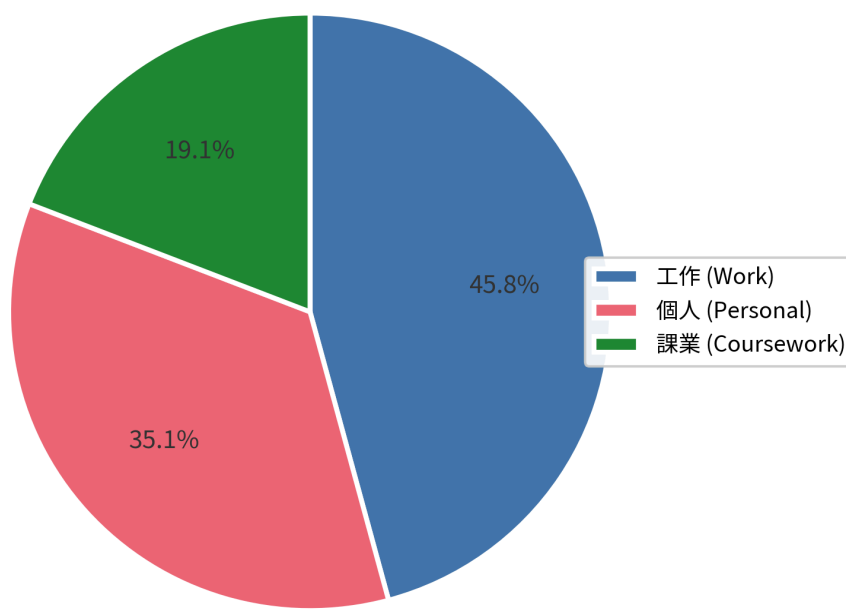

**圖 A1. 台灣 Claude 使用者使用場景分布（N=7,729）**

工作場景佔比最高（45.8%），其次為個人用途（35.1%），課業場景佔 19.1%。三者分布呈現明確的梯度結構，工作用途約佔半數，但個人與課業合計亦超過半數（54.2%），顯示 Claude 的使用並非侷限於職場。

值得注意的是，課業場景佔近五分之一。此佔比暗示學生與研究者構成不可忽視的使用族群。對人文社會科學領域而言，此數據意味著相當比例的使用者可能為研究生或學術工作者，將 Claude 用於課業報告、論文寫作、文獻翻譯等學術任務。

### 2.2 人機協作模式

圖 A-2 呈現六種人機協作模式的分布。

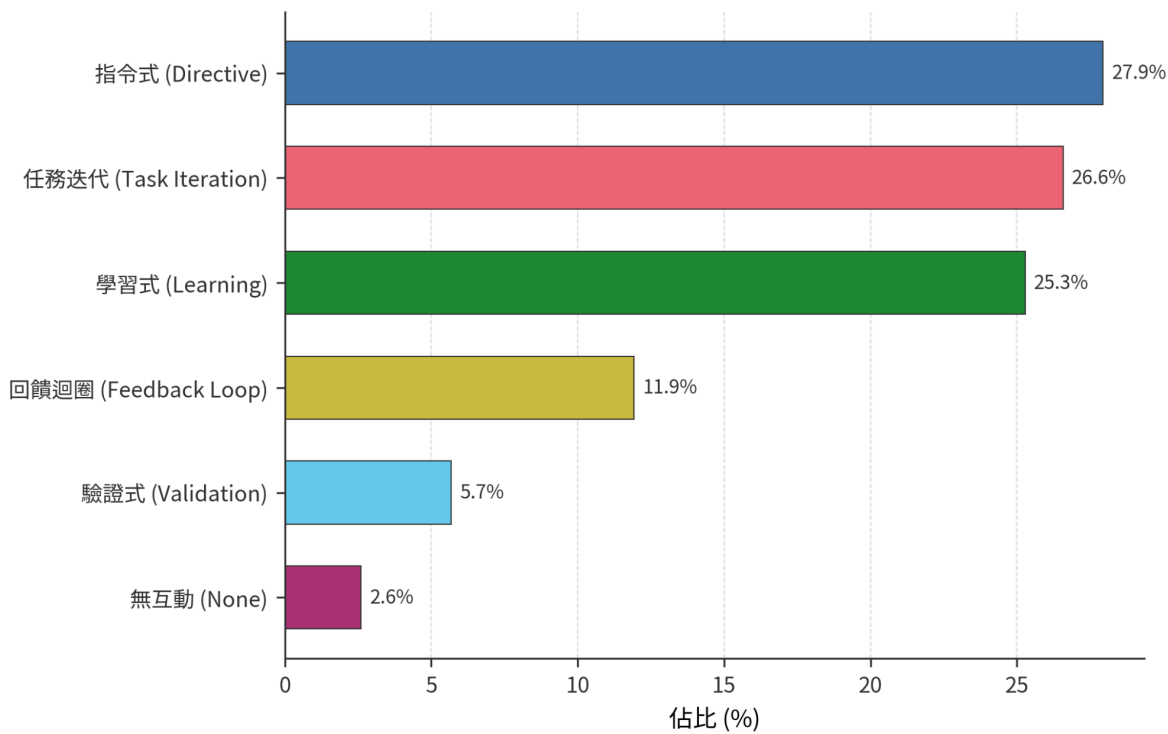

**圖 A2. 台灣 Claude 使用者人機協作模式分布（N=7,729）**

協作模式分布的最顯著特徵為三者佔比相近的均衡格局：指令式（directive, 27.9%）、任務迭代（task iteration, 26.6%）與學習式（learning, 25.3%）三者合計達 79.8%，且無單一模式佔據絕對多數。此一分布顯示，台灣使用者並非僅以單向指令操作 AI，而是在相當程度上進行多輪對話式互動。

回饋迴圈（feedback loop, 11.9%）加上任務迭代（26.6%），意味著 38.5% 的對話涉及反覆修正。此比例契合學術寫作的工作本質——論文草稿通常需要經歷多輪修訂，AI 可在此迭代過程中扮演即時回應的協作角色。

學習式使用佔 25.3%，顯示四分之一的使用者將 Claude 視為學習管道。此模式對缺乏程式設計或統計背景的人文社會科學研究者尤為重要：AI 可作為「技術中介」，降低跨領域方法的入門門檻。

---

# 第 3 節　讀懂這份資料——O*NET 與任務分類方法

本節為背景知識說明，旨在協助不熟悉 AEI 資料結構的讀者理解後續分析所使用的分類框架。

## 3.1 什麼是 O*NET

O*NET（Occupational Information Network）是美國勞工部建立的職業資訊網絡資料庫，涵蓋約 1,000 種職業及其數千項具體工作任務的標準化描述。每項職業任務以一段精確的英文文字定義，例如「修改現有軟體以修正錯誤、適應新硬體或改善其效能」即為軟體開發人員的一項典型任務。

Anthropic 在 AEI 資料中使用 AI 分類器，判斷每筆 Claude 對話內容最接近 O*NET 資料庫中哪一項職業任務。例如，「modify existing software to correct errors...」佔台灣對話的 5.49%，代表 5.49% 的對話內容在性質上最接近軟體開發人員日常執行的這項工作。

O*NET 分類回答的核心問題是：台灣人用 Claude 做的事，在傳統職業世界裡對應到誰的什麼工作？ 此分類使我們得以將 AI 使用行為與既有的勞動市場結構對接，評估 AI 對不同職業領域的潛在影響。

須注意的限制：O*NET 係以美國勞動市場為基礎建立，對台灣的適用性存在落差。此外，台灣資料中有 39.18% 被標為 not_classified（無法對應至任何現有職業任務），另有 4.58% 標為 none。合計近 44% 的對話無法歸類，反映出相當比例的 AI 使用行為超出了傳統職業任務的描述範疇。

### 3.2 什麼是 Request 任務分類層級

除了 O*NET 職業任務對應之外，Anthropic 另以階層式分類法（Request taxonomy）將每筆對話歸入三層任務類別：

- Level 2（最粗）：大類別，涵蓋範圍最廣。例如「翻譯、編輯、格式化、摘要文件並輔助語言學習」。
- Level 1（中間）：歸併相近的 Level 0 類別。例如「翻譯各語言間的文件」。
- Level 0（最細）：非常具體的任務描述。例如「在亞洲語言與英語之間翻譯文件」。

以圖書館分類作類比：Level 2 如同「文學類」書架，Level 1 是「小說」區域，Level 0 則是「日本推理小說」。 三個層級共同構成一套由粗到細的任務分類體系，使分析者可在不同粒度上觀察 AI 使用行為的分布。

### 3.3 任務複雜度的衡量指標

AEI 資料包含六項數值型指標，用以量化每筆對話所涉及的任務複雜度與 AI 介入程度。表 A-2 整理各指標的定義與台灣實際數據。

表 A-2：台灣 Claude.ai 任務複雜度指標一覽（N=7,729）

| 指標英文名 | 中文說明 | 台灣數據 | 白話解讀 |
|---|---|---|---|
| human_education_years | 完成該任務所需的人類教育年數 | $M$ = 12.49, 95% CI [12.41, 12.58] | 約等於大學畢業程度 |
| ai_education_years | AI 表現等效的教育年數 | $M$ = 12.38, 95% CI [12.31, 12.46] | AI 表現約等於大學畢業生 |
| human_only_time | 人類獨力完成所需時間 | $Mdn$ = 1.75 小時 | 通常是一到兩小時的工作 |
| human_with_ai_time | 有 AI 協助的完成時間 | $Mdn$ = 12 分鐘 | 大幅壓縮至十幾分鐘 |
| ai_autonomy | AI 自主程度（1-5 分） | $Mdn$ = 4.0 | 使用者讓 AI 高度自主產出 |
| human_only_ability | 人類能否獨力完成 | 82.9% 可以 | 大部分是「能做但讓 AI 做」 |

綜合上述指標，可以勾勒出台灣 Claude 使用的典型畫像：一項大學程度、人類本來就能做、但做起來要一兩個小時的工作——使用者選擇讓 AI 以高自主性在十幾分鐘內完成。這是一幅效率加速器的使用圖像，而非能力替代。

---

## 第 4 節　任務類型排行

圖 A-3 呈現 Request Level 2 層級（最粗分類）中佔比最高的 12 類任務，排除 not_classified。

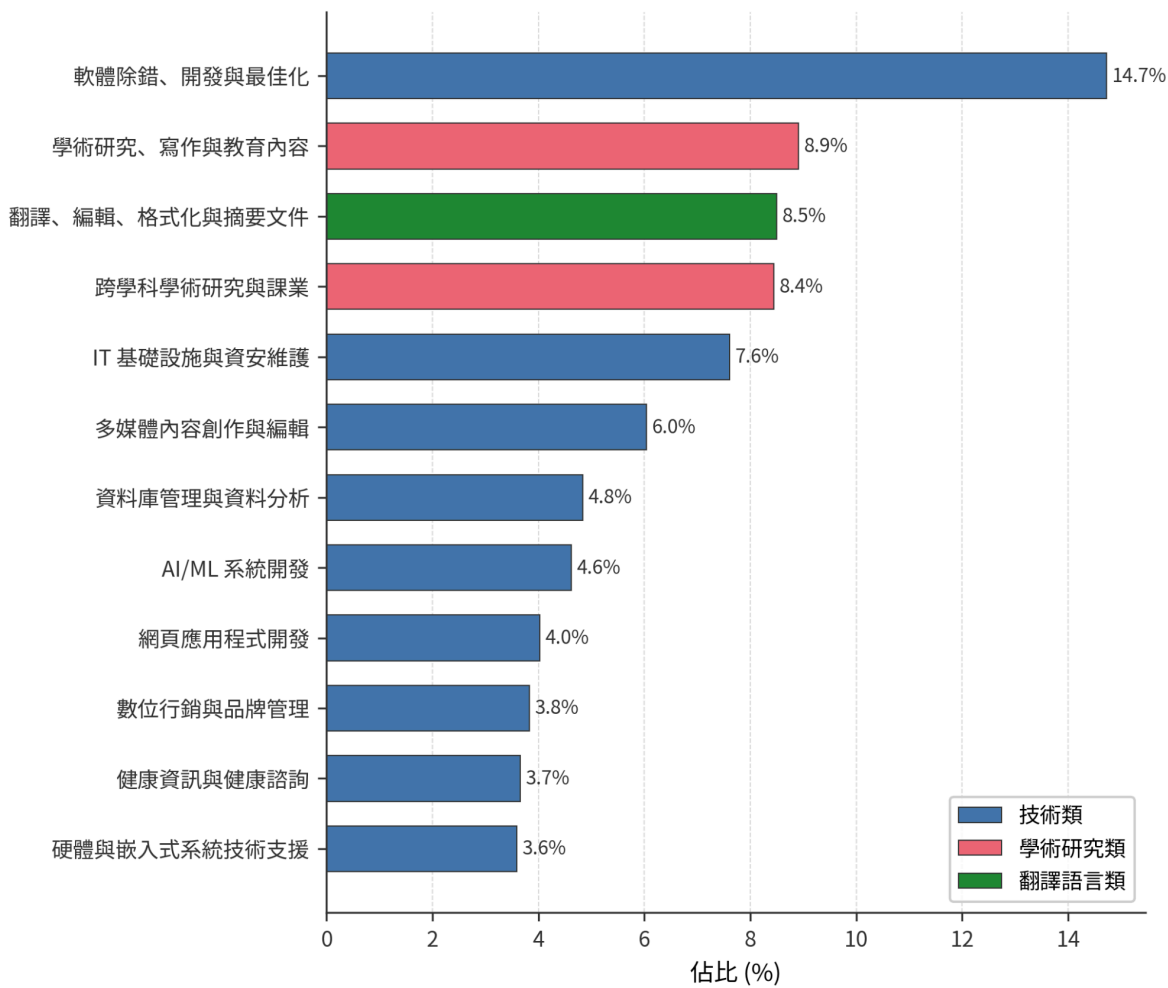

**圖 A3. 台灣 Claude 使用者主要任務類型排行（Request L2, Top 12, N=7,729）**

軟體除錯、開發與最佳化以 14.7% 佔據首位，反映程式開發為 Claude 最普遍的用途。然而，排名第二至第四的類別均與學術或語言工作直接相關：

- 學術研究、寫作與教育內容：8.9%
- 翻譯、編輯、格式化與摘要文件：8.5%
- 跨學科學術研究與課業：8.4%

學術研究寫作兩個類別合計佔 17.3%，翻譯文件處理佔 8.5%，三者合計約 25.8%——超過四分之一的 Claude 對話與學術或語言任務相關。此佔比僅略低於整體技術類任務（軟體 + IT + AI/ML 合計約 27%），顯示學術場域已是台灣 AI 使用的第二大場域。

對人文社會科學研究者而言，翻譯與文件處理的高佔比（8.5%）尤其值得關注。在台灣中英雙語學術環境中，翻譯與文稿編修為研究者日常工作的核心組成，此類別的高排名暗示研究者已廣泛運用 AI 處理語言相關任務。

## 第 5 節　O*NET 職業任務分析

### 5.1 Top 15 O*NET 任務

圖 A-4 呈現排除 not_classified 和 none 後，佔比最高的 15 項 O*NET 職業任務。

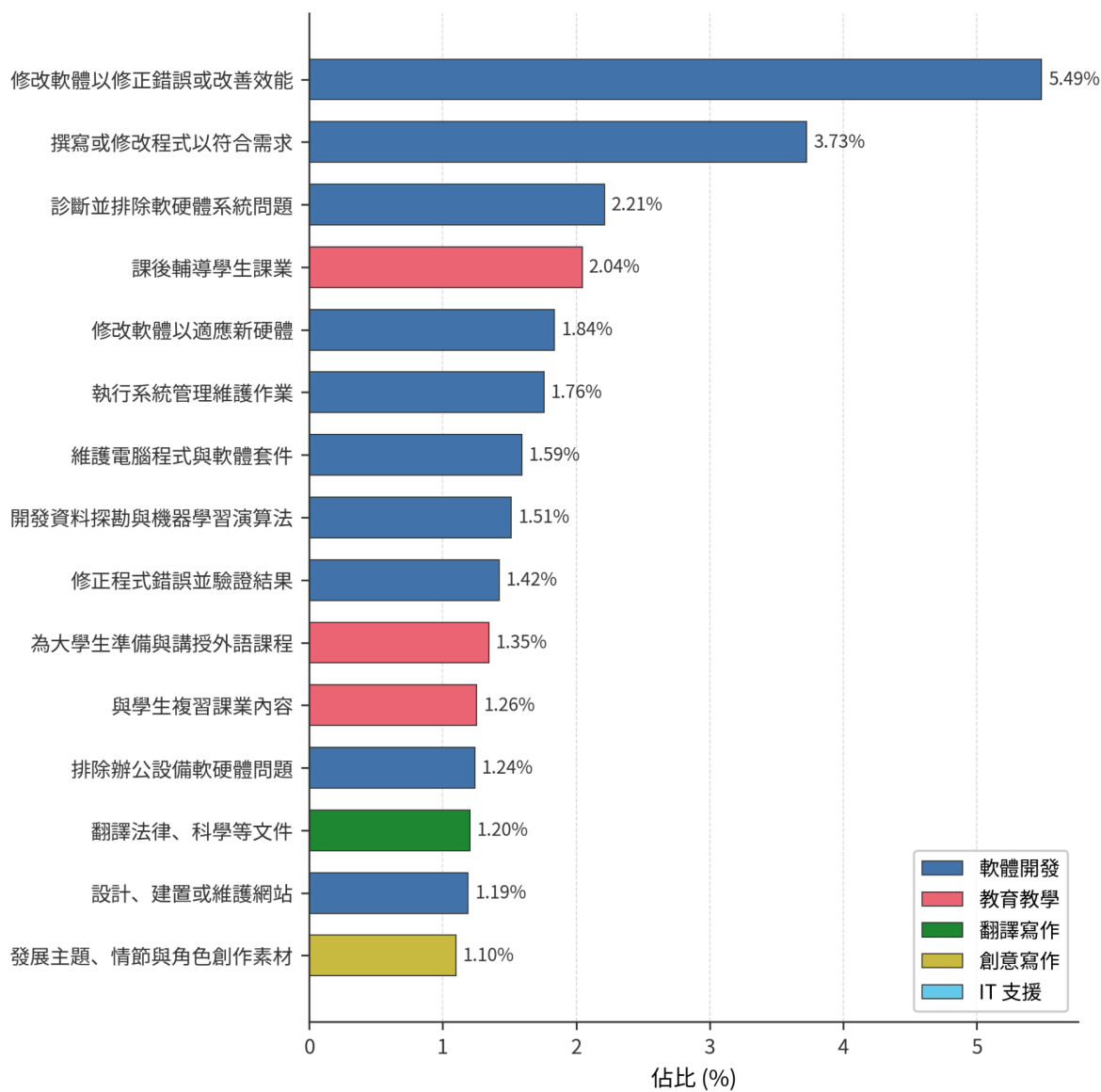

**圖 A4. 台灣 Claude 使用者 O*NET 職業任務對應排行（Top 15, N=7,729）**

前 15 項任務以軟體開發相關為主（藍色），但教育教學（紅色）與翻譯寫作（綠色）亦佔有一席之地。教育相關任務中，「課後輔導學生課業」（2.04%）、「為大學生準備與講授外語課程」（1.35%）、「與學生複習課業內容」（1.26%）合計 4.65%，顯示教學活動在 AI 使用中的顯著存在。

## 5.2 五大領域歸納

表 A-3 將 O*NET 任務歸納為五大領域，呈現各領域的代表性任務、合計佔比及其影響的台灣學科領域。

表 A-3: O*NET 任務五大領域歸納

| 領域 | 代表性任務 | 合計佔比 | 影響的台灣學科領域 |
|------|-----------|---------|-----------------|
| 軟體開發與 IT | 修改軟體修正錯誤、撰寫程式、系統管理 | ~31.7% | 資訊工程、資訊管理 |
| 教育與教學 | 課後輔導、講授外語、複習課業 | ~13.5% | 教育學、外語教學、各學科教學 |
| 翻譯、寫作與編輯 | 翻譯文件、發展創作素材 | ~7.5% | 外國語文、翻譯研究、中文系、傳播學 |
| 諮商與輔導 | 提供社會服務、心理輔導相關對話 | ~1.3% | 社會工作、心理諮商 |
| 學術研究 | 文獻分析、資料處理、研究方法 | ~0.5% | 跨領域學術研究 |

教育教學(13.5%)加上翻譯寫作與編輯(7.5%)合計約 21%。值得注意的是，這些正是人文社會科學研究者的核心工作技能——備課教學、翻譯文獻、撰寫論文、編輯文稿。O*NET 分類雖以美國職業體系為基礎，但此結果仍清楚指向：AI 正在介入人文社會科學研究者最日常的工作內容。

學術研究本身的佔比(~0.5%)看似極低，但這反映的是 O*NET 分類的特性——O*NET 並無專門的「學術研究者」任務類別，學術研究工作被拆解為「教學」「翻譯」「寫作」「資料分析」等具體任務後，散布於多個類別之中。因此，學術研究對 AI 的實際使用量遠高於 O*NET 直接歸類的 0.5%。

# 第 6 節 教育程度與時間效益

## 6.1 教育年數對比

圖 A-5 比較人類教育年數與 AI 等效教育年數。

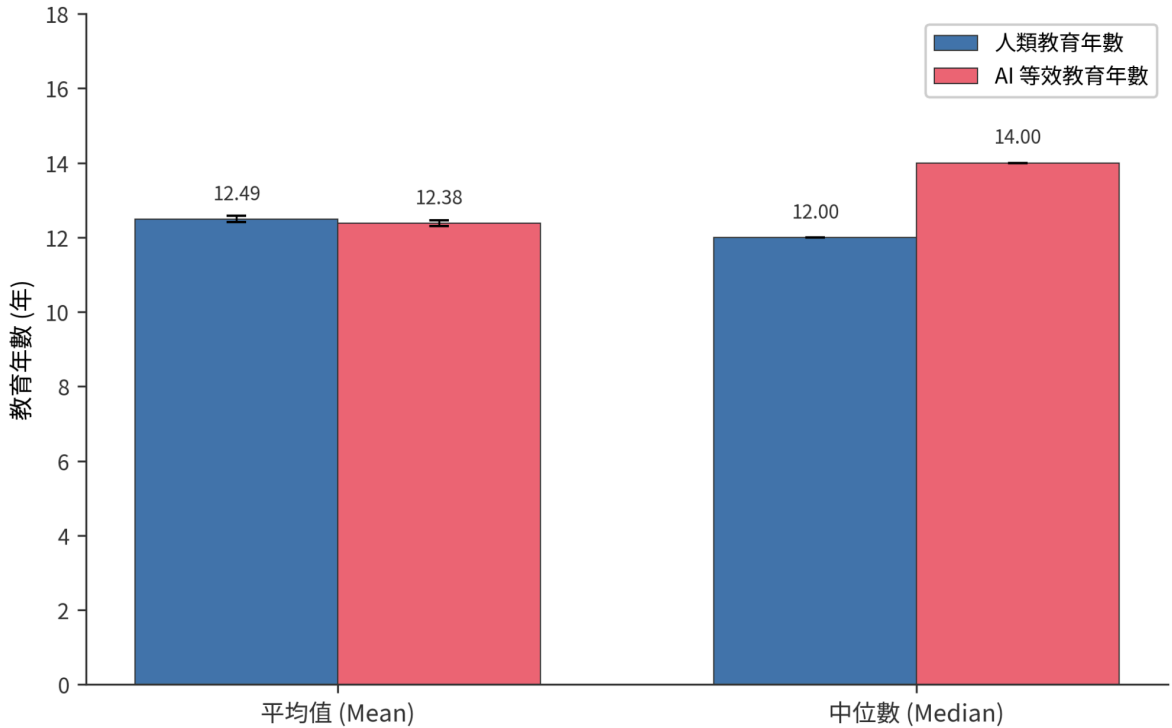

**圖 A5. 任務教育年數與 AI 等效教育年數對比（N=7,729，平均值含 95% CI 誤差線）**

五項關鍵數據：

| 指標 | 數值 |
|---|---|
| 任務所需人類教育年數 | *M* = 12.49 年, 95% CI [12.41, 12.58] |
| AI 等效教育年數 | *M* = 12.38 年, 95% CI [12.31, 12.46] |
| 人類教育年數中位數 | *Mdn* = 12.00 年 |
| AI 等效教育年數中位數 | *Mdn* = 14.00 年 |
| 平均值差異 | 0.11 年（CI 重疊, 差異不顯著） |

人類教育年數平均 12.49 年與 AI 等效教育年數平均 12.38 年高度接近，兩者的 95% 信賴區間存在重疊，顯示差異在統計上不具顯著性。此結果意味著：使用者指派給 AI 的任務複雜度，大致等同於自身教育程度所能處理的水準。AI 的表現被評估為與大學畢業程度的人類相當。

中位數方面，AI 等效教育年數（14.0 年）略高於人類教育年數（12.0 年），暗示在「典型」任務中，AI 的表現甚至略優於任務本身所需的教育水準。

## 6.2 時間效益對比

圖 A-6 比較人類獨力完成時間與 AI 協助時間。

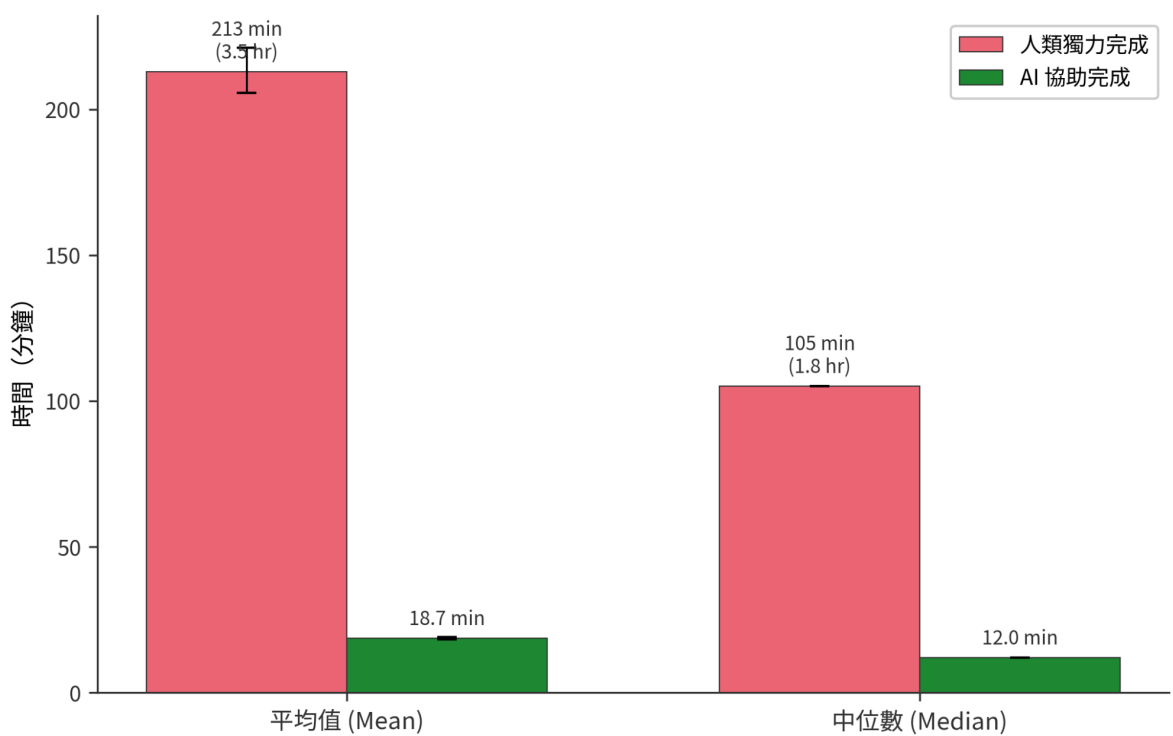

**圖 A6. 人類獨力完成時間 vs. AI 協助時間對比（N=7,729，平均值含 95% CI 誤差線）**

| 指標 | 數值 |
| --- | --- |
| 人類獨力時間（平均） | *M* = 3.55 小時（212.8 分鐘）, 95% CI [3.43, 3.68] hr |
| AI 協助時間（平均） | *M* = 18.7 分鐘, 95% CI [18.2, 19.2] min |
| 人類獨力時間（中位數） | *Mdn* = 1.75 小時（105 分鐘） |
| AI 協助時間（中位數） | *Mdn* = 12.0 分鐘 |
| 中位數時間節省率 | 約 89%（105 分鐘 → 12 分鐘） |

以中位數計算，人類獨力完成需要 1.75 小時，AI 協助下僅需 12 分鐘，時間節省率約 89%。即便以較保守的中位數估計，AI 仍將典型任務的完成時間壓縮至原本的約九分之一。

對人文社會科學研究者而言，此效率增益在翻譯校對（典型耗時 2-3 小時的英文論文翻譯可壓縮至 15-20 分鐘）、教案製作（備課材料整理從數小時縮短至半小時）等

場景中影響尤為顯著。然而，須注意此處比較的是「人類估計獨力完成時間」與「實際 AI 協助互動時間」，前者為估計值，後者為實測值，兩者的測量基礎不完全相同。

## 第 7 節　對人文社會科學領域學術研究者的影響推論

圖 A-7 呈現與人文社會科學學術直接相關的 14 項 Request L1 任務類別。

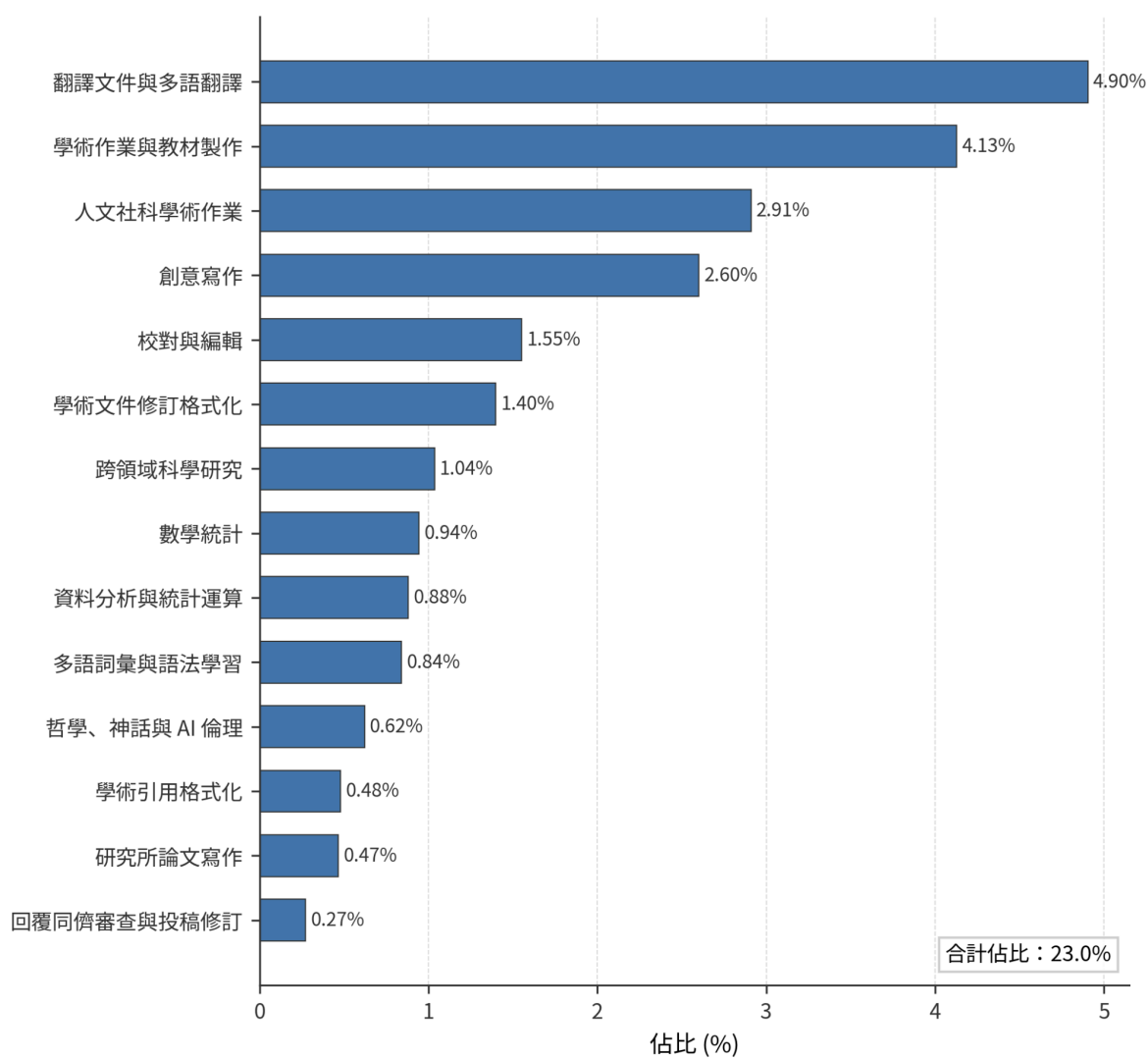

**圖 A7. 人文社會科學學術相關任務類型分布（Request L1, N=7,729）**

14 項人文社會科學相關任務合計佔全部對話的 23.0%，接近四分之一。以下從五個面向推論 AI 對台灣人文社會科學領域學術研究者的影響。

推論一：翻譯與多語處理是最普遍的 AI 輔助場景

翻譯文件與多語翻譯在 L1 層級佔 4.90%，為人社相關任務中佔比最高者。在 L0 層級中，「亞洲語言與英語之間翻譯文件」佔 2.91%，多語詞彙與語法學習另佔 0.84%。跨語言工作是台灣人文社會科學研究者最直覺的 AI 切入點。

此現象反映台灣學術環境的結構性特徵：研究者需要大量閱讀英文文獻並以中文撰寫報告（或反之），翻譯為日常工作中佔比極高但附加價值相對低的任務。AI 正大幅降低此一語言障礙，長遠而言可能重塑「英語能力」在學術能力評估中的權重——當翻譯不再是瓶頸，語言能力的稀缺性將降低，學術評價體系可能需要重新校準。

推論二：學術寫作的「全流程 AI 輔助」正在成形

從資料中可辨識出學術寫作各環節均已有對應的 AI 輔助任務：

| 寫作環節 | 對應 L1 任務 | 佔比 |
|---|---|---|
| 文獻翻譯 | 翻譯文件與多語翻譯 | 4.90% |
| 草稿撰寫 | 學術作業與教材製作；人文社科學術作業 | 4.13% + 2.91% |
| 修訂格式化 | 學術文件修訂格式化 | 1.40% |
| 引用管理 | 學術引用格式化 | 0.48% |
| 回覆審查 | 回覆同儕審查與投稿修訂 | 0.27% |

AI 已不只是「寫作助手」，而是貫穿從文獻翻譯、草稿撰寫、修訂格式化、引用管理到回覆審查意見的全流程協作工具。此趨勢意味著學術寫作的每一個環節都已進入 AI 可介入的範圍，「AI 輔助寫作」正從單點工具演變為系統性的工作流程。

推論三：課業場景的比重暗示研究生高度依賴

課業場景佔整體使用的 19.1%，人文社會科學學術作業佔 2.91%，學術文件修訂佔 1.40%，學習式協作模式佔 25.3%。綜合觀之，相當比例的使用者將 Claude 當作學習工具或知識顧問，而非純粹的代勞工具。

研究生群體可能是此模式的主要驅動者。研究生同時身兼「學習者」與「研究者」雙重角色，其使用 AI 的模式兼具學習式（理解新概念、探索方法論）與任務迭代式（撰寫並修改論文）。25.3% 的學習式協作佔比支持此推論：使用者在相當程度上將 AI 視為「可以隨時提問的指導者」。

推論四：時間節省效果對研究生產力有根本性影響

中位數時間節省率約 89%（1.75 小時 → 12 分鐘），此效率增益若投射至學術工作的具體場景：一篇英文論文的翻譯校對從 3 小時縮短至 20 分鐘，一學期若有 20 篇文獻需要翻譯處理，累計可節省約 50 小時。

此一生產力提升具有根本性影響：研究者可將節省的時間投入更具創造性的工作——深度思考、理論建構、田野調查——而非耗費於結構化的文書處理。然而，時間節省亦帶來「AI 輔助」與「AI 代寫」之間界線模糊的學術倫理挑戰。當 AI 可在數分鐘內產出大學程度的文字（ai_education_years 中位數 = 14.0 年），如何界定合理使用的邊界，將成為學術機構無法迴避的議題。

### 推論五：人類仍可獨力完成大部分任務，AI 扮演加速角色

82.9% 的任務被評估為人類可獨力完成，但 AI 自主性中位數高達 4.0/5.0。此一組合描繪出「能做但讓 AI 做」的新常態：使用者具備完成任務的能力，但選擇將執行委託給 AI，以節省時間與認知負荷。

對人文社會科學學界而言，此數據挑戰了「AI 將取代人文學者」的極端論述，同時也否定了「人文學者不需要 AI」的保守立場。更準確的描述是：AI 正在重新分配人文社會科學研究者的認知資源配置——將結構化、重複性的工作（翻譯、格式化、文獻整理）交由 AI 加速處理，釋出時間與注意力投入需要人類判斷力的核心工作（理論建構、脈絡化詮釋、倫理反思）。

### 影響總結

表 A-4：AI 對台灣人文社會科學學術研究者的五大影響面向

| 面向 | 數據依據 | 對研究者的具體影響 | 風險與挑戰 |
|---|---|---|---|
| 語言障礙降低 | 翻譯佔 L1 第二高（4.90%），亞洲語言翻譯（2.91%） | 英文論文閱讀與寫作門檻大幅降低；非英語母語研究者的國際發表機會增加 | 翻譯品質的專業術語準確性仍需人類審查；過度依賴可能弱化研究者自身語言能力 |
| 寫作效率提升 | 學術寫作全流程均有對應 AI 任務（草稿→修訂→引用→審查回覆） | 論文產出週期縮短；多輪修訂效率提升 | 「AI 輔助」與「AI 代寫」界線模糊；學術原創性認定標準需重新定義 |

| 教學模式轉型 | 教育教學佔 O*NET 13.5%；課業場景佔 19.1% | 教師可運用 AI 製作教材、設計評量；學生以 AI 作為課後學習輔助 | 傳統考評方式（課堂報告、期末論文）的有效性受到挑戰 |
| --- | --- | --- | --- |
| 研究方法擴展 | 資料分析（0.88%）、數學統計（0.94%）、學習式協作（25.3%） | 缺乏量化背景的人社研究者可借助 AI 跨越統計分析與程式設計門檻 | AI 輔助的統計分析可能導致「黑箱」式方法應用，研究者未必理解所使用方法的假設與限制 |
| 學術倫理新課題 | 82.9% 人類可獨力完成但選擇 AI 代勞；AI 自主性中位數 4.0/5.0 | 「能做但讓 AI 做」成為新常態，需建立明確的 AI 使用揭露規範 | 學術誠信判定標準尚未跟上技術發展；師生間對 AI 合理使用的認知可能存在落差 |

## 第 8 節　結論與建議

台灣 Claude.ai 使用資料描繪了一幅以「效率加速」為核心的使用圖像：82.9% 的任務人類本可獨力完成，但借助 AI 後完成時間中位數從 1.75 小時壓縮至 12 分鐘。學術相關任務（含翻譯）佔比達 25.8%，顯示 AI 在台灣學術場域的滲透已具一定規模。對人文社會科學研究者而言，翻譯校對、學術寫作與教學備課為最主要的 AI 使用場景，而「全流程 AI 輔助寫作」的趨勢正從可能變為現實。

基於上述分析，提出以下五點建議：

1. 建立明確的 AI 使用指引：學術機構應制定具體且可操作的 AI 使用規範，區分「合理輔助」（如翻譯校對、格式整理）與「學術不誠實」（如未揭露的 AI 代勞），而非以籠統的禁令回應。
2. 將 AI 素養納入研究方法論課程：與其一味禁止，不如在研究方法課程中系統性地教授 AI 工具的合理使用方式、限制與風險，培養研究生具備「批判性使用 AI」的能力。
3. 重新評估語言能力在學術評量中的角色：當 AI 翻譯品質持續提升，「英語寫作能力」作為學術能力指標的權重可能需要調整。學術評量應更聚焦於研究設計、分析深度與理論原創性，而非語言表達的流暢度。
4. 關注 AI 對深度思考與原創洞見的潛在侵蝕：AI 的效率優勢可能使研究者傾向於快速產出而非深度思考。學術社群應有意識地維護「慢思考」的價值，確保 AI 加速的是例行性工作，而非取代了需要時間沉澱的知識創造。
5. 善用效率優勢，將節省時間投入研究創新：89% 的時間節省率意味著巨大的認知資源釋放。研究者應有策略地運用此效率紅利——將省下的時間投入田野調

查、跨學科對話、理論反思等 AI 尚無法替代的高價值工作，而非僅以 AI 加速既有的工作模式。

---

資料來源聲明

本報告分析所使用之資料來自 Anthropic Economic Index（AEI）第四版公開原始資料。

Anthropic. (2025). *The Anthropic Economic Index (4th ed.)*. Retrieved from https://www.anthropic.com/research/economic-index-primitives

資料集：`aei_raw_claude_ai_2025-11-13_to_2025-11-20.csv`（台灣子集）資料下載：https://huggingface.co/datasets/Anthropic/EconomicIndex 資料區間：2025 年 11 月 13 日至 2025 年 11 月 20 日 平台：Claude AI (Free and Pro) 地理範圍：台灣（geo_id = TW）樣本數：N = 7,729 筆對話
本報告中所有數據均直接提取自上述原始資料，未經任何統計推論或模型估計。圖表由 Python（matplotlib 3.10）產生，原始碼存於 `scripts/generate_all_charts.py`。

---

附錄參考文獻

Acemoglu, D., & Restrepo, P. (2019). Automation and new tasks: How technology displaces and reinstates labor. *Journal of Economic Perspectives*, 33(2), 3-30. https://doi.org/10.1257/jep.33.2.3

Eloundou, T., Manning, S., Mishkin, P., & Rock, D. (2024). GPTs are GPTs: An early look at the labor market impact potential of large language models. *Science*, 384(6702), 1306-1308. https://doi.org/10.1126/science.adj0998